\documentclass[11pt,en,cite=authoryear]{note}

\title{Sparse Kronecker Product Decomposition:  A General Framework of Signal Region Detection in Image Regression}
\author{Sanyou Wu \\ sanyouwu@connect.hku.hk \and Long Feng \\ lfeng@hku.hk}
\institute{Department of Statistics \& Actuarial Science, The University of Hong Kong}

\date{}

\usepackage{appendix}
\usepackage{array}
\usepackage{subfigure}
\usepackage{graphicx}
\usepackage{amssymb,amsmath,amsthm}

\usepackage{xcolor}
\usepackage{graphicx}
\graphicspath{
	{note/figure/}{note/image}{./image/}
}
\let\savehash\hash
\let\hash\relax
\usepackage{mathabx}
\let\hash\savehash
\definecolor{pinegreen}{rgb}{0.0, 0.47, 0.44}
\usepackage{algorithm}
\usepackage{algorithmic}
\usepackage{amsmath}
\usepackage{amsfonts}
\newtheorem{thm}{\textbf{Theorem}}

\newtheorem{cor}{\textbf{Corollary}}
\newtheorem{condition}{\textbf{Condition}}

\def\bbB{\widebar{\mathbf{B}}}
\def\bbA{\widebar{\mathbf{A}}}
\def\bbb{\widebar{\mathbf{b}}}
\def\bba{\widebar{\mathbf{a}}}
\def\hbbA{\widehat{\widebar{\mathbf{A}}}}
\def\hbbB{\widehat{\widebar{\mathbf{B}}}}
\def\hbba{\widehat{\widebar{\mathbf{a}}}}
\def\hbbb{\widehat{\widebar{\mathbf{b}}}}

\def\tbbA{\tilde{\bbA}}
\def\tbba{\tilde{\bba}}
\def\tbbX{\widetilde{\widebar{\bX}}}
\def\bmB{\mathbf{\mathcal{B}}}
\def\bmA{\mathbf{\mathcal{A}}}
\def\bmC{\mathbf{\mathcal{C}}}
\def\mC{\mathcal{C}}
\def\bmB{{\bf{\mathcal{B}}}}
\def\bmA{{\bf{\mathcal{A}}}}
\def\bmX{{\bf{\mathcal{X}}}}
\def\bmC{{\bf{\mathcal{C}}}}

\def\tr{\text{tr}}
\def\mR{\mathcal{R}}

\def\argmin{\operatorname{argmin} \displaylimits}

\def\d{\delta}

\def\E{\mathbb{E}}

\def\T{\top}

\def\mC{\mathcal{C}}

\def\tvec{\text{vec}}

\def\mbR{\mathcal{\widebar{R}}}

\newcommand{\bel}{\begin{eqnarray}\label}
\newcommand{\eel}{\end{eqnarray}}
\newcommand{\bes}{\begin{eqnarray*}}
\newcommand{\ees}{\end{eqnarray*}}
\newcommand{\bei}{\begin{itemize}}
\newcommand{\beiftnt}{\begin{itemize}\footnotesize}
\newcommand{\eei}{\end{itemize}}

\def\benu{\begin{enumerate}}
\def\eenu{\end{enumerate}}

\def\argmin{\mathop{\rm arg\, min}}

\def\E{{\mathbb{E}}}

\def\P{{\mathbb{P}}}

\def\complex{\mathop{{\rm I}\kern-.58em\hbox{\rm C}}\nolimits}

\def\diag{\hbox{\rm diag}}

\def\rank{\hbox{\rm rank}}

\def\sgn{\hbox{\rm sgn}}

\def\mathbold{\boldsymbol} 

\def\ba{\mathbold{a}}
\def\hba{{\widehat{\ba}}}\def\tba{{\widetilde{\ba}}}
\def\bA{\mathbold{A}}
\def\hbA{{\widehat{\bA}}}

\def\bb{\mathbold{b}}
\def\hbb{{\widehat{\bb}}}
\def\bB{\mathbold{B}}
\def\hbB{{\widehat{\bB}}}

\def\hbC{{\widehat{\bfC}}}

\def\bC{\mathbold{C}}
\def\hbC{{\widehat{\bC}}}

\def\bD{\mathbold{D}}

\def\bE{\mathbold{E}}

\def\bh{\mathbold{h}}
\def\hbh{{\widehat{\bh}}}

\def\bI{\mathbold{I}}

\def\bM{\mathbold{M}}

\def\bU{\mathbold{U}}

\def\bv{\mathbold{v}}

\def\bV{\mathbold{V}}

\def\bw{\mathbold{w}}

\def\bW{\mathbold{W}}

\def\bX{\mathbold{X}}\def\tbX{{\widetilde{\bX}}}

\def\by{\mathbold{y}}

\def\bz{\mathbold{z}}

\def\bZ{\mathbold{Z}}

\def\ep{\varepsilon}\def\eps{\epsilon}
\def\bep{\mathbold{\ep}}

\def\ttheta{\widetilde{\theta}}

\def\bTheta{\mathbold{\Theta}}

\def\lam{\lambda}

\def\bSigma{\mathbold{\Sigma}}

\newcommand\lfR[1]{{\color{black} #1}}

\begin{document}
\maketitle

\begin{abstract}
This paper aims to present the first Frequentist framework on signal region detection in high-resolution and high-order image regression problems.
Image data and scalar-on-image regression are intensively studied in recent years. However, most existing studies on such topics focused on outcome prediction, while the research on image region detection is rather limited, even though the latter is often more important. 
In this paper, we develop a general framework named Sparse Kronecker Product Decomposition (SKPD) to tackle this issue.
The SKPD framework is general in the sense that it works for both matrices (e.g., 2D grayscale images) and (high-order) tensors (e.g., 2D colored images, brain MRI/fMRI data) represented image data. Moreover, unlike many Bayesian approaches, our framework is computationally scalable for high-resolution image problems.
Specifically, our framework includes: 1) the one-term SKPD; 2) the multi-term SKPD; and 3) the nonlinear SKPD. We propose nonconvex optimization problems to estimate the one-term and multi-term SKPDs and develop path-following algorithms for the nonconvex optimization. The computed solutions of the path-following algorithm are guaranteed to converge to the truth with a particularly chosen initialization even though the optimization is nonconvex. Moreover, the region detection consistency could also be guaranteed by the one-term and multi-term SKPD. The nonlinear SKPD is highly connected to shallow convolutional neural networks (CNN), particular to CNN with one convolutional layer and one fully connected layer. Effectiveness of SKPDs is validated by real brain imaging data in the UK Biobank database. 
\end{abstract}

\section{Introduction}\label{sec-1}
This paper aims to address an important challenge in high-dimensional image regression problems: signal region detection. Specifically, we aim to develop a general framework to detect the signal regions in image data (represented as matrices or tensors) that are associated with a scalar outcome.
Our study is first motivated by using brain imaging data to understand the mechanisms of intellectual disability. By \citet{daily2000identification}, there are 2\%-3\% of general population that are affected by intellectual disability. On the other hand, more than 60\% of intellectual disabilities still have unknown causes \citep{vos2015global}. Therefore, understanding how different parts of brain are related to various intellectual disabilities is an increasingly important goal of psychiatry.

Variable selection has been intensively studied in high-dimensional regression models over the past two decades. 
However, signal region detection is far more than  generalizations of variable selection due to the uniqueness of image data. In general, a 2D grayscale image is represented as a matrix $\bX\in\mathbb{R}^{D_1\times D_2}$, while a colored 2D image is represented as a three order tensor $\mathcal{X}\in\mathbb{R}^{D_1\times D_2\times D_3}$, with $D_3=3$ indicating three color channels: red, green, blue. Beyond 2D image, a brain magnetic
resonance imaging (MRI) scan produces a three order tensor  $\mathcal{X}\in\mathbb{R}^{D_1\times D_2\times D_3}$, while functional MRI (fMRI) scan produces an even higher-order tensor. 
As all the imaging data contains rich spatial and structural information, simply vectorizing the matrix/tensor image and treating obtained pixels/voxels as independent variables would not only generate ultra high-dimensional vectors and face computational problems, but also omit the spatial structure and breakdown the signal region.
Therefore, a statistical approach that could effectively detect signal regions in both matrix and tensor represented image data in a uniform way is urgently desired. This is the methodological motivation of this project.

Although image data and image regression problems have been intensively studied in recent years, most existing research focus on outcome prediction, while the studies on signal region detection are relatively limited. The most related studies are from Bayesian perspectives, where regression coefficient is first vectorized and then modeled with certain prior distributions to detect signal regions. For example, the Ising prior is used in \citet{goldsmith2014smooth} and \citet{li2015spatial}; the soft-thresholdings of a latent Gaussian process is proposed by \citet{kang2018scalar}; and continuous shrinkage priors is applied on \citet{jhuang2019spatial}. However, due to the restrictions on posterior computation, most Bayesian approaches are difficult to handle large 2D image, not mentioning large MRI or fMRI data that contains significantly more pixels/voxels. We also note that Bayesian approaches have also been applied to other image regression problems beyond signal region detection in the literature \citep{boehm2015spatial,feng2019bayesian}.

Image regression problems have also been studied from Frequentist perspective, although to the best of our knowledge, signal region detection was not addressed directly. Total Variation (TV) \citep{rudin1992nonlinear,rudin1994total} and fused Lasso \citep{tibshirani2005sparsity} based approaches have been commonly applied for image denoising and recovery. In particular, \citet{wang2017generalized} proposed a TV based penalization approach to promote the piecewise smoothness of image coefficients. Moreover, \citet{reiss2010functional} extended functional principal component regression for image data 
and used B-splines to approximate the image coefficients and enforce smoothness.
\citet{reiss2015wavelet} proposed a set of wavelet procedures for image regression and conducted a permutation-based approach to test the effects of image predictors.
When image data is represented as high-order tensors, dimension reduction has become a core problem in the analysis. \citet{zhou2013tensor} proposed a  tensor image regression framework that uses canonical polyadic decomposition (CPD) to reduce the image coefficients dimension. \citet{feng2020brain} further proposed an Internal Variation (IV) penalization approach built on CPD to mimic the effects of Total Variation and promote smoothness of tensor coefficients. 

Beyond statistics community, the convolutional neural networks (CNN; \citet{fukushima1982neocognitron,lecun1998gradient}) is arguably the most popular approach for image prediction problems in recent years. With the advancement of modern computational power, CNN could introduce  thousands and even millions of unknown parameters in the composition of many nonlinear functions to obtain the optimal prediction accuracy.
On the other hand, with these many parameters presented in a ``black box'', it is extremely difficult to interpret a CNN model, not mentioning detecting the signal regions. In fact, improving the interpretability of CNN has a become popular topic in the computer vision community, we refer to Section \ref{sec-5} for a detailed literature review.

This paper aims to provide the first Frequentist framework on signal region detection in high-resolution and high-order image regression problems.  Toward this goal, we explore the potential of Kronecker product and propose a series of models named Sparse Kronecker Product Decomposition (SKPD). The SKPD models include two components: the ``dictionaries'' and ``location indicators''. The ``dictionaries'' aim to catch the ``shapes'' and ``intensities'' of the signal, while the ``location indicators'' are assumed to be sparse and aim to find the locations of the signal.
In the literature, Kronecker product has been commonly used when analyzing matrix-valued data and has become a powerful tool for 
matrix dimension reduction. For example, \citet{cai2019kopa} used Kronecker product for matrix approximation and denoising; \citet{hafner2020estimation} proposed a Kronecker product model for covariance or correlation matrix estimation; \citet{chen2020constrained} investigated Kronecker product for matrix autoregressive models, etc.  

The SKPD framework includes: one-term SKPD, multi-term SKPD, and nonlinear SKPD.    
This framework is general in the sense that it works for both matrices and tensors represented image data. We propose nonconvex optimization problems to estimate the one-term and multi-term SKPDs and develop path following algorithms for the nonconvex optimization. Under an restricted isometry property (RIP), the computed solutions of the path following algorithm are guaranteed to converge to the truth with a particularly chosen initialization even though the optimization is nonconvex. Moreover, the region detection consistency could also be guaranteed by the one-term and multi-term SKPD models given on a coherence condition. 
The nonlinear SKPD model is closely related to a shallow CNN, particularly to a CNN with one convolutional layer and one fully-connected layer. The ``dictionaries" in nonlinear SKPD can be viewed as the filters in CNN, while the ``location indicators'' can be viewed as the coefficients in the fully-connected layer. However, different from standard CNN, the ``dictionaries'' in SKPD convolves with the input with no overlap. Such design not only enables signal region detection, it also significantly improves the interpretability of standard CNN. Finally, a comprehensive simulation study and a real MRI analysis with the UK Biobank data further validated the effectiveness of SKPDs on signal region detection problems.

The rest of the paper is organized as follows. In Section \ref{sec-2}, we introduce the one-term SKPD for matrix image and tensor image, along with the path-following algorithm to solve one-term SKPD. In section \ref{sec-3}, we study the multi-term SKPD. In Section \ref{sec-5}, we propose the nonlinear SKPD and discuss its connections with CNN.  Section \ref{sec5-2} contains tuning parameter selection. In Section \ref{sec-4}, we provide theoretical guarantees on the coefficients estimation and region detection of SKPD. We conduct simulation studies in Section \ref{sec-6} and a real brain MRI data analysis in Section \ref{sec-7}.

\noindent 
{\bf Notations: } For a vector $\bv=(v_1,...v_p)^\T$, $\|\bv\|_q=\sum_{1\le j\le p} (\|v_j\|^q)^{1/q}$ is the $\ell_q$ norm, $\|\bv\|_0$ the number of nonzero entries.  For a matrix $\bM=\{M_{i,j}, 1\le i\le n, 1\le j\le m\}$, 
$\|\bM\|_F=(\sum_{i,j}M_{i,j}^2)^{1/2}$ is the Frobenius norm, $\|\bM\|_{op}$ the operator norm (the top singular value), and $\tvec(\bM)$ the vectorization of $\bM$. 
For a tensor $\mathcal{T}=\{T_{i,j,k}, 1\le i\le n, 1\le j\le m, 1\le k\le p\}$, 
$\|\mathcal{T}\|_F=(\sum_{i,j,k}T_{i,j,k}^2)^{1/2}$ is the Frobenius norm, $\tvec(\mathcal{T})$ is the vectorization of tensor $\mathcal{T}$. 
In addition, we use $\bI_n$ to denote an identity matrix of dimension $n\times n$, $\langle\cdot,\cdot\rangle$ to denote inner product, and $\otimes$ to denote the Kronecker product. Finally, the notation $f(n,p)\asymp g(n,p)$ means that there exist constant $c_1, c_2>0$ such that $c_1 g(n,p)\le f(n,p)\le c_2 g(n,p)$.  We use $c$ to refer a generic constant that may differ from line to line.

\section{The One-term SKPD}\label{sec-2}
\subsection{The matrix image model}\label{sec2-1}
We start with the regression problem for 2D grayscale image data. The study for 2D color image or general 3D image will be deferred to Section \ref{sec2-2}. Consider the model
\bel{model}
y_i=\langle\bX_i, \bC \rangle +\eps_i, \ \ i=1,\ldots, n.
\eel
where $\by_i\in \mathbb{R}$ and $\bX_i\in\mathbb{R}^{D_1\times D_2}$ are respectively the observed continuous outcome and image data for observation $i$, $\bC\in\mathbb{R}^{D_1\times D_2}$ is the unknown coefficients matrix, $\eps_i$ are i.i.d. noises. 
To focus on image regression, other design
variables, such as age and sex, are not considered here since they can be added to the regression easily. We first propose to use an one-term Kronecker Product Decomposition (KPD) to model the coefficients matrix $\bC$:
\bel{kpd}
\bC=\bA\otimes\bB, \ \ \ \|\bA\|_F=1.
\eel
Here $\otimes$ is the Kronecker product, $\bA$ and $\bB$ are unknown matrices of dimension $p_1\times p_2$ and $d_1\times d_2$, respectively. However, we only know the dimensions of $\bC$, i.e., $(D_1, D_2)$, while the dimensions of $\bA$ and $\bB$, i.e., $(p_1,p_2)$ and $(d_1,d_2)$, are unknown.
But they certainly need to satisfy $D_1=p_1\times d_1$ and $D_2=p_2\times d_2$. 

To detect the signal regions, it is essential to assume that 
there are only a few blocks of unknown shapes in the coefficients matrix contain signal. The KPD model provides a convenient way to impose such region sparseness assumption. We assume that the matrix $\bA$ is sparse: 
\bel{sparse1}
\|\bA\|_{0} \le s
\eel
for some unknown sparsity level $s$ with $1\le s\le (p_1 p_2)$. 
We name the model (\ref{model})-(\ref{sparse1}) the one-term Sparse KPD (SKPD) model. 

We note that although this paper focuses on the linear model (\ref{model}), SKPD could be easily extended to a generalized linear model by allowing certain link function $g(\cdot)$:
$g\left(\E(y_i)\right) = \langle \bX_i, \bC \rangle$. Then the coefficients $\bC$ could still be modeled as in (\ref{kpd}) and (\ref{sparse1}).

In a one-term SKPD model,  
the small block matrix $\bB$ can be viewed as the ``dictionary'' of the original coefficient, which contains the ``shape'' and ``intensity'' information of the signal.
On the other hand, the matrix $\bA$ is the ``location indicator'' for the  dictionary. Among these $p_1\times p_2$ blocks, there are at most $s$ of them contain signal, while the others are zero. For example, $A_{j,k}\neq 0$ for some $1\le i\le p_1$, $1\le j\le p_2$ suggests that the region $\left[\left((i-1)d_1+1\right): i d_1;\  \left((j-1)d_2+1\right): j d_2\right]$ contains signal. 

We then consider the following penalized minimization problem for estimating $(\bA,\bB)$,
\begin{align}\label{obj}
	&(\hbA, \hbB)\in \argmin_{\bA, \bB} \left\{\frac{1}{2n} \sum_{i=1}^n\left(y_i-\langle\bX_i, \bA\otimes\bB \rangle\right)^2 + \lambda\|\tvec(\bA)\|_{1}\right\},\cr &
	\text{subject to} \quad \|\bA\|_F = 1,
\end{align}
where $\lam$ is a regularization parameter and will be discussed in detail later. Here we impose the $\ell_1$-norm on $\bA$ to account for its sparsity, while the Frobenius-norm on $\bA$ is imposed to guarantee that the scale of the estimation is consistent with that of the truth, although such condition would not change the estimation of $\hbC=\hbA\otimes \hbB$.

We shall mention that the decomposition (\ref{kpd}) is not identifiable for $\bA$ and $\bB$ in general. However, when the dimensions of $\bA$ and $\bB$, i.e., $p_1$, $p_2$, $d_1$ and $d_2$, are known, $\bA$ and $\bB$ are identifiable up to a sign change, i.e., $\bA\otimes\bB=(-\bA)\otimes(-\bB)$.
Suppose for now that the dimensions are known. For any matrix $\bC$ that is a $p_1\times p_2$ array of blocks of the same block size $d_1\times d_2$, let $C_{j,k}^{d_1,d_2}$ be the $(j,k)$-th block, $1\le j\le p_1$, $1\le k\le p_2$. Further let the operator $\mR: \mathbb{R}^{(p_1d_1)\times (p_2d_2)}\rightarrow \mathbb{R}^{(p_1p_2)\times (d_1d_2)}$ be a mapping from any matrix $\bC$ to 
\bel{Rmat}
\mR(\bC)=\left[\tvec(C_{1,1}^{d_1,d_2}),\ldots, \tvec(C_{1,p_2}^{d_1,d_2}),\ldots, \tvec(C_{p_1,1}^{d_1,d_2}),\ldots,\tvec(C_{p_1,p_2}^{d_1,d_2})\right]^\T.
\eel
When applying the operator $\mR$ to a Kronecker product $\bA\otimes \bB$, it holds that
\bel{Rmat2}
\mR(\bA\otimes \bB)=\tvec(\bA)[\tvec(\bB)]^\T.
\eel
The property (\ref{Rmat2}) would be of great use in our analysis. Let  $\ba=\tvec(\bA)$, $\bb=\tvec(\bB)$ and $\mR(\bX_i)=\tbX_i$. It follows from (\ref{Rmat2}) that the optimization (\ref{obj}) can be rewritten as the following bi-linear problem 
\begin{align}\label{obj2}
	&(\hba, \hbb)\in \min_{\ba, \bb} \left\{\frac{1}{2n} \sum_{i=1}^n\left(y_i- \ba^\T\tbX_i \bb \right)^2 + \lambda\|\ba\|_{1} \right\}
	\cr &
	\text{subject to} \quad \|\ba\|_2 = 1.
\end{align}

Given an appropriate initialization, the optimization problem can be solved by alternatively updating $\ba$ and $\bb$. Specifically, updating $\bb$ given $\ba$ is a standard OLS problem, while updating $\ba$ given $\bb$ reduces to a Lasso \citep{TibshiraniR96}. Consequently, $\bA$ and $\bB$ can be obtained as $\hbA=\tvec^{-1}(\hba)$ and $\hbB=\tvec^{-1}(\hbb)$, with $\tvec^{-1}(\cdot)$ denoting the inverse operation of $\tvec(\cdot)$. We defer to Section
\ref{sec2-3} to discuss the initialization and the alternating algorithm in detail.
In addition, recall that the implementation of (\ref{obj2}) depends on the known dimensions of $\bA$ and $\bB$.
We defer to Section \ref{sec-5} to discuss the dimension selection of SKPD models.


\subsection{The tensor image model}\label{sec2-2}

The image regression for matrix image can be directly extended to its tensor version.
It would allow us to address 2D colored image or general 3D image data. 
For ease of presentation, we demonstrate our analysis for three-order tensor here. Further generalizations to higher order tensors can be achieved in the same fashion.

Given two tensors ${\bf \mathcal{A}} \in \mathbb{R}^{p_1\times p_2\times p_3}$ and ${\bf \mathcal{B}} \in \mathbb{R}^{d_1\times d_2\times d_3}$, the tensor Kronecker product of ${\bf \mathcal{A}}$ and ${\bf \mathcal{B}}$, still written as ${\bf \mathcal{A}}  \otimes {\bf \mathcal{B}}$, is defined as
\bes
\bmA\otimes \bmB\in \mathbb{R}^{(p_1d_1)\times (p_2d_2)\times (p_3d_3)}, \ \ \ ({\bf \bmA}  \otimes {\bf \mathcal{B}})_{\cdot \cdot k}={\bA}_{\cdot \cdot k_1}  \otimes \ {\bB}_{\cdot \cdot k_2}, \ \  
\ees
where 
\bes
k_1=\lceil (k-1)/d_3\rceil +1, \ \ k_2=k-(k_1-1)\times d_3.
\ees
Here $\lceil x \rceil$ stands for the largest integer no greater than $x$.  
Given this definition, we consider the following tensor image model: 
\bel{model2}
y_i=\langle{\bf \mathcal{X}}_i, {\bf \mathcal{C}}\rangle +\eps_i,  \ \ {\bf \mathcal{C}}={\bf \mathcal{A}}  \otimes {\bf \mathcal{B}}, \ \ i=1,\ldots, n
\eel
Similar to the matrix image regression, we impose the sparsity assumption on tensor $\mathcal{A}$:
\bes
\|\mathcal{A}\|_0\le s.
\ees
The resultant objective function becomes
\begin{align}\label{obj5}
&(\widehat{\bmA},\widehat{\bmB})\in \argmin_{\bmA, \bmB} \left\{\frac{1}{2n} \sum_{i=1}^n\left(y_i-\langle{\bf \mathcal{X}}_i, \bmA\otimes {\bf \mathcal{B}}\rangle\right)^2 + \lambda\|\tvec(\bmA)\|_{1} \right\}
\cr &
\text{subject to} \quad \|\bmA\|_F = 1.
\end{align}
For example, we may model standard 2D colored image data with ${\bf \mathcal{A}}$ being a matrix 
and ${\bf \mathcal{B}}$ being a tensor with the third mode equal to 3, representing the three color channels. When the sizes of $\bmB$ is known, the objective function (\ref{obj5}) can be written as
\begin{align}\label{obj6}
&(\hba, \hbb_1,\hbb_2,\hbb_3)\in \min_{\ba, \bb_1,\bb_2,\bb_3} \left\{\frac{1}{2n} \sum_{i=1}^n\left(y_i- \sum_{k=1}^3\ba^\T\tbX_{i,\cdot\cdot k} \bb_k \right)^2 + \lambda\|\ba\|_{1} \right\},
\cr &
\text{subject to} \quad \|\ba\|_2 = 1.
\end{align}
where for each channel $k=1,2,3$, $\bb_k=\tvec(\bB_{\cdot\cdot k})$, and $\tbX_{i,\cdot\cdot k}=\mR(\bX_{i,\cdot\cdot k})$, with $\bX_{i,\cdot\cdot k}$ representing the $k$-th channel of the image in the $i$-th observation.
For a general tensor, we define the operator $\mbR:  \mathbb{R}^{(p_1d_1)\times (p_2d_2)\times (p_3d_3)}\rightarrow \mathbb{R}^{(p_1p_2p_3)\times (d_1d_2d_3)}$ for tensor as
\begin{align}\label{Rtensor}
\mbR(\bmC)=&\Big[\tvec(\mC_{1,1,1}^{d_1,d_2,d_3}),\ldots, \tvec(\mC_{1,1,p_3}^{d_1,d_2,d_3}),\ldots, \tvec(\mC_{1,p_2,1}^{d_1,d_2,d_3}),\ldots,\tvec(\mC_{1, p_2,p_3}^{d_1,d_2,d_3}),\ldots,
\cr &\ \ \ \tvec(\mC_{p_1,1,1}^{d_1,d_2,d_3}),\ldots, \tvec(\mC_{p_1,1,p_3}^{d_1,d_2,d_3}),\ldots, \tvec(\mC_{p_1,p_2,1}^{d_1,d_2,d_3}),\ldots,\tvec(\mC_{p_1, p_2,p_3}^{d_1,d_2,d_3})\Big]^\T,
\end{align}
where $\mC_{j,k,l}^{d_1,d_2,d_3}$ is the $(j,k,l)$-th block of $\bmC$ of dimension $d_1\times d_2\times d_3$.
This is the tensor generalization of the operator (\ref{Rmat}) and  similar properties in (\ref{Rmat2}) also holds:
\bel{Rtensor2}
\mbR(\bmA\otimes \bmB)=\tvec(\bmA)[\tvec(\bmB)]^\T.
\eel
Due to (\ref{Rtensor2}), when the dimensions $(d_1,d_2,d_3)$ are given, by letting $\ba=\tvec(\bmA)$ and $\bb=\tvec(\bmB)$, the optimization problem (\ref{obj5}) reduces to the form in (\ref{obj2}) and alternating minimization approach could still be adopted. 

The transformation (\ref{Rtensor}) allows us to analyze tensor SKPD using matrix properties. Indeed, as the operator $\mR(\cdot)$ will transform a tensor into a matrix, all the theoretical properties of matrix SKPD could be extended to the tensor version straightforwardly. Such a property is a major advantage of SKPD. It is well recognized that the tensor decomposition (both the canonical polyadic decomposition and the Tucker decomposition) are much more complicated compared to matrix decomposition. The SKPD framework allows us to avoid analyzing the complicated tensor decomposition and obtain an unified theorem (to be shown in Section \ref{sec-4}).


\subsection{A path following algorithm}\label{sec2-3}
As discussed before, the one-term SKPD model can be computed easily by alternatively updating $\hba$ and $\hbb$ when the sizes of $\bA$ (or $\bB$) are given. 
In this section, we introduce a path following algorithm to consider a sequence of 
regularization parameters $\lam^{(t)}$ of decreasing order and obtain approximate solutions of (\ref{obj2}) corresponding to the sequence of $\lam^{(t)}$. Our algorithm applies for both matrix and tensor represented image.

We start by considering the initialization. Denote $\tba^{(t)}$ and $\hba^{(t)}$ as the estimation for $\ba$ before and after normalization in the $t$-th step, i.e.,  $\hba^{(t)}=\tba^{(t)}/ \|\tba^{(t)}\|_2$. We initialize $\tba^{(0)}=\hba^{(0)}$ as the top-1 left singular vector of $\sum_{i=1}^n\tbX_iy_i$ with $\tbX_i=\mathcal{R}(\bX_i)$ for matrix image or $\tbX_i=\mbR(\bX_i)$ for tensor image. In Section \ref{sec2-3} we will show that such initialization is close to the truth with a desired precision. Given the normalized $\hba^{(t-1)}$, $t=1,2, \ldots$, we update $\hbb^{(t)}$ by
\bel{ols}
\hbb^{(t)} \leftarrow \min\limits_{\bb} {\frac{1}{2n}\sum_{i=1}^{n}(y_i -  (\hba^{(t-1)})^\T\tbX_i\bb)^2 }.
\eel
It is clear that (\ref{ols}) is a standard OLS problem and can be solved easily. Given $\hbb^{(t)}$, we in the $t$-th step consider the regularization parameter of the following form:
\bel{lam}
\lam^{(t)}=\lam^{(0)}\kappa^t \|\hbb^{(t)}\|_2,\  \ \ \kappa\in (0,1),   \ \ \ t=1,2,\cdots.
\eel
Here $\lam^{(0)}$ is an initialization constant. By (\ref{lam}), we have $\lam^{(t+1)}/\lam^{(t)}=\kappa \|\hbb^{(t+1)}\|_2/\|\hbb^{(t)}\|_2$. When $\|\hbb^{(t+1)}\|_2\approx \|\hbb^{(t)}\|_2$, $\lam^{(t+1)}/\lam^{(t)}\approx \kappa <1$, so that $\lam^{(t)}$ is in a decreasing order.  Given $\hbb^{(t)}$ and $\lam^{(t)}$, we update $\hba^{(t)}$ and $\tba^{(t)}$ by
\bel{lasso}
\tba^{(t)} &\leftarrow &\min\limits_{\ba}{\frac{1}{2n}\sum_{i=1}^{n}(y_i - (\hbb^{(t)})^\T \tbX_i^\T \ba)^2 + \lambda^{(t)}\left\|\ba\right\|_1},
\cr \hba^{(t)}&\leftarrow &\tba^{(t)}/ \|\tba^{(t)}\|_2.
\eel
The Lasso problem (\ref{lasso}) can be solved by standard approach, e.g., coordinate gradient descent. We denote $\lam^{tgt}$ as the target regularization parameter in (\ref{obj2}), the total number of iterations is
\bes
T=\left\lceil\frac{\log \lam^{tgt}}{\log \kappa}\right\rceil + T_0,
\ees
where $T_0$ is certain positive integer related to $\lam^{(0)}$ and $\|\bb^{(t)}\|_2$. In practice, the target regularization parameter $\lam^{tgt}$ can be chosen by cross-validation or modified BIC \citep{wang2009shrinkage}. We defer to Section \ref{sec5-2} for more details on tunning parameter selection. The path-following algorithm is summarized in Algorithm \ref{alg:1} below. 
\begin{algorithm}[H]
\renewcommand{\algorithmicrequire}{\textbf{Input:}}
\renewcommand{\algorithmicensure}{\textbf{Output:}}
\caption{Alternating Minimization for the one-term SKPD}
\label{alg:1}
\begin{algorithmic}[1]
	\REQUIRE $\by_i$ and $\bX_i$, $i=1,\ldots, n$
	\STATE Initialization: $\hba^{0}$ is taken as the top-1 left singular vectors of $\sum_{i}\tbX_i y_i$ with $\tbX_i=\mathcal{R}(\bX_i)$ for matrix image or $\tbX_i=\mbR(\bX_i)$ for tensor image, and $\lam^{(t)}$, $t=1,\ldots, T$.
	\FOR{t in $0,1,2,...T-1$}
	\STATE {$\hbb^{(t+1)} \leftarrow \min\limits_{\bb} {\frac{1}{2n}\sum_{i=1}^{n}(y_i -  \tba^{(t)}\tbX_i\bb)^2 }$}
	\STATE {$\tba^{(t+1)} \leftarrow \min\limits_{\ba}{\frac{1}{2n}\sum_{i=1}^{n}(y_i - (\hbb^{(t+1)})^\T \tbX_i^\T \ba)^2 + \lambda^{(t+1)}\left\|\ba\right\|_1}$}
	\STATE {Normalization:  $\hba^{t+1} \leftarrow \tba^{t+1}/ \|\tba^{t+1}\|_2$}
	\ENDFOR
	\STATE \textbf{return} $\hbA^{(T)}=\tvec^{-1}\left(\hba^{(T)}\right)$, $\hbB^{(T)}=\tvec^{-1}\left(\hbb^{(T)}\right)$
\end{algorithmic}  
\end{algorithm}
\vspace{-0.2in}

\section{The Multi-term SKPD} \label{sec-3}
\medskip
Under many scenarios, the coefficients structure could be much more complex and difficult to be modeled by a one-term SKPD. For example, when there are many different signal regions of different shapes, we may need multiple different dictionaries and associated location indicators to approximate the coefficients.
Therefore, we generalize the one-term SKPD to the following multi-term model,
\bel{rkpd}
\bC=\sum_{r=1}^R \bA_r\otimes \bB_r.
\eel
Here $R$ is the number of terms in the multi-term SKPD. The $\bB_r$ can be viewed as different dictionaries for different shapes of the signal, and $\bA_r$ are the location indicators for these dictionaries. Similar to the one-term SKPD, we impose sparsity assumption on $\bA_r$ to achieve signal region detection, i.e., we suppose
\bel{sparse2}
\|\bA_r\|_{0} \le s_r, \ \ r=1,\ldots, R,
\eel
for some possibly different $s_1,\ldots,s_R$. That is to say, different sparsity levels are allowed for the matrices $\bA_r$. 

The decomposition (\ref{rkpd}) is general. Indeed, by \cite{van1993approximation}: for any matrix $\bC\in\mathbb{R}^{D_1\times D_2}$, and any given $d_1$ and $d_2$ (that can be divided by $D_1$ and $D_2$ respectively), there is a Kronecker Product Decomposition (KPD):
$\bC = \sum_{r = 1}^{R} \bA_r \otimes \bB_r$, where $\bB_r \in \mathbb{R}^{d_1\times d_2}$, $\bA_r \in \mathbb{R}^{p_1\times p_2}$ with $\|\bA_r\|_F=1$ and $(p_1,p_2)=(D_1/d_1, D_2/p_2)$, and $R = \min\{p_1p_2, d_1d_2\}$. That to say, for any given $(d_1,d_2)$, there is a corresponding  $R$ that satisfies the decomposition (\ref{rkpd}). Moreover, this paper is motivated to detect sparse and small signal regions in medical imaging data.  
Under such scenario, a rank $R$ that is much smaller than the theoretical upper bound ($\min\{p_1p_2, d_1d_2\}$) could be sufficient for the Kronecker decomposition.
In the following figure, we demonstrate such a decomposition under different signal shapes and grid sizes. 
\begin{figure}[H]
\centering
\includegraphics[width=0.7\columnwidth]{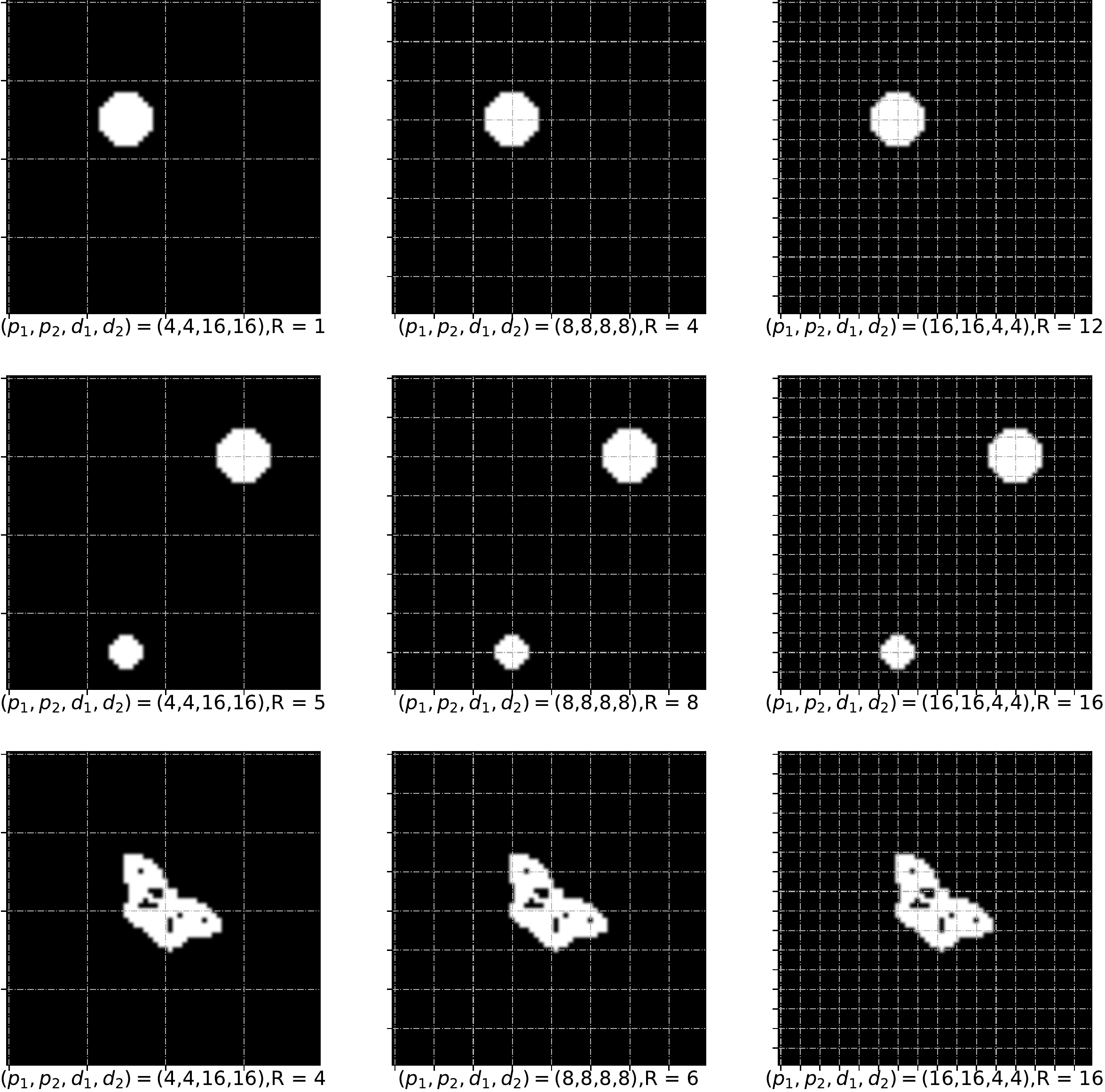}
\caption{The number of terms needed under different signal shapes and grid sizes. Note that each of the plot demonstrates an obvious but not unique decomposition. It is possible to find other KPD such that fewer number of terms are needed.}
\label{fig:study3-revision2}
\end{figure}



The decomposition (\ref{rkpd}) is not unique even if the sizes of $\bB_1,\ldots,\bB_R$ (or $\bA_1,\ldots,\bA_R$) are equal and correctly specified. Therefore, to make $\bA_r$ and $\bB_r$ identifiable, we assume the following orthonormal condition on $\bA_r$: 
\bel{orth}
\tr(\bA_r^\T\bA_l)=\left\{\begin{matrix}
1, \quad \text{if } r= l,
\\0, \quad  \text{if } r\neq l.
\end{matrix}\right. 
\eel
and decreasing norm condition on $\bB_r$:
\bel{Bnorm}
\|\bB_1\|_F\ge\|\bB_2\|_F\ge\cdots\ge\|\bB_R\|_F.
\eel
In general, it is impossible to impose the orthogonal assumption on both $\bA_r$ and $\bB_r$ when $\bA_r$ are assumed to be sparse. This can be seen when we apply the transformation $\mR(\cdot)$ in (\ref{Rmat}) on $\bC$ in (\ref{rkpd}). Suppose the sizes of $\bB_r$ are equal and known, then after $\mR(\cdot)$ transformation, we have
\bel{rmat3}
\mR(\bC)=\sum_{r=1}^R\ba_r\bb_r^\T,
\eel
where $\ba_r=\tvec(\bA_r)$ and  $\bb_r=\tvec(\bB_r)$. If the orthogonal assumption on both $\ba_r$ and $\bb_r$  are imposed, then (\ref{rmat3}) can be viewed as a singular value decomposition (SVD) on $\mR(\bC)$ with singular values incorporated into the singular vectors $\bb_r$. However, the sparsity assumption on $\bA_r$ may conflict with the SVD as the singular vectors are usually not sparse. Therefore, the orthogonal assumption could only be imposed on either $\bA_r$ or $\bB_r$. We impose such assumption on $\bA_r$ from a heuristic perspective. Considering the case that the true coefficients contains multiple non-overlap signal regions of different shapes. An ideal decomposition of (\ref{rkpd}) would be $\bA_r$ representing the non-overlap regions and $\bB_r$ representing signal shapes and intensities.
Through this decomposition, $\bA_r$ are naturally orthogonal to each other. As a consequence, the assumptions (\ref{orth}) and (\ref{Bnorm}) are imposed for identifiability.

To solve the multi-term SKPD model, we propose the following optimization problem
\begin{align}\label{obj3}
&(\hbA_1,\ldots\hbA_R, \hbB_1,\ldots, \hbB_R)
\cr \in & \argmin_{\substack{\bA_1,\ldots,\bA_R\\ \bB_1,\ldots,\bB_R}}  \left\{\frac{1}{2n} \sum_{i=1}^n\left(y_i-\langle\bX_i, \sum_{r=1}^R \bA_r\otimes \bB_r \rangle\right)^2 + \lambda\sum_{r=1}^R\|\tvec(\bA_r)\|_{1} \right\}.
\cr & \text{suject to }\quad \|\bA_r\|_F = 1, \ \ \tr(\bA_r^\T\bA_l)=0, \ \ r\neq l, \ \ 1\leq r,l \leq R.
\end{align}
When the sizes of $\bB_r$ are known, applying (\ref{rmat3}) gives us
\begin{align}\label{obj4}
&(\hba_1,\ldots\hba_R, \hbb_1,\ldots, \hbb_R)\in  \argmin_{\substack{\ba_1,\ldots,\ba_R\\ \bb_1,\ldots,\bb_R}}  \left\{\frac{1}{2n} \sum_{i=1}^n\left(y_i- \sum_{r=1}^R \ba_r^\T\tbX_i \bb_r\right)^2 + \lambda\sum_{r=1}^R\|\ba_r\|_{1} \right\},
\cr & \text{suject to }\quad \|\ba_r\|_2 = 1, \ \ \ba_r^\T\ba_l=0, \ \ r\neq l, \ \ 1\leq r,l \leq R.
\end{align}
We shall note that the decreasing norm condition (\ref{Bnorm}) is not imposed in the optimization (\ref{obj3}) or (\ref{obj4}) as it has no effect on the estimation of $\hbC=\sum_{r=1}^R\hbA_r\otimes \hbB_r$.

The resultant optimization problem (\ref{obj4}) could still be solved by alternatively updating $(\ba_1,\ldots,\ba_R)$ and $(\bb_1,\ldots,\bb_R)$. Let $\bbA=[\ba_1,\ba_2,\ldots,\ba_R]\in\mathbb{R}^{(p_1p_2)\times R}$ and $\bbB=[\bb_1,\bb_2,\ldots,\bb_R]\in\mathbb{R}^{(d_1d_2)\times R}$ be the combined matrices of $\bA_r$ and $\bB_r$ across $r$-terms respectively, and $\hbbA^{(t)}$ and $\hbbB^{(t)}$ be the corresponding estimations at stage $t$. 
We initialize $\hbbA^{(0)}$ as the top-R left singular vectors of $\sum_{i=1}^n\tbX_iy_i$ with $\tbX_i=\mathcal{R}(\bX_i)$. Given $\hbbA^{(t-1)}$, we update $\hbbB^{(t)}$ by
\bel{ols2}
\hbbB^{(t)} \leftarrow \min\limits_{\bbB} {\frac{1}{2n}\sum_{i=1}^{n}\left(y_i - \left[\tvec(\bbB)\right]^\T \tvec\big(\tbX_i^\T\hbbA^{(t-1)}\big)\right)^2 }.
\eel
It is still an OLS problem and can be solved easily when $n>Rd_1d_2$.  Given $\hbb^{(t)}$, we consider the regularization parameter of similar form to the one-term case
\bel{lam2}
\lam^{(t)}=\lam^{(0)}\kappa^t \|\hbbB^{(t)}\|_F,  \ \ \ t=1,2,\cdots,\  \ \ \kappa\in (0,1),
\eel
and update $\hbbA^{(t)}$ by
\begin{align}
\tbbA^{(t)} &\leftarrow \min\limits_{\bbA} {\frac{1}{2n}\sum_{i=1}^{n}\left(y_i - \left[\tvec(\bbA)\right]^\T \tvec\big(\tbX_i\hbbB^{(t-1)}\big)\right)^2 }+ \lambda^{(t)}\left\|\tvec(\bbA)\right\|_1\label{Rlasso1},
\\ \hbbA^{(t)}&\leftarrow \tbbA^{(t)} \left[(\tbbA^{(t)})^\T\tbbA^{(t)}  \right]^{-1/2}.\label{Rlasso2}
\end{align}
The update (\ref{Rlasso2}) is to guarantee that $\hbbA^{(t)}$ is an orthonormal matrix, i.e., $(\hbbA^{(t)})^\T\hbbA^{(t)}=\bI_R$, and match the orthogonality assumption (\ref{orth}) on the true coefficients $\bA_r$. We shall note that the orthonomalization step allow us to find the nearest orthonormal matrix to $\tbbA^{(t)} $. However, it  is only for the identifiability consideration and does not change the estimation of $\hbC^{(t)}=\hbbA^{(t)}(\hbbB^{(t)})^\T$. 
When $\hbbA^{(t)}$ is not column-wise full-rank, the orthonormalization step (\ref{Rlasso2}) can be modified slightly to $\hbbA^{(t)}\leftarrow \tbbA^{(t)} \left[(\tbbA^{(t)})^\T\tbbA^{(t)} +\eta\bI_R \right]^{-1/2}$, where $\eta$ is a small constant, such as $1/n$. We summarize the path following algorithm for R-term SKPD below.

\begin{algorithm}[H]
\renewcommand{\algorithmicrequire}{\textbf{Input:}}
\renewcommand{\algorithmicensure}{\textbf{Output:}}
\caption{Alternating Minimization for the R-term SKPD}
\label{alg:2}
\begin{algorithmic}[1]
	\REQUIRE $\by_i$ and $\bX_i$, $i=1,\ldots, n$
	\STATE Initialization: $\hbbA^{0}$ is taken as the top-R left singular vectors of $\sum_{i}\tbX_i y_i$ with $\tbX_i=\mathcal{R}(\bX_i)$ for matrix image or $\tbX_i=\mbR(\bX_i)$ for tensor image, and $\lam^{(t)}$, $t=1,\ldots, T$.
	\FOR{t in $0,1,2,...T-1$}
	\STATE {$\hbbB^{(t+1)} \leftarrow \min\limits_{\bbB} {\frac{1}{2n}\sum_{i=1}^{n}\left(y_i - \left[\tvec(\bbB)\right]^\T \tvec\big(\tbX_i^\T\hbbA^{(t)}\big)\right)^2 }$}
	\STATE {$\tbbA^{(t+1)} \leftarrow \min\limits_{\bbA} {\frac{1}{2n}\sum_{i=1}^{n}\left(y_i - \left[\tvec(\bbA)\right]^\T \tvec\big(\tbX_i\hbbB^{(t+1)}\big)\right)^2 }+ \lambda^{(t+1)}\left\|\tvec(\bbA)\right\|_1$}
	\STATE {Orthonormalization:  $ \hbbA^{(t+1)}\leftarrow \tbbA^{(t+1)} \left[(\tbbA^{(t+1)})^\T\tbbA^{(t+1)}  \right]^{-1/2}$}
	\ENDFOR
	\STATE \textbf{return} $\hbbA^{(T)}$, $\hbbB^{(T)}$ and $\hbC^{(T)}=\hbbA^{(T)}(\hbbB^{(T)})^\T$
\end{algorithmic}  
\end{algorithm}

\section{The Nonlinear SKPD and its connections to CNN} \label{sec-5}
The proposed approaches share many similarities with a shallow convolutional neural network (CNN). Consider the problem of predicting a scalar outcome with a $D_1\times D_2$ image using a simple CNN with one convolutional layer and one fully-connected layer. Suppose that there are $R$ unknown filters and each of dimension $d_1\times d_2$ in the convolutional layer. Each of these $R$ filters convolves with the input features $\bX$ with stride $(s_1, s_2)$ on two dimensions to obtain a single feature map of dimension 
$\lceil(D_1-d_1+1) /s_1\rceil\times\lceil(D_2-d_2+1) /s_2\rceil$.
When $s_1<d_1$ and  $s_2<d_2$, the filters convolves with $\bX$ with overlap, otherwise with no overlap.
The outputs of convolutional layers are then followed by nonlinear activation functions, such as ReLU(Rectified Linear Unit). Finally in the fully connected layer, the $R$ activated feature maps are used to predict the final output. In this process, the unknown parameters are 1) R filters, each of dimension $d_2\times d_2$, and 2) $R$ matrices for the activated feature map in the fully connected layer, each of dimension $\lceil(D_1-d_1+1) /s_1\rceil\times\lceil(D_2-d_2+1) /s_2\rceil$.

In an $R$-term SKPD model, the block matrices $\bB_r$ can be viewed as the unknown filters in CNN, while $\bA_r$ can be understood as the matrices for the feature map in the fully connected layer. 
Then, the filters in our approach convolves with the input feature $\bX$ with the stride size $s_1=d_1$ and  $s_2=d_2$. So such convolutions are non-overlapped. Consequently, the resulted feature map for each filter is of dimension $(D_1/d_1)\times (D_2/d_2)$ when $D_1$ and $D_2$ are multiplications of $d_1$ and $d_2$. 
More rigorously, define the non-overlapped convolution operator $*$ for matrix $\bX \in \mathbb{R}^{D_1\times D_2}, \bB \in \mathbb{R}^{d_1\times d_2}$ as
\bel{conv-def1}
\bX * \bB \in \mathbb{R}^{p_1\times p_2}, \quad p_1 = D_1/d_1, \quad p_2 = D_2/d_2
\eel
with the $(j,k)$-th component being
\bel{conv-def2}
(\bX * \bB)_{j,k} = \langle \bX_{j,k}^{d_1,d_2}, \bB \rangle, \quad 1\leq j \leq p_1, 1 \leq k \leq p_2.
\eel
Here $\bX_{j,k}^{d_1,d_2}$ is the $(j,k)$-th block of $\bX$ and is of dimension $d_1\times d_2$.
Building on this convolution operator, the one-term SKPD model (\ref{model}) and (\ref{kpd}) can be rewritten as 
\bel{conv-one-term}
y_i = \langle \bA , \ \bX_i * \bB \rangle + \eps_i.
\eel 
Similarly, the $R$-term SKPD model (\ref{model}) and (\ref{rkpd}) can be rewritten as 
\bel{conv-r-term}
y_i = \sum_{r = 1}^{R} \langle \bA_r , \ \bX_i * \bB_r \rangle + \eps_i.
\eel 
By writing SKPD into the forms of (\ref{conv-one-term}) and  (\ref{conv-r-term}), it is clear that the SKPD is equivalent to a two-layers CNN with one convolutional layer, one fully-connected layer and an identity activation function. 

The identity activation function in (\ref{conv-r-term}) could be extended to a general nonlinear activation function $g(\cdot)$. This leads to the following nonlinear SKPD:
\bel{conv-nonlinear}
y_i = \sum_{r = 1}^{R} \langle \bA_r , \ g(\bX_i * \bB_r) \rangle + \eps_i
\eel 
where $g(\bv)=[g(v_1), g(v_2),\ldots, g(v_p)]^\T$ for any $\bv\in\mathbb{R}^p$. Popular choices of $g(\cdot)$ include ReLU, $g(v)=\sigma(v)=(v+|v|)/2$ and Sigmoid, $g(v)=(1+e^{-v})^{-1}$. The optimization problem resulted from (\ref{conv-nonlinear}) becomes
\begin{align}\label{obj4a}
&(\hbA_1,\ldots\hbA_R, \hbB_1,\ldots, \hbB_R)\cr \in  &\argmin_{\substack{\bA_1,\ldots,\bA_R\\ \bB_1,\ldots,\bB_R}}  \left\{\frac{1}{2n} \sum_{i=1}^n\left(y_i- \sum_{r=1}^R\langle \bA_r, g(\bX_i *  \bB_r)\rangle\right)^2 + \lambda\sum_{r=1}^R\|\tvec(\bA_r)\|_{1}\right\}, \cr
& s.t. \ \ \|\bA_r\|_F = 1 \quad r = 1 \ldots R.
\end{align}
We shall note that the orthogonality condition is not imposed in (\ref{obj4a}) due to the appearance of nonlinear activation, which enables us to avoid the identifiability issue that concerns linear SKPD. 
Clearly, when $g(v)=v$ and orthogonality condition imposed, the problem (\ref{obj4a}) reduces to (\ref{obj3}). 
We omit the discussion of computing (\ref{obj4a})
as it can be solved easily with standard CNN implementation tools, such as {\it Pytorch}.
For nonlinear SKPD, the coefficient matrix $\bC$ cannot be written as the Kronecker product form $\bC=\sum_{r=1}^R \bA_r\otimes \bB_r$. However, the nonlinear SKPD could able be used for region detection as the locations of the non-zero coefficients in  $\bC=\sum_{r=1}^R \bA_r\otimes \bB_r$ still match the signal regions. Specifically, if for some $j,k$ that $C_{j,k}=0$, then the $(j,k)$-th coefficient of $\bX_i$ would be independent with $y_i$ for $i=1,\ldots, n$. While if for some $j,k$ such that $C_{j,k}\neq 0$, the  $(j,k)$-th coefficient of $\bX_i$ may affect the outcome $y_i$ even after the nonlinear activation.

The non-overlapping design of SKPD not only significantly reduces the parameter dimension, it is also the key to achieve region detection. When the filters convolve with input features with overlaps as in CNN, the true signals are contained in multiple feature blocks. Consequently, the signal regions are difficult to be identified.
In a deep CNN, the overlapped features in many layers entangled 
together and thousands of parameters presented like a ``black box'', region detection becomes an even more difficult task.
Indeed, substantial effort has been made in the computer vision literature to improve the interpretability of deep CNN. For example, 
\cite{zeiler2014visualizing} proposed a multi-layered Deconvolutional Network to project the feature activations back to the input pixel space. \cite{zhou2016learning} proposed to learn a weight matrix to locate the class-discriminative regions in each
image. Similar strategy has also been adopted by \cite{selvaraju2017grad} and \citet{ramaswamy2020ablation}.

Even with these efforts made in visualizing CNN, we emphasize that these deep learning models still may not be good options for our task --- brain region detection. Deep CNNs are believed to work well for many computer vision tasks because they exploit hierarchies of visual features: the earlier layers usually aim to learn small pattern such as edges, while later layers put the learned small patterns together into larger patterns.
However, brain region detection is significantly different from those computer vision tasks in many different perspectives. To list a few, 1) the signal regions are much smaller, 2) the signal intensities are much weaker, 3) signal regions usually have no clear edge or boundary, 4) sample size are much smaller. 
Due to these differences, those hierarchies exploited in deep models may not benefit our region detection problem.
In fact, it is well recognized that small object detection is an extremely challenging problem for deep models \citep{liu2021survey}, not mentioning the much weaker signals and far less samples in our problem.
As a comparison, the SKPD that could improve model interpretability, reduce model dimension, and enjoy theoretical guarantees (to be shown in Section \ref{sec-4} below) is clearly a better option.


\section{Tuning parameters selection in SKPD}\label{sec5-2}
The unknown parameters involved in SKPD include: the block sizes, i.e., $(d_1, d_2)$ (or equivalently $(p_1,p_2)$, the number of blocks), the ranks $R$ and the regularization level $\lambda$. To select the unknown parameters, our strategy is to fix the block size to be ``moderately small'' and tune the rank $R$ and regularization $\lam$ for the given block size due to the following reasons.

First, as discussed in Section \ref{sec-3}, due to \citet{van1993approximation}: for any given $(d_1,d_2)$, there is a corresponding  $R\le \min\{p_1p_2,d_1d_2\}$ that satisfies the Kronecker product decomposition. Moreover, note that this paper mainly concerns detecting small and sparse signal regions. Under such scenario, the number of theoretical ranks $R$ could be much smaller than $\min\{p_1p_2,d_1d_2\}$, as illustrated in Fig. \ref{fig:study3-revision2}. Second, unlike many computer vision tasks that have a strong signal, the signals in medical imaging are often weak and the signal regions usually have no clear boundaries/edges. Consequently, 
it would be difficult or even impossible to 
perfectly
detect these regions and capture their pixel-wise shapes. 
In this sense, we do not intend to carefully tune the grid sizes as long as they fall in an appropriate range, i.e., ``moderately small'', especially considering that the rank $R$ could be adjusted for the  block size.

To tune rank $R$ and penalization strength $\lambda$, we propose to 
minimize the following modified BIC \citep{wang2009shrinkage} criteria
\bel{MBIC}
\text{BIC}(\lambda, R)=&&\log\left(\frac{1}{n}\sum_{i=1}^{n}\left(y_i - \left(\tvec\big[\hbbA(\lam,R)\big]\right)^\T \tvec\big[\tbX_i\hbbB(\lam,R)\big]\right)^2\right) \cr &&+ \frac{C_n\log(n)}{n}\times \left\|\tvec\big[\hbbA(\lam,R)\big]\right\|_0, 
\eel
where $C_n$ is certain constant that need to be specified. 
\citet{wang2009shrinkage} suggested that $C_n$ could be chosen as $\log\log(p)$, where $p$ is the number of parameters in a high-dimensional regression problem. In our case, we follow their suggestion and take $C_n=\log\log(Rp_1p_2)$. We refer to \citet{wang2009shrinkage} for more details on modified BIC.

To better illustrate the effects of $R$ in SKPD, in the supplementary material \ref{appendix:sumulation}, we conducted a simulation study to demonstrate region detection and coefficients estimation performance under different $R$. The separate terms estimated by R-term SKPD are also recorded. We find that with sufficient samples, the coefficients estimation performance could be significantly improved with an enlarged $R$. But in terms of region detection, an 1-term SKPD under many cases is already sufficient.
In practice, we usually suggest to  implement the 1-term SKPD first before carefully tuning for $R$, especially when the sample size $n$ is limited.  We refer to the supplementary material \ref{appendix:sumulation} for more details.

On the other hand, we note that if there is prior knowledge about the size of true signal, the same scale of block size would be preferred. This would allow us to have a Kronecker product decomposition with even smaller $R$. 
In all our simulation and real image analysis in Section \ref{sec-6} and \ref{sec-7}, we fix the size of $\bB$ to be $d_1=d_2=8$ for matrix images and $d_1=d_2=d_3=8$ for tensor images. The SKPD performs consistently well with such grid sizes.


To help interested readers implement SKPD, we developed a Python package named ``SKPD'', available at \href{https://pypi.org/project/SKPD/}{https://pypi.org/project/SKPD} with specified PyEnv. 
In addition, more examples and source code can be found at Github: \href{https://github.com/SanyouWu/SKPD}{https://github.com/SanyouWu/SKPD}. 

\section{Theoretical Results}\label{sec-4}
In this section, we present our main theoretical results for the linear SKPDs. Specifically, 
we first prove that the path following algorithm described in Section \ref{sec2-3} and Section \ref{sec-3} converge to the truth even though the optimization is nonconvex. We then show the region detection consistency of SKPD by proving the sign consistency of $\hbA$. For ease of presentation, the results are presented for matrix images, although all the results also work for tensor images with slight change of statement.

\subsection{Estimation consistency of one-term SKPD}\label{sec3-1}
In this subsection, we provide sharp theoretical upper bounds for $\|\hbA^{(t)}-\bA\|_F$, $\|\hbB^{(t)}-\bB\|_F$ and $\|\hbA^{(t)}\otimes \hbB^{(t)}-\bA\otimes \bB\|_F$ when the dimensions of $\bB$ are correctly specified in a one-term SKPD. This is equivalent to bound $\|\hba^{(t)}-\ba\|_2$, $\|\hbb^{(t)}-\bb\|_2$ and $\|\hba^{(t)}(\hbb^{(t)})^\T-\ba\bb^\T\|_2$ as $\hba^{(t)}=\tvec(\hbA^{(t)})$, $\hbb^{(t)}=\tvec(\hbB^{(t)})$.
We shall mention that the true matrices $\bA$ and $\bB$ are subject to sign change even if their dimensions are given. In other words, if $(\bA,\bB)$ is the true coefficients matrix, then $(-\bA,-\bB)$ also satisfies $\bC=(-\bA)\otimes (-\bB)$. Therefore, we refer $\|\hbA^{(t)}-\bA\|_F$ and $\|\hbB^{(t)}-\bB\|_F$ to the smallest corresponding errors, i.e., $\|\hbA^{(t)}-\bA\|_F=\left(\|\hbA^{(t)}\|_F^2+\|\bA\|_F^2-2|\langle\hbA^{(t)},\bA\rangle|\right)^{1/2}$ and $\|\hbB^{(t)}-\bB\|_F=\left(\|\hbB^{(t)}\|_F^2+\|\bB\|_F^2-2|\langle\hbB^{(t)},\bB\rangle|\right)^{1/2}$. 

We first recall the operator $\mathcal{R}(\cdot)$ in (\ref{Rmat}) and denote $\mR(\bX_i)=\tbX_i$. Then we define the following matrices
\bes
\tbX&=&\left(\tvec(\tbX_{1}),\ldots,\tvec(\tbX_{n})\right)^\T\in\mathbb{R}^{n\times (p_1p_2d_1d_2)},
\cr \tbX_{jk} &=& \left(\tvec\big(\{\bX_i\}^{d_1,d_2}_{jk}\big),\ldots,\tvec\big(\{\bX_i\}^{d_1,d_2}_{jk}\big)\right)^\T\in\mathbb{R}^{n\times (d_1d_2)}, \ \ 1\le j\le p_1, 1\le k\le p_2.
\ees
For any $\tbX_{jk}$, we let $\ttheta_{j,k}=(1/\sqrt{n})\big\|\tbX_{jk}\big\|_{2}$ be its scaled top singular value. Further let $\ttheta=\max_{j,k}\ttheta_{j,k}$. For $\tbX$, 
We assume that it satisfies the following Restricted Isometric Property (RIP):
\begin{condition}[Restricted Isometric Property] \label{RIP}
For each integer $r=1,2,\ldots$, a matrix $\bX\in\mathbb{R}^{n \times (D'D'')} $ is said to satisfy the $r$-RIP with constant $\delta_r$, if for all for matrices $\bM \in \mathbb{R}^{D' \times D''}$ that rank at most $r$, the following holds
\begin{align}\label{rip1}
	(1-\d_r)\|\bM\|_F^2 \leq \frac{1}{n}\|\bX\tvec(\bM)\|_2^2\leq (1+\d_r)\|\bM\|_F^2.
\end{align}
\end{condition}
The RIP is first proposed by \citet{candes2005decoding} for sparse vector, and later generalized by \citet{recht2010guaranteed} for low rank matrices as in Definition \ref{RIP}. The RIP condition is satisfied by many random matrices with sufficiently large number of independent observations, such as the sub-Gaussian matrices  \citep{recht2010guaranteed}.

Now let $\bep=(\eps_1,\ldots,\eps_n)^\T$ be the additive noises and define the following related quantities
\bel{err}
\tau_1=\max_{j,k}\frac{1}{n}\left\|\tbX_{jk}^\T\bep\right\|_2,
\quad    \tau_2 = \left\|\tvec^{-1}\big(\tbX^\T\bep/n\big)\right\|_{op}.
\eel
We will provide probabilistic upper bounds for $\tau_1$ and $\tau_2$ later. Before that, we introduce the conditions on the initialization and show that the initialization described in Section \ref{sec2-3} satisfies such conditions. Let $\hba^{(0)}$ be the normalized initialization with $\|\hba^{(0)}\|_2=1$ and $\mu_0=\|\hba^{(0)}-\ba\|_2$ be the initial error.
We assume that 
\bel{kappa1}
\kappa_1= (1/2)\mu_0+ \delta_2(1-\delta_2)^{-1} <1,
\eel
and 
\bel{kappa2}
\cr \kappa_2 =\frac{6\ttheta\sqrt{s}(1+\delta_2)^{1/2}}{(1-\kappa_1\mu_0)(1-\delta_2)^{-1} -(\tau_2/\|\bb\|_2)} \in (0,1).
\eel
We argue that the initialization requirements (\ref{kappa1}) and (\ref{kappa2}) could be satisfied easily when $\tba^{(0)}$ is chosen to be the first left singular vector of $\sum_{i}\tbX_i y_i$. 
Here we provide some intuitions about this argument, while the formal statement is deferred to Theorem \ref{th-3}.  First note that $\mu_0\le \sqrt{2}$ holds for any $\hba^{(0)}$ due to normalization and the definition of $\|\hba-\ba\|_2$. Thus, the requirement (\ref{kappa1}) holds for any initialization as long as $\delta_2<0.22$, while larger $\delta_2$ is also possible for some carefully chosen initialization. For the condition (\ref{kappa2}), we shall need to understand the scale of $\ttheta$. 
To have some heuristics, consider the extreme case that each block of $\bX$, i.e., $\tbX_{jk}$, is identical across $j,k$. Then we have 
$\ttheta^2=\max_{j,k}\ttheta_{j,k}^2=(p_1p_2)^{-1}\sum_{j,k}\theta_{jk}^2\le (p_1p_2)^{-1} (1+\delta_1)$.
So we see that $\ttheta$ is of order of $(p_1p_2)^{-1/2}$. Consequently, (\ref{kappa2}) can be satisfied as long as $s\ll p_1p_2$ and $\tau_2 < (1-\kappa_1\mu_0)(1-\delta_2)\|\bb\|_2$, where the first inequality holds when $\bA$ is sparse enough, the second inequality holds when $\kappa_1\mu_0<1$ and $\tau_2$ is small enough.

Finally, we define the following quantities that will be used in our Theorem \ref{th-1} below.
\bel{nu}
\nu_1=\frac{\tau_2}{(1-\delta_2)\|\bb\|_2}, \ \  \ \ \nu_2=\frac{6\tau_1\sqrt{s}}{\|\bb\|_2(1-\kappa_1\mu_0)(1-\delta_2) -\tau_2}.
\eel
\begin{thm}[Non-asymptotic]
\label{th-1}
Suppose model (\ref{model})-(\ref{sparse1}) hold and Algorithm \ref{alg:1} is implemented under the true dimension with regularization parameters $\lam^{(t)}$, $t=1,2,\cdots$. Assume that $\tbX$ satisfies the 2-RIP condition with constant $\delta_2$. Let $\mu_0=\|\hba^{(0)}-\ba\|_2$ be the initialization error and $\ttheta=(1/\sqrt{n})\max_{j,k}\big\|\tbX_{jk}\big\|_{op}$. Let $\lam^{(t)}=2\|\hbb^{(t)}\|_2 \left\{\tau_1+\ttheta(1+\delta_2)^{1/2}\left[\kappa_1^t\kappa_2^{t-1} \mu_0 + (\kappa_1\nu_2 +\nu_1)(1-\kappa_1\kappa_2)^{-1}\right]\right\}$. Then, if $\mu_0$ satisfies (\ref{kappa1}) and (\ref{kappa2}) and $\mu_0\ge (\kappa_2\nu_1+\nu_2)(1-\kappa_1\kappa_2)^{-2}$, we have 
\begin{align}
	\|\hba^{(t)}-\ba\|_2 &\leq 
	(\kappa_1\kappa_2)^t \mu_0 + \frac{\kappa_2\nu_1+\nu_2}{1-\kappa_1\kappa_2},\label{th-1-1}\\
	\frac{\|\hbb^{(t+1)}-\bb\|_2}{\|\bb\|_2}&\le\kappa_1^{t+1}\kappa_2^{t} \ \mu_0 + \frac{\kappa_1\nu_2+\nu_1}{1-\kappa_1\kappa_2},\label{th-1-2}\\
	\frac{\|\hbC^{(t+1)}-\bC\|_F}{\|\bC\|_F} &\le (1+\kappa_2) \frac{\|\hbb^{(t+1)}-\bb\|_2}{\|\bb\|_2} + \nu_2.\label{th-1-3}
\end{align}
\end{thm}

\begin{remark}
The requirement $\mu_0\ge (\kappa_2\nu_1+\nu_2)(1-\kappa_1\kappa_2)^{-2}$ is almost negligible as if otherwise, 
we can simply take the initialization $\hba^{(0)}$ as our final estimation and it has already achieved desired estimation accuracy.
\end{remark}

\begin{remark}
We note that the seemingly different regularization parameter $\lam^{(t)}$ is asymptotically the same to that in (\ref{lam}). We refer to Theorem \ref{th-2}, the asymptotic version of Theorem \ref{th-1} for more details.
\end{remark}

Theorem \ref{th-1} provides the finite sample results of one-term SKPD with all the constants being explicit. Based only on the RIP condition and the initialization requirements, Theorem \ref{th-1} suggests that
$\hba^{(t)} $, $\hbb^{(t)}$ and $\hbC^{(t)}$ converge to their  corresponding truth geometrically even if (\ref{obj2}) is a nonconvex optimization problem. 
Moreover,  if the noise term $\eps$ is sub-Gaussian, we have the following probabilistic upper bound for $\tau_1$ and $\tau_2$ when $n\rightarrow \infty$,
\bel{asy}
\tau_1=O_p\left(\sqrt{\frac{d_1d_2\log(p_1p_2)}{n}}\right), \ \ \tau_2= O_p\left(\sqrt{\frac{\log(n)}{n}}\right).
\eel
This leads to the Theorem \ref{th-2} below. 

\begin{thm}[Asymptotic]
\label{th-2}
Suppose model (\ref{model})-(\ref{sparse1}) hold and Algorithm \ref{alg:1} is implemented under the true dimension with regularization parameters $\lam^{(t)}$, $t=1,2,\cdots$.  
Suppose $\tbX$ satisfies the 2-RIP condition with constant $\delta_2$, $\bep$ is a sub-Gaussian vector, and the initialization error $\mu_0$ satisfies (\ref{kappa1}) and (\ref{kappa2}) with $\tau_2=0$ in (\ref{kappa2}). 
Let 
\bes
\lam^{(t)}\propto\|\hbb^{(t)}\|_2 (\kappa_1\kappa_2)^t.
\ees
Then, when $n\rightarrow \infty$ and $sd_1d_2\log(p_1p_2)\ll n$, we have
after 
\bes
t\ge t_0+\frac{\log\left(n^{-1}[\log(n)+sd_1d_2\log(p_1p_2)]\right)}{2[\log(\kappa_1)+\log(\kappa_2)]}
\ees
times iteration, 
\bel{th-2-1}
\|\hba^{(t)}-\ba\|_2\asymp \frac{\|\hbb^{(t+1)}-\bb\|_2}{\|\bb\|_2}\asymp \frac{\|\hbC^{(t+1)}-\bC\|_F}{\|\bC\|_F}\asymp \sqrt{\frac{\log(n)+sd_1d_2\log(p_1p_2)}{n}},
\eel
holds with high probability,  where $t_0$ is a certain constant.
\end{thm}

The form of $\lam^{(t)}$ in Theorem \ref{th-2} match that in (\ref{lam}). It suggests that the $\kappa$ in (\ref{lam}) may be taken as $\kappa=\kappa_1\kappa_2$. Furthermore, the following Theorem \ref{th-3} shows that when the initialization $\hba^{(0)}$ is taken as the first left singular vector of $\sum_{i}\tbX_i y_i$, both $\kappa_1$ and $\kappa_2$ fall in the range $(0,1)$ and the conditions (\ref{kappa1}) and (\ref{kappa2}) are satisfied easily.
\begin{thm}[Initialization]
\label{th-3}
Suppose model (\ref{model})-(\ref{sparse1}) hold and Algorithm \ref{alg:1} is implemented under true dimension. Suppose  $s\ll p_1p_2$, $\tbX$ satisfies the 2-RIP condition with constant $\delta_2 < 0.1$, and error term $\bep$ is bounded and satisfies $\|\bep\|_2\le 0.1(1-\delta)\|\bb\|_2$. Then, when $\hba^{(0)}$ is taken as the first left singular vector of $\sum_{i}\tbX_i y_i$, we have 
\bes
\mu_0=\|\hba^{(0)}-\ba\|_2<1.
\ees
As a consequence, (\ref{kappa1}) holds with
\bes
\kappa_1\in(0,0.61).
\ees	
If in addition $n\rightarrow \infty$ and $\ttheta\sqrt{s}<1/20$, we have (\ref{kappa2}) holds with
\bes
\kappa_2\in (0,0.70).
\ees
\end{thm}

\subsection{Estimation consistency of multi-term SKPD}\label{sec3-2}
In this subsection, we generalize the theoretical results for one-term SKPD to the multi-term version. Recall the notation $\bbA=[\ba_1,\ba_2,\ldots,\ba_R]\in\mathbb{R}^{(p_1p_2)\times R}$ and $\bbB=[\bb_1,\bb_2,\ldots,\bb_R]\in\mathbb{R}^{(d_1d_2)\times R}$.
By definition, 
$\sum_{r=1}^R\|\hbA^{(t)}_r-\bA_r\|_F^2=\|\hbbA^{(t)}-\bbA\|_F^2$, $\sum_{r=1}^R\|\hbB^{(t)}_r-\bB_r\|_F^2=\|\hbbB^{(t)}-\bbB\|_F^2$, and $\left\|\sum_{r=1}^R\hbA^{(t)}_r\otimes\hbB^{(t)}_r-\sum_{r=1}^R\bA_r\otimes\bB_r\right\|_F=\|\hbbA^{(t)}(\hbbB^{(t)})^\T-\bbA\bbB^\T\|_F$. 
Our target is to bound the three quantities above. 
Similar to the one-term case, we refer $\|\hbbA^{(t)}-\bbA\|_F$ and $\|\hbbB^{(t)}-\bbB\|_F$ to the smallest corresponding errors.


We first denote the initial error as $\mu_0=\|\hbbA^{(0)}-\bbA\|_F$, the total sparsity level as $s=\sum_{r=1}^R s_r$, and recall the error terms $\tau_1$ and $\tau_2$ in (\ref{err}).
Suppose the $2R$-RIP Condition holds with constant $\delta_{2R}$. 
We now generalize the initialization requirements (\ref{kappa1}) and (\ref{kappa2}) for $R$-term SKPD:
\bel{kappa3}
\kappa_1(R)=\frac{\mu_0}{2} + \frac{\d_{2R}}{1-\d_{2R}}<1,
\eel
and
\bel{kappa4}
\kappa_2(R)=\frac{6\sqrt{R}\ttheta\sqrt{s}(1+\delta_{2R})^{1/2}}{[1-\kappa_1(R)\mu_0](1-\delta_{2R})^{-1} -\tau_2/\|\bbB\|_2}\in (0,1).
\eel
Similarly, we define the $\nu_1(R)$ and $\nu_2(R)$ as generalizations of $\nu_1$ and $\nu_2$ in (\ref{nu}):
\bel{nu2}
\nu_1(R)=\frac{\tau_2}{(1- \d_{2R})\|\bbB\|_F}, \quad \nu_2(R)=\frac{6\tau_1\sqrt{s}}{\|\bbB\|_F(1-\kappa\mu_0)(1-\delta)^{-1} -\tau_2}.
\eel
For the convenience of notation, we omit $\cdot(R)$ in $\kappa_1(R)$, $\kappa_2(R)$, $\nu_1(R)$ and  $\nu_2(R)$ and write them respectively as $\kappa_1$, $\kappa_2$, $\nu_1$ and $\nu_2$ below. 
\begin{thm}[Non-asymptotic]
\label{th-4}
Suppose the model (\ref{model}) and (\ref{rkpd})-(\ref{orth}) hold. 
Suppose the number of terms $R$ is correctly specified and Algorithm \ref{alg:2} is implemented under the true dimension. Let $\tau_1$ and $\tau_2$ be as in (\ref{err}), $\nu_1$ and $\nu_2$ be as in (\ref{nu2}) and $s=\sum_{r=1}^R s_r$. Suppose $\tbX$ satisfies the $2R$-RIP condition with constant $\delta_{2R}$ and initial error $\mu_0$ satisfies (\ref{kappa3}) and (\ref{kappa4}).
Let $\lam^{(t)}=2\sqrt{\sum_{r=1}^R\|\bB\|_F^2} \left\{\tau_1+\sqrt{R}\ \ttheta(1+\delta_2)^{1/2}\left[\kappa_1^t\kappa_2^{t-1} \mu_0 + (\kappa_1\nu_2 +\nu_1)(1-\kappa_1\kappa_2)^{-1}\right]\right\}$. Then, if $\mu_0\ge (\kappa_2\nu_1+\nu_2)(1-\kappa_1\kappa_2)^{-2}$, we have 
\begin{align}
	\sqrt{\sum_{r=1}^R\|\hbA^{(t)}-\bA|_F^2} &\leq 
	(\kappa_1\kappa_2)^t \mu_0 + \frac{\kappa_2\nu_1+\nu_2}{1-\kappa_1\kappa_2},\label{th-4-1}\\
	\sqrt{\frac{\sum_{r=1}^R\|\hbB^{(t)}-\bB|_F^2}{\sum_{r=1}^R\|\bB\|_F^2}}&\le\kappa_1^{t+1}\kappa_2^{t} \ \mu_0 + \frac{\kappa_1\nu_2+\nu_1}{1-\kappa_1\kappa_2}.\label{th-4-2}\\
	\frac{\left\|\sum_{r=1}^R\hbA^{(t)}\otimes\hbB^{(t)}-\bC\right\|_F}{\left\|\bC\right\|_F} &\le (\sqrt{R}+\kappa_2) 	\sqrt{\frac{\sum_{r=1}^R\|\hbB^{(t)}-\bB|_F^2}{\sum_{r=1}^R\|\bB\|_F^2}} + \nu_2.\label{th-4-3}
\end{align}
\end{thm}
Theorem \ref{th-4} is a direct generalization of Theorem \ref{th-1} to the R-term case. We note that all the constants in Theorem \ref{th-4} are explicit. By applying the probabilistic upper bound (\ref{asy}) on $\tau_1$ and $\tau_2$, we have the Corollary \ref{cor-1} below. 
\begin{cor}[Asymptotic]
\label{cor-1}
Suppose model (\ref{model}) and (\ref{rkpd})-(\ref{orth}) hold. Suppose the number of terms $R$ is correctly specified and Algorithm 2 is implemented under the true dimension. Let $\tau_1$ and $\tau_2$ be as in (\ref{err}), $\nu_1$ and $\nu_2$ be as in (\ref{nu2}). Suppose $\tbX$ satisfies the $2R$-RIP condition, $\bep$ is a sub-Gaussian vector, and initialization error $\mu_0$ satisfies (\ref{kappa3}) and (\ref{kappa4}) with $\tau_2=0$ in (\ref{kappa4}).
Let 
\bes
\lam^{(t)}\propto \sqrt{R} \|\hbb^{(t)}\|_2 (\kappa_1\kappa_2)^t.
\ees
Then, when $n\rightarrow \infty$ and $sd_1d_2\log(p_1p_2)\ll n$, we have
after 
\bes
t\ge t_0+ \frac{\log\left(n^{-1}[\log(n)+sd_1d_2\log(p_1p_2)]\right)}{2[\log(\kappa_1)+\log(\kappa_2)]}
\ees
times iteration,
\bel{cor-1-1}
\sqrt{\sum_{r=1}^R\|\hbA^{(t)}-\bA|_F^2}\asymp 	\sqrt{\frac{\sum_{r=1}^R\|\hbB^{(t)}-\bB|_F^2}{\sum_{r=1}^R\|\bB\|_F^2}}\asymp \sqrt{\frac{\log(n)+sd_1d_2\log(p_1p_2)}{n}}
\eel
and
\bel{cor-1-2}
\frac{\left\|\sum_{r=1}^R\hbA^{(t)}\otimes\hbB^{(t)}-\bC\right\|_F}{\left\|\bC\right\|_F}\asymp \sqrt{\frac{R\log(n)+Rsd_1d_2\log(p_1p_2)}{n}}
\eel
hold with high probability,  where $t_0$ is a certain constant.
\end{cor}
The algorithm \ref{alg:2} is initialized with the top-R left singular vectors of $\sum_{i=1}^n\tbX_iy_i$.
However, unlike one-term SKPD, it is difficult to prove that such initialization satisfies the conditions (\ref{kappa3}) and (\ref{kappa4}). But in our numerical studies, we found that such choice is very stable across a large range of settings. We refer to Section \ref{sec-6} and \ref{sec-7} for more details.

\subsection{Region detection consistency }

In this subsection, we study the region detection consistency of the SKPDs. We shall focus on the multi-term SKPD as it includes the one-term version as a special case. Because the signal regions are indicated by the non-zero elements of $\hbA_r$, we only need to consider the variable selection consistency of $\hbA_r$. 

We first note that estimating $\bbA$ given $\hbbB$ 
could be viewed as a high-dimensional regression problem with noise in the design matrix.  Let $\bba=\tvec(\bbA)$, $\hbba=\tvec(\hbbA)$, $\bbb=\tvec(\bbB)$  and $\hbbb=\tvec(\hbbB)$. Then we may treat $\tbX_i (\hbbb-\bbb)$ as the noise in the designs: 
\bes
y_i&=& \bba^\T\tvec\left(\tbX_i \bbB\right) +\eps_i
\cr\tbX_i \hbbB&=&\tbX_i \bbB + \tbX_i (\hbbB-\bbB).
\ees
The resulted optimization problem could be viewed as
\bes
\min_{\bba} \left\{\frac{1}{2n} \sum_{i=1}^n\left(y_i- \bba^\T\tvec\big(\tbX_i \hbbB\big) \right)^2 + \lambda\|\bba\|_{1} \right\}.
\ees
The high-dimensional regression problem with noise in the design matrix has been studied in the literature, for example, \citet{rosenbaum2010sparse, LohW11,datta2017cocolasso}. In general, the sign consistency of the estimated coefficients is hard to be guaranteed when the design matrix is subject to noise. However, we may follow the strategy of \citet{rosenbaum2010sparse}, where the sign consistency is proved for a hard-thresholded estimator. Specifically, define the hard-thresholded estimator as  $\hbba^{HT}=(\hat{a}_1^{HT},\ldots,\hat{a}^{HT}_{Rp_1p_2})^\T$, 
where
\bes
\hat{a}_j^{HT}=\hat{a}_j{\bf 1}\left(|\hat{a}_j|>c\sqrt{\frac{\log(n)+d_1d_2\log(p_1p_2)}{n}}\right)
\ees
for certain constant $c$. We show that under the following {\it coherence condition}, $\hbba^{HT}$ is sign consistent with true coefficient $\bba$ as long as it is not small.
\begin{condition}[Coherence condition] \label{coh}
Define the matrix 
\bes
\Psi=\frac{1}{n}\sum_{i=1}^n \tvec\big(\tbX_i \bbB\big)\left[\tvec\big(\tbX_i \bbB\big)\right]^\T \in\mathbb{R}^{(Rp_1p_2)\times (Rp_1p_2)}.
\ees
The matrix $\Psi$ satisfies the coherence condition with constants $\psi_1, \psi_2$ if 1) all the diagonal elements satisfy $\min_{j}\Psi_{jj}\ge \psi_1\|\bbb\|_2^2$ and 2) all the off-diagonal elements satisfy $\max_{j}|\Psi_{jk}|\le \psi_2\|\bbb\|_2^2$.
\end{condition}
The coherence condition has been used to study the variable selection consistency in the literature, e.g., \citet{rosenbaum2010sparse}. It is also related to the restricted eigenvalue condition \citep{bickel2009simultaneous} and irrepresentable condition \citet{ZhaoY06}. We refer to \citet{rosenbaum2010sparse} for more discussion.

\begin{thm}[Region Detection Consistency]
\label{th-rd}
Suppose the conditions of Corollary \ref{cor-1} hold. Let $\bba_{\text{nzi}}$ be the indices of nonzero elements of $\bba$ and $s=\sum_{r=1}^R s_r=\|\bba\|_0$. Assume the coherence condition holds with constants $\psi_1$ and $\psi_2$ satisfy $\psi_2s+\ttheta\sqrt{s}<\psi_1$. Let $\hbba^{HT}$ be the hard-thresholded estimator after 
\bes
t\ge t_0+\frac{\log\left(n^{-1}[\log(n)+sd_1d_2\log(p_1p_2)]\right)}{2[\log(\kappa_1)+\log(\kappa_2)]}
\ees
times iteration, where $t_0$ is a certain constant. Then, if
\bes
\min|\bba_{\text{nzi}}|> c\sqrt{\frac{\log(n)+d_1d_2\log(p_1p_2)}{n}},
\ees 
we have
\bes
\sgn(\hbba^{HT})= 	\sgn(\bba). 
\ees
As a result, the region detection consistency can be guaranteed.
\end{thm}

\section{Simulation studies}\label{sec-6}
In this section, we conduct comprehensive numerical studies to demonstrate the region detection and estimation performance of proposed SKPDs under both linear and nonlinear model settings. 

\subsection{Study I: linear models}\label{sec-6-1}
In this subsection, we conduct a simulation study under the linear model 
\bel{sim0}
y_i=\langle\bX_i, \bC \rangle +\eps_i, \quad \eps_i\sim \mathcal{N}(0,\sigma^2),\quad i=1,\ldots,n.
\eel
We investigate the effects of sample size, noise level and signal shape to region detection and estimation. 
Specifically, we fix the image size at $128\times 128$, and consider two different sample size settings, 
$n=500, 1000$, along with two different noise levels $\sigma=1, 3$. The images $\bX_i$ are i.i.d drawn from a $\mathcal{N}(0,1)$ distribution. 

\begin{figure}
\centering
\includegraphics[width=0.9\columnwidth]{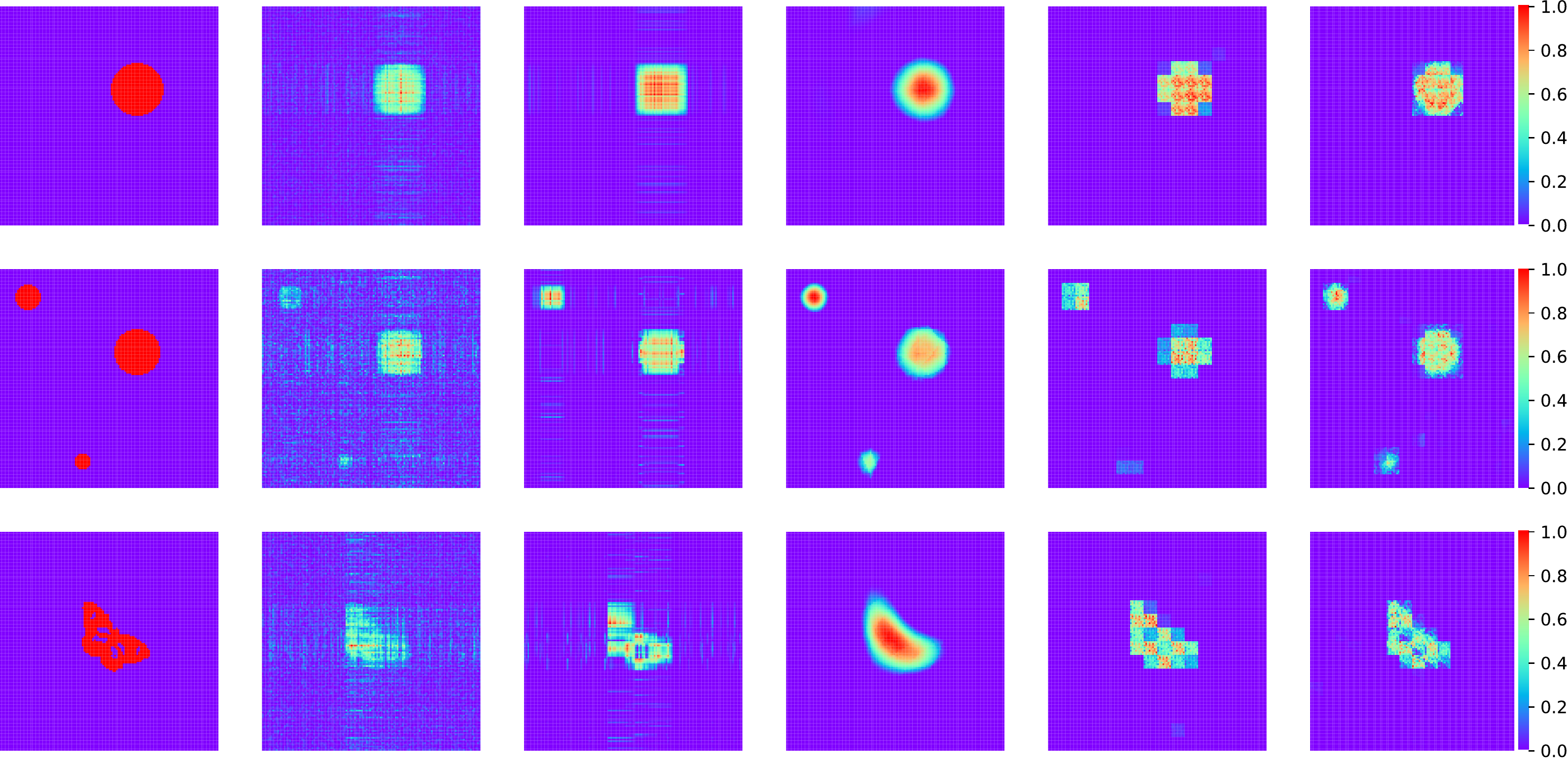}
\caption{An illustration of estimated coefficients $\hbC\in\mathbb{R}^{128\times 128}$ in the linear model simulation with $n=1000$ and noise level $\sigma = 1$. From left to right columns: True signals; MatrixReg; TR Lasso; STGP; one-term SKPD; R-term SKPD. }
\label{fig:heatmap_all}
\end{figure}

Three different coefficients $\bC$ are considered, namely, ``one circle'', ``three circles'' and ``butterfly''. Specifically, we let $\bC_{i,j}=1$ when $(i,j)$ falls in the ``one circle'', ``three circles'' and ``butterfly'' region, and $\bC_{i,j}=0$ otherwise.  When the true signal is ``one circle'', the true coefficients
$\bC$ can be written as $\bA\otimes \bB$ with $\bA=[0,0,0,0;0,0,1,0;0,0,0,0;0,0,0,0]\in\mathbb{R}^{4\times 4}$ and  $\bB\in\mathbb{R}^{32\times 32}$ representing a centered circle with radius 15.
When the true signal is ``three circles'', $\bC$ can be written as $\sum_{r=1}^3\bA_r\otimes \bB_r$ with $\bA_1=[1,0,0,0;0,0,1,0;0,0,0,0;0,0,0,0]$,  $\bA_2=[0,0,0,0;0,0,1,0;0,0,0,0;0,0,0,0]$, $\bA_3=[0,0,0,0;0,0,0,0;0,0,0,0;0,1,0,0]\in\mathbb{R}^{4\times 4}$, and  $\bB_r\in\mathbb{R}^{32\times 32}$ representing three circles with radius 4,13 and 7, respectively. Obviously, the true coefficients $\bC$ could be represented as other combinations of $r$ and $\bA_r$, $\bB_r$ of different sizes. Here we only present one illustration.
When the true signal is ``butterfly'', there is no clear decomposition of $\bC$ (except for the naive approach with $\bA=1$ and $\bB=\bC$). We use the ``butterfly'' to demonstrate complex signals, while use ``one circle'' and ``three circles'' to represent the scenarios of single signal and multiple signals, respectively. See Fig. \ref{fig:heatmap_all} for the signal shape illustrations.

We implement the one-term SKPD and R-term SKPD with rank $R$ tuned by BIC. For both SKPDs, we fix the block size to be $8\times 8$. Note that such a block size is inconsistent with the truth. 
This setup would allow us to better mimic real data scenario and test the performance of SKPDs with mis-specified block sizes.

We evaluate the region detection and estimation performance of SKPDs along with the implementation time. To evaluate region detection performance, the False Positive Rate (FPR) and the True Positive Rate (TPR) are calculated. Specifically, define the FPR as $ \sum\limits_{i=1}^{D_1}\sum\limits_{j=1}^{D_2}\frac{\bI(\hbC_{ij} \neq 0)\cdot\bI(\bC_{ij} = 0)}{\bI(\bC_{ij} = 0)}$ and TPR as $\sum\limits_{i=1}^{D_1}\sum\limits_{j=1}^{D_2}\frac{\bI(\hbC_{ij} \neq 0)\cdot\bI(\bC_{ij} \neq 0)}{\bI(\bC_{ij} \neq 0)}$, where $\hbC$ is the estimated coefficients and $\bI(\cdot)$ is the indicator function. To evaluate estimation performance, we measure the Root Mean Squared Errors (RMSE): $\|\hbC - \bC\|_F/\sqrt{D_1D_2} $. 

The performance of SKPDs are compared with three competing methods: the nuclear norm regularized matrix regression \citep[][denoted as MatrixReg]{zhou2014regularized}, Tensor Regression with Lasso regularization \citep[][denoted as TR Lasso]{zhou2013tensor} and the Bayesian approach based on soft-threshold Gaussian process 
\citep[][denoted as STGP]{kang2018scalar}. 
We note that although nuclear norm based penalization approaches are commonly used in matrix regression, they are unable to detect signal regions as the produced coefficients are non-sparse. Thus, the MatrixReg method will not be compared for region detection, but only for coefficients estimation and computation efficiency. More details on the implementation of different methods are referred to the supplementary material \ref{appendix:b5}.

Our simulation study is based on 100 independent datasets, expect for the  Bayesian approach STGP. Due to the heavy computation of STGP, it takes more than 2 hours for a single implementation on a Linus cluster with Intel Xeon E5-2686 under Amazon Web Services (AWS) when sample size $n=1000$. 
It is impractical to implement it for 100 times. So the STGP is only implemented on the first 5 generated datasets. Considering the fact that we are not studying a very high-resolution image problem ($D_1=D_2=128$), the computational burden is a major obstacle for applying Bayesian approaches to image data analysis.



We report the median of RMSE, FPR, TPR and computation time for different methods in Table \ref{table:all}. In addition, we plot in Fig. \ref{fig:heatmap_all} the estimated coefficients for different signals with sample size $n = 1000$ and noise level $\sigma = 1$ in one repetition. Note that the coefficients estimated by STGP
consists many small (close but not equal to zero) signals that cannot be displayed in Fig. \ref{fig:heatmap_all}. 
By Table \ref{table:all} and Fig. \ref{fig:heatmap_all}, it is clear that both the 1-term and R-term SKPD demonstrate competitive performance on all three aspects: estimation, region detection, and computation time. 
Specifically, for region detection, we can see that both 1-term and R-term SKPD are able to detect more than 95\% of the true signals under most circumstances. Although STGP could achieve a slightly higher TPR, but it  pays a price of larger FPR. Indeed, the FPR of 1-term and R-term SKPD are below 5\% under nearly all the settings, while the FPR of STGP is over 70\% in the butterfly case. For TR Lasso, its region detection performance is very unstable with different signal shapes. For instance, it could only detect less than 50\% true signals under the ``three circles'' case when $n=500$. 

In terms of coefficients estimation, the R-term SKPD and STGP achieve the best  overall performance. 
In particular, when the sample size is large, e.g., $n=1000$, the R-term SKPD achieved the smallest RMSE under the ``butterfly'' case. The performance of 1-term SKPD is not as good as its R-term counterpart when $n=1000$, but still demonstrate competitive performance under most settings. The performance of TR Lasso is still not stable depending on the signal shapes. While for MatrixReg, it present the largest RMSE under nearly all the settings. 

\begin{table}
\caption{\label{table:all}Simulation results for the linear model with different sample size, noise level and signal shapes. The best and second best results are marked by {\color{pinegreen}\bf green} and {\bf bold} respectively.}
\setlength{\tabcolsep}{1mm}{
\begin{tabular}{*{13}{c}}
\hline
\multicolumn{7}{c}{FPR ($\times 100\%$) }& \multicolumn{4}{c}{TPR ($\times 100\%$)}&  \\
Signal &n&$\sigma$ &TR Lasso&STGP&1-tm  &R-tm &&TR Lasso&STGP&1-tm &R-tm  &  \\ 
Circle
&500		&1	&15.6& 43.1	& \bf{2.3}&$\textcolor{pinegreen}{\bf{1.8}}$ &&$\textcolor{pinegreen}{\bf{100.0}}$&$\textcolor{pinegreen}{\bf{100.0}}$ 	& $\bf{98.7}$	&96.7 & \\  
&500		&3	&16.7&34.6	&  \bf{2.3}	&$\textcolor{pinegreen}{\bf{1.9}}$& &$\textcolor{pinegreen}{\bf{100.0}}$&$\textcolor{pinegreen}{\bf{100.0}}$  & $\bf{98.0}$ & 96.7 & \\ 
&1000		&1	&19.1&44.4	&$\textcolor{pinegreen}{\bf{2.6}}$&$\textcolor{pinegreen}{\bf{2.6}}$	 &&$\textcolor{pinegreen}{\bf{100.0}}$&$\textcolor{pinegreen}{\bf{100.0}}$ & $\textcolor{pinegreen}{\bf{100.0}}$  & $\textcolor{pinegreen}{\bf{100.0}}$ 	 &\\ 
&1000		&3	&20.8& 64.3& $\textcolor{pinegreen}{\bf{2.6}}$ &  \bf{2.8}	&&$\textcolor{pinegreen}{\bf{100.0}}$&$\textcolor{pinegreen}{\bf{100.0}}$ 	& $\textcolor{pinegreen}{\bf{100.0}}$ & $\textcolor{pinegreen}{\bf{100.0}}$ 	 & \\ 

3 circles   
&500	&1	&29.6& 19.1	& \bf{4.1} & $\textcolor{pinegreen}{\bf{3.5}}$ & &48.8&$\textcolor{pinegreen}{\bf{100.0}}$ 	&$\bf{93.8}$  & 91.7	& \\ 
&500	&3	&30.6&  19.2	&\bf{4.0} & $\textcolor{pinegreen}{\bf{3.5}}$	&&45.6&$\textcolor{pinegreen}{\bf{100.0}}$ 	&$\bf{93.8}$  	& 91.1	&\\ 
&1000	&1	&44.0& 24.3	&$\textcolor{pinegreen}{\bf{4.1}}$ &  \bf{5.0}	&&94.8&$\textcolor{pinegreen}{\bf{100.0}}$ &96.8 	& $\bf{98.4}$ 	&  \\ 
&1000	&3	&46.2& 28.5&$\textcolor{pinegreen}{\bf{4.0}}$ &  \bf{5.5}&&94.8&$\textcolor{pinegreen}{\bf{100.0}}$ & 96.7 &$\bf{98.4}$ 	&  \\ 

Butterfly   
&500	&1	&29.2& 72.5	& \bf{3.1} & $\textcolor{pinegreen}{\bf{3.0}}$&	&51.1&$\textcolor{pinegreen}{\bf{100.0}}$ 	&$\bf{95.1}$  & 94.4	&  \\ 
&500	&3	&28.8& 77.5	&\bf{3.3} & $\textcolor{pinegreen}{\bf{3.0}}$	& &45.2&$\textcolor{pinegreen}{\bf{100.0}}$ 	&$\bf{95.1}$  &94.7   		& \\ 
&1000	&1	&47.9& 25.4	& $\textcolor{pinegreen}{\bf{3.4}}$ &  \bf{3.5}& &\bf{99.7}&$\textcolor{pinegreen}{\bf{100.0}}$ 		& 97.5  & 98.4 	&  \\ 
&1000	&3	&46.7& 43.6	&$\textcolor{pinegreen}{\bf{3.3}}$ &  \bf{3.6}	&&\bf{99.7}&$\textcolor{pinegreen}{\bf{100.0}}$ 		& 97.5  &98.4	& \\ 
\multicolumn{11}{c}{RMSE($\times 100$)} \\
Signal &n&$\sigma$ &MatrixReg&TR Lasso&STGP&\multicolumn{2}{c}{1-tm SKPD}&\multicolumn{2}{c}{R-tm SKPD}&  \\
Circle    
&500		&1 & 16.3& \bf{8.2} & $\textcolor{pinegreen}{\bf{6.3}}$	&\multicolumn{2}{c}{9.4}  & \multicolumn{2}{c}{10.0}	 & \\ 
&500		&3& 16.5 & \bf{8.2} &$\textcolor{pinegreen}{\bf{6.1}}$	& \multicolumn{2}{c}{9.5}  & \multicolumn{2}{c}{10.1}	& \\ 
&1000		&1& 10.1&$\textcolor{pinegreen}{\bf{6.1}}$ & \bf{7.2}	&\multicolumn{2}{c}{8.7}  & \multicolumn{2}{c}{\bf{7.2}} & \\ 
&1000		&3& 10.6&$\textcolor{pinegreen}{\bf{6.0}}$& \bf{7.4}	& \multicolumn{2}{c}{8.8}  & \multicolumn{2}{c}{\bf{7.4}}	 & \\ 		

3 circles    
&500	&1& 20.5&25.9& $\textcolor{pinegreen}{\bf{7.6}}$		&\multicolumn{2}{c}{\bf{14.4}}  & \multicolumn{2}{c}{15.4}& \\ 
&500	&3& 20.5&26.6& $\textcolor{pinegreen}{\bf{7.6}}$		& \multicolumn{2}{c}{\bf{14.4}}  & \multicolumn{2}{c}{15.9}&\\ 
&1000	&1& 16.3&10.6&$\textcolor{pinegreen}{\bf{8.4}}$	&\multicolumn{2}{c}{13.3}  & \multicolumn{2}{c}{\bf{10.1}}&  \\ 
&1000	&3& 16.4&10.8& $\textcolor{pinegreen}{\bf{8.5}}$	&\multicolumn{2}{c}{13.3}  & \multicolumn{2}{c}{\bf{10.4}}&  \\

Butterfly    
&500	&1& 19.7&25.1& $\textcolor{pinegreen}{\bf{11.6}}$&\multicolumn{2}{c}{\bf{12.6}}  & \multicolumn{2}{c}{14.0}	&  \\ 
&500	&3& 19.9&25.6& $\textcolor{pinegreen}{\bf{11.7}}$	& \multicolumn{2}{c}{\bf{12.7}}  & \multicolumn{2}{c}{14.1}	& \\ 
&1000	&1& 16.1&11.8&  \bf{10.8}	&\multicolumn{2}{c}{11.7}  & \multicolumn{2}{c}{$\textcolor{pinegreen}{\bf{10.6}}$}	&  \\ 
&1000	&3& 16.3&12.0&  \bf{11.0}	&\multicolumn{2}{c}{11.7}  & \multicolumn{2}{c}{$\textcolor{pinegreen}{\bf{10.8}}$}	& \\
\multicolumn{11}{c}{Computational time under butterfly case, minutes }\\
Signal&$n$&$\sigma$&MatrixReg&TR Lasso&STGP&	\multicolumn{2}{c}{1-tm SKPD} & 	\multicolumn{2}{c}{R-tm SKPD}\\ 
Butterfly& 500    &1& $\textcolor{pinegreen}{\bf{0.05}}$&0.31  & 48.09	& \multicolumn{2}{c}{\bf 0.08}&\multicolumn{2}{c}{0.42}  & \\ 
Butterfly&1000   & 1 & $\textcolor{pinegreen}{\bf{0.07}}$ & 2.12   &130.38	&\multicolumn{2}{c}{\bf{0.13}}& \multicolumn{2}{c}{0.76} &
\\
\hline
\end{tabular}}
\end{table}

\begin{table}
\caption{\label{table:nonlinear}Simulation results for nonlinear model under the ``three circles'' signal, $n = 1000, \sigma = 1$.}
\setlength{\tabcolsep}{1.5mm}{
\begin{tabular}{*{8}{lcccccc}}
\hline
Measures       &CNN&TR Lasso&STGP&MatrixReg&1-tm SKPD& R-tm SKPD&NL SKPD& \\ 

FPR($\times 100\%$)    &x&22.5& 24.7& x&$\textcolor{pinegreen}{\bf{1.9}}$&{\bf 1.9} &7.0 & \\ 
TPR($\times 100\%$) &x& 74.4&$\textcolor{pinegreen}{\bf{100.0}}$&x &\bf{88.5} &86.6&84.1 &\\
Prediction error  &$\textcolor{pinegreen}{\bf{11.5}}$&23.2& 21.7 &23.5 & 22.9 &18.5& \bf{14.9}  &\\ 
\hline
\end{tabular}}
\end{table}%

In terms of computation efficiency, 
the SKPD also demonstrate clear advantages. Compared to the TR Lasso, the one-term SKPD is on average 16 times faster when $n=1000$. Compared to the Bayesian approach STGP, the advantage of SKPD is even more significant. 
When the images are of higher resolutions, such advantage could be more significant. 
Note that the Matrixreg is unable to achieve region detection. We list its performance here as a reference.  Also note that the reported time includes all the parameter tunings. 

\subsection{Study II: non-linear model}\label{sec-6-2}
In this subsection, we investigate the performances of SKPDs in the nonlinear model 
\bel{sim1}
y_i=\sum_{r=1}^R\langle \bA_r , \sigma(\bX_i * \bB_r)\rangle+\eps_i,\quad i=1,\ldots, n
\eel
where $*$ is the non-overlapped convolution operator defined in (\ref{conv-def1}) and (\ref{conv-def2}), sample size $n=1000$, $\eps_i\sim \mathcal{N}(0,1)$ and $\sigma(\cdot)$ is the ReLU activation function. The elements of $\bX_i$ are i.i.d generated from $\mathcal{N}(0,1)$ distribution. For the true coefficients $\bA_r$ and $\bB_r$, $r=1,\ldots, R$, we let $R=3$ and $(\bA_r, \bB_r)$ forms the ``three circles'' signal considered in the previous subsection. 


In the nonlinear model,  $\bC=\sum_{r=1}^R \bA_r\otimes \bB_r$ are no longer the true coefficients. Therefore, we do not need to report the estimation error. But as discussed in Section \ref{sec-5}, we could still use the FPR and TPR on $\bC$ to measure the region detection performance in the nonlinear model. 

Beside the region detection performance, we also measure the test set prediction error of different methods. Specifically, we generate an independent test set of size $n_{test}=200$, written as $(\tbX_i^{(new)},y_i^{(new)})$, $i=1,\ldots,n_{test}$. The RMSE of the prediction error is measured by $\sqrt{(1/n_{test})\sum_{i=1}^{n_{test}} \left(\widehat{y}_i^{(new)}-y_i^{(new)}\right)^2}$.

We test the region detection and prediction performance of both linear and nonlinear SKPDs. Specifically, one-term SKPD, R-term SKPD, and R-term nonlinear SKPD with ReLU activation are considered. For all three SKPD approaches, we fix the block size to be $8\times 8$.  Again notice that the dimension of $\hbA$ and $\hbB$ are inconsistent with the truth.
Besides the TR Lasso, STGP, MatrixReg and SKPD models, we also implement a standard CNN with one convolutional layer, one fully connected layer and ReLU activation. 
In the convolutional layer, three filters of size $8\times 8$ are considered and the stride size are set to be $(1,1)$. 
We implement a CNN with stride 1 because it is an ideal benchmark for outcome prediction. Indeed, a the filters in a stride-1 CNN convolute with all the possible blocks of an image, although such a CNN is unable to detect signal regions.  
As in the linear case, we repeat the simulation 100 times and report the median results. The STGP results are still based on 5 times repetition due to computational limitation.



We summarize the results of Study II in Table \ref{table:nonlinear}. The NL SKPD stands for nonlinear SKPD. It is surprised to see that the linear SKPDs still demonstrate satisfactory region detection performance under such a nonlinear model.  By Table \ref{table:nonlinear}, both 1-term and R-term linear SKPDs achieved a TPR greater than 88\% while maintaining a FPR below 2.6\%. Such results further demonstrate the robustness of SKPD. While for nonlinear SKPD, it showed its clear advantages on prediction accuracy and at the same time maintained a competitive performance on region detection. The nonlinear R-term SKPD obtained a prediction error of 14.9, second only to CNN, which is arguably the best approach for image prediction in recent years. 
Besides, the performance of STGP shows a similar pattern as in the linear model. It is able to detect nearly all the signal pixels, but pays more price on FPR. While for TR Lasso, although it also presents an reasonable region detection performance, but still unable to catch up with the SKPDs. 

In summary, we conclude that all SKPDs are able to achieve satisfactory region detection performance even with mis-specified model. While for prediction, the nonlinear SKPD and CNN provide better choices when the true model is nonlinear.

\section{The UK Biobank Study}\label{sec-7}
In this section, we study real brain MRI data collected from a large-scale biomedical database: UK Biobank \href{https://www.ukbiobank.ac.uk/}{(https://www.ukbiobank.ac.uk/)}.
UK Biobank contains in-depth genetic and health information from half a million UK participants. 

This study aims to use the the T1-weighted brain imaging data to detect brain regions that affect individual's visual search ability and psychomotor speed.
A widely used approach to measure such ability is through the Trail Making Test (TMT). In a TMT, the participants are required to link 25 circles marked by numbers 1 to 25 in sequential order as quickly as possible.  The TMT score is then the time taken to correctly link all the 25 circles. More details on the TMT are referred to the UK Biobank description: \href{https://biobank.ctsu.ox.ac.uk/crystal/refer.cgi?id=8481}{https://biobank.ctsu.ox.ac.uk/crystal/refer.cgi?id=8481}. We consider 1500 participants that are involved in the T1-weighted imaging scan and completed the Trail Making Test.


For each partcipants, the brain MRI scan produced a tensor of size $182 \times 218 \times 182$. To improve analytical efficiency, we crop original images to remove the layers of zero-valued voxels. This leads to the processed images of size $144\times 184 \times 144$. We further conduct interpolation to resize the images and the finally obtained images are of size $80 \times 96 \times 80$. We shall note that the T1-weighted images in the UK Biobank have been registered with a MNI template, otherwise further preprocessings, including location registration and intensity normalization would be needed. 

In this problem, how to evaluate the detected regions is not an easy task as we do not know what truly happens in brain. Therefore, in section \ref{sec:study2}, we first consider a simulation study with real brain MRI scan but simulated signals (and responses). This simulation allows us to mimic a real brain region detection problem and evaluate the performance of different methods. While in Section \ref{sec:study3}, we study the real data problem with the response being the TMT score.

\subsection{Real Image and Simulated Response}\label{sec:study2}
In this subsection, we evaluate the performance of SKPDs in the tensor image model:
\bes
y_i=\langle{\bf \mathcal{X}}_i, {\bf \mathcal{C}}\rangle +\eps_i,  \ \ {\eps_i \sim \mathcal{N}(0,\sigma^2)}, \ \ i=1,\ldots, n.
\ees
Here the sample size $n = 1500$ and noise level $\sigma = 3$. We let $\bmX_i\in\mathbb{R}^{80 \times 96 \times 80}$ be the real MRI images in the UK Biobank. Two signal shapes are considered for the true coefficients $\mC$: ``one ball'' and ``two balls''. Specifically, we let $\bmC_{i,j,k} = 1$ when $(i,j,k)$ falls in the ``one ball'' or "two balls'' regions, and $\bmC_{i,j,k} = 0$ otherwise. When the true signal is ``one ball'', the coefficient $\bmC$ can be written as $\bmA \otimes \bmB$ when $\bmA$ and $\bmB$ are of size $5\times 6 \times 5$ and $16\times 16\times 16$, respectively. The true $\bmA\in \mathbb{R}^{5\times 6 \times 5}$ satisfies $\bmA_{3,3,3} = 1$ and $\bmA_{i,j,k}=0$ for $(i,j,k)\neq (3,3,3)$, while the true $\bmB \in \mathbb{R}^{16\times 16 \times 16}$ represents a centered ball with radius 6. When the true signal is ``two balls'', 
$\bmC$ could be written as $\sum_{r=1}^2\bmA_r \otimes \bmB_r$ with $\bmA_1,\bmA_2\in \mathbb{R}^{5\times 6 \times 5}$ and $\bmB_1,\bmB_2\in \mathbb{R}^{16\times 16 \times 16}$. The true $\bmA_1$ and $\bmA_2$ satisfy $(\bmA_1)_{3,3,3} = 1$, $(\bmA_2)_{1,1,3} = 1$ and otherwise zero, while $\bmB_1$ and $\bmB_2$ represents two centered balls with radius 6 and 4, respectively. 
We shall note that although both the ``one ball'' and ``two balls'' signals could be written as the Kronecker product form with a specified size on $\bB$, we still implement the SKPDs under a mis-specified size with $d_1=d_2=d_3=8$. Such mis-specification allows us to further demonstrate the robustness of SKPD.

\begin{figure}[H]
\includegraphics[width=\columnwidth]{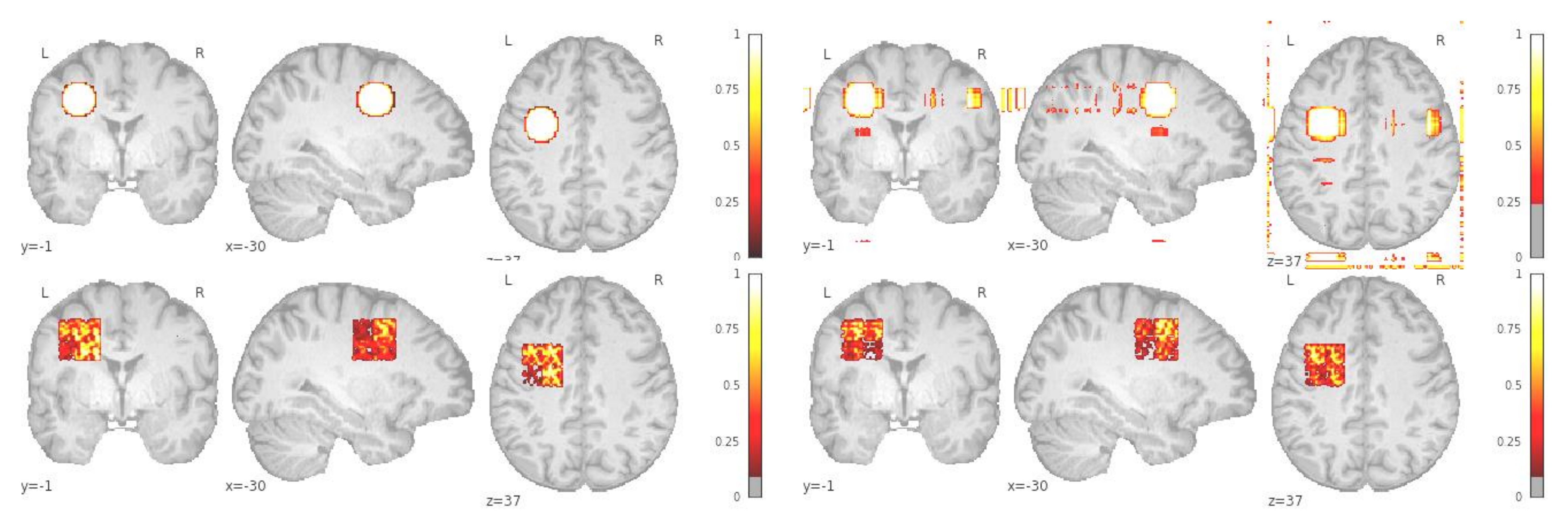}
\vspace{0.1in}
\includegraphics[width=\columnwidth]{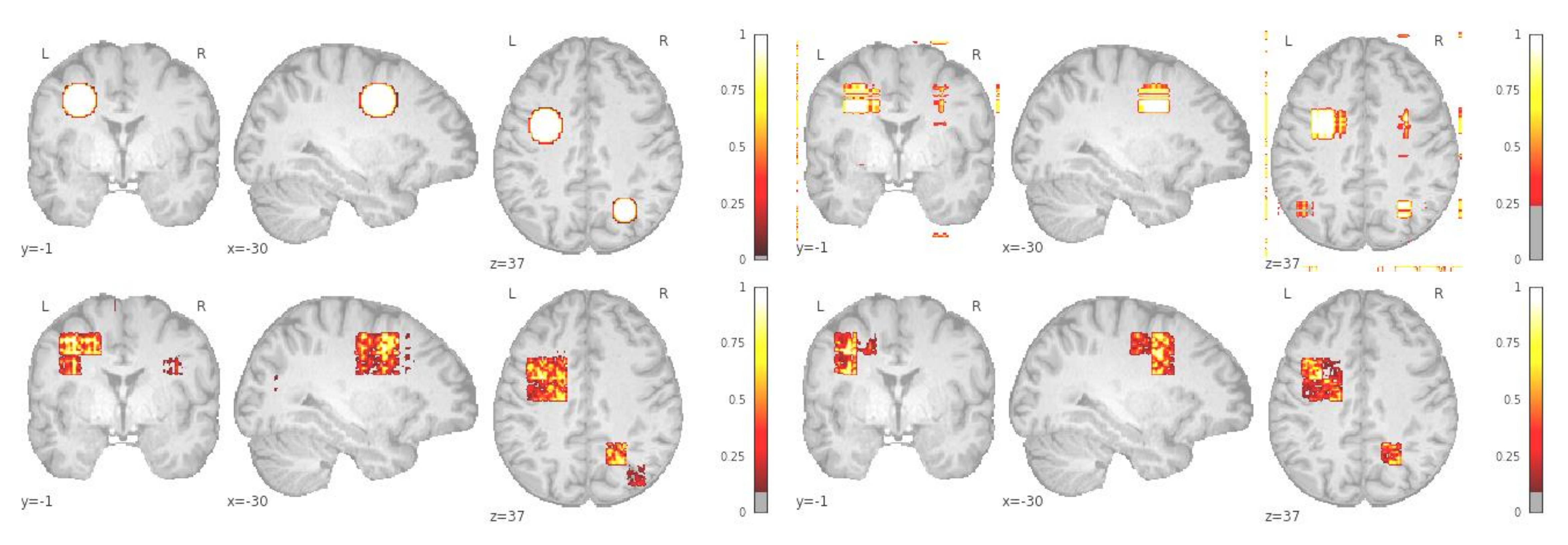}
\caption{An illustration of estimated tensor coefficients in the UK Biobank data. The true signal is ``one ball'' (top two rows) and ``two balls" (bottom two rows). 
For both ``one ball'' and ``two balls'' signal, the plots in a left-to-right and top-to-bottom order are respectively: the true signal in a brain MRI template from the coronal, sagittal and horizontal sections; the TR Lasso estimates; the one-term SKPD estimates; the R-term SKPD estimates.}
\label{fig:Study2}
\end{figure}

The methods implemented in this study include the TR Lasso, one-term and R-term SKPDs. Note that the MatrixReg is no longer applicable under the tensor case. While the STGP can be applied for tensor image theoretically, but practically the computational issue hindered its implementation. 
We plot in Fig. \ref{fig:Study2} the true and estimated tensor coefficients of different methods. The plots illustrate the sagittal, coronal and horizontal sections of the signals.
The median of FPR, TPR and RMSE  under 100 independent repetitions are reported in Table \ref{table:study2}.

\begin{table}
\caption{\label{table:study2}Simulation results under the UK Biobank study, $n = 1500, \sigma = 3$. The best and second best results are marked by {\color{pinegreen}\bf green} and {\bf bold} respectively.}
\centering
\begin{tabular}{*{8}{lcccccc}}
\hline
&Measures &TR Lasso&1-term SKPD& R-term SKPD& \\ 

``One ball''&	FPR($\times 100\%$)    &10.7&\bf{7.7}&$\textcolor{pinegreen}{\bf{3.2}}$& \\ 
&TPR($\times 100\%$)     &\bf{98.2}&$\textcolor{pinegreen}{\bf{100.0}}$ &$\textcolor{pinegreen}{\bf{100.0}}$&\\
&RMSE   &9.4& $\textcolor{pinegreen}{\bf{0.3}}$& $\textcolor{pinegreen}{\bf{0.3}}$  &\\ 
``Two balls''&	FPR($\times 100\%$)  &10.1&\bf{3.7}&$\textcolor{pinegreen}{\bf{3.3}}$& \\ 
&TPR($\times 100\%$)    &76.9&$\textcolor{pinegreen}{\bf{85.2}}$ &\bf{80.7}&\\
&RMSE  &0.9&  $\textcolor{pinegreen}{\bf{0.3}}$& \bf{0.4} &\\ 
\hline
\end{tabular}
\end{table}%

By Table \ref{table:study2} and Fig. \ref{fig:Study2}, it is clear that the SKPD performed consistently well under this tensor case. In the ``one ball'' signal, both the 1-term and R-term SKPDs are able to capture 100\% signals with a small price of FPR (3.2\% for R-term SKPD and  7.7\% for one-term SKPD). Even under the ``two balls'' case, the SKPDs still achieved a TPR over 80\% and a FPR below 5\%. 
We shall also note that the 1-term SKPD demonstrated more advantages compared to R-term SKPD in this case. 
Indeed, under such a tensor case with limited samples, it is more favorable to use 1-term SKPD over the R-term version.
As a comparison, although the TR Lasso could also capture a large part of signal regions, it also include quite some noise in their estimation. 
Consequently, the performance of TR Lasso is dominated by the SKPDs in FPR, TPR and RMSE all three measures. In conclusion, this real MRI based simulation further validated the superior performance of SKPD under different image types.

\subsection{Real data analysis}\label{sec:study3}
In this subsection, we use the TMT score discussed earlier to detect brain regions that affect individual's visual attention and task switching ability. We consider the same $n=1500$ participants. 
The mean and standard error of the TMT scores of the 1500 participants are 37.1 and 11.3, respectively. In addition to brain imaging, we include two additional covariates in this study: sex and age. We first regress the TMT scores to sex and age to remove their effects. Then the residuals are used in the image regression problem.

\begin{figure}[H]
\centering
\includegraphics[width=0.8\columnwidth]{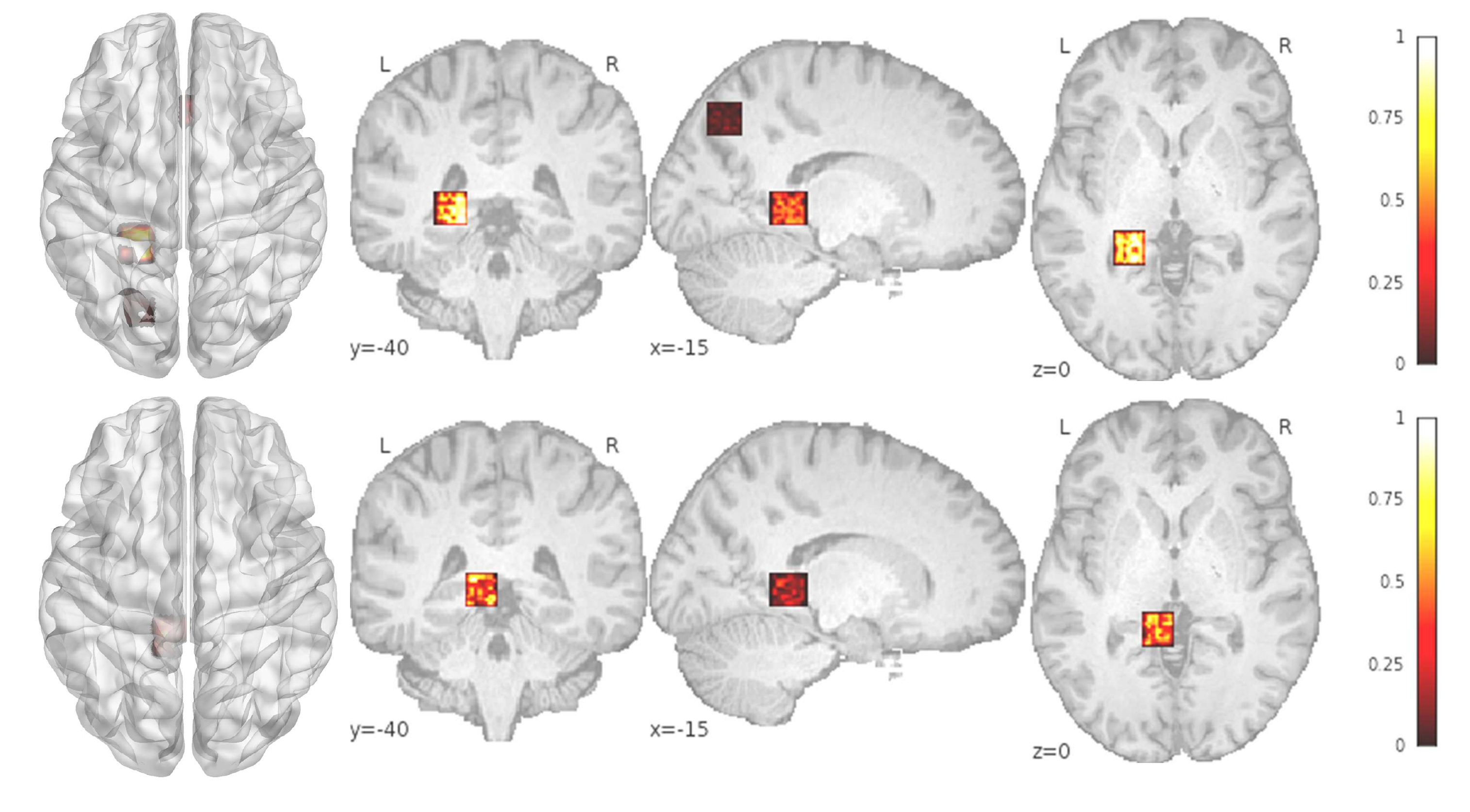}
\caption{(Real data) Tensor coefficients estimated by TR Lasso (first row), one-term SKPD (second row) and R-term SKPD (third row) in the UK Biobank Data. }
\label{fig:Study3}
\end{figure}

Fig. \ref{fig:Study3} plots the estimated tensor coefficients of TR Lasso, one-term and R-term SKPDs in a brain template on coronal, sagittal and horizontal section. We see that two SKPD methods detect clear regions in this study. Moreover, if we look closely, the strongest signal regions detected by both one-term and R-term SKPD are consistent. In contrast, the signals detected by TR Lasso appears to be sporadic.



The strongest signal region detected by both SKPDs is located across the splenium of the corpus callosum and the pineal gland. 
To validate our findings, we first calculated the R-square explained by SKPD detected region: 0.31 (1-term) and 0.34 (R-term). As a comparison, the R-square explanined by TR Lasso region is 0.08.
Moreover, we conducted a permutation test to demonstrate the stability and significance of the detected regions. 
Specifically, we permute the detected region across different patients so that this region would not match with the outcome. We implement SKPDs on the permuted data and check whether this region could be detected. This process is repeated for $N=500$ times. 
Consequently, there are 15 out of the 500 times that the region was detected for one-term SKPD. While for multi-term SKPD, it is 11 times. Under the null hypothesis that such region is independent with the outcome, we may calculate the p-values for finding this region are 15/500=0.03 and 11/500=0.022 for one-term and R-term SKPDs respectively. With a significance level of 0.05, we may reject the null and claim our findings to be significant.

The splenium is the most posterior part of corpus callosum. It contains a large proportion of thick fibers, which is believed to support fusing the hemirepresentation of the visual field \citep{aboitiz1992fiber}.
Moreover, the splenium is connected to the occipital lobe, which has been recognized as the visual processing center of brain 
\citep{grill1998sequence,ungerleider2000mechanisms}.
On the other hand, the pineal gland is a midline brain structure. It produces melatonin and modulate sleep and temperature regulation in both circadian and seasonal cycles \citep{macchi2004human,arendt2005melatonin}. 

Recall that the TMT is designed to test individual's visual search ability and psychomotor speed. It is also considered to be sensitive to frontal lobe damage \citep{macpherson2015handbook} and dementia \citep{salmon2009neuropsychological}. Clearly, the detected regions by SKPD are strongly consistent with the medical findings. Moreover, our results also suggest to further investigate the potential cooperation between the splenium and the pineal gland in visual field. To the best of our knowledge, such 
cooperation has not been explored in the literature yet.

In summary, brain region detection, or brain localization is a fundamental problem in psychology, psychiatry, neuroscience and cognitive science. We also aware that there are still debates between brain region detection, or brain localization, and the holistic aspects of brain function. In particular, the book of \cite{uttal2001new} attacks the idea of brain localization and raise the concerns about ``neo-phrenology''.
On the other hand, \citet{hubbard2003discussion} and \citet{landreth2004localization} disagree with  \cite{uttal2001new} and believe that the two perspectives should not be put on the opposite position. They believe the two theory are, to some extent,  complementary to each other.
We, as statistians, expect our research could provide
statistical tools and supportive evidences for medical researchers to analyze brain imaging data and further contribute to the understanding of human brain.

\newpage
\bibliography{reference}
\appendix
\appendixpage
\addappheadtotoc
\noindent In the supplementary material, we provide the proofs of Theorem \ref{th-1} to Theorem \ref{th-rd} along with three additional lemmas. Moreover, we provide additional simulation results along with more details on the implementions.
\section{Proofs}\label{appendix:proofs}
\noindent{\it \bf Proof of Theorem \ref{th-1}.}
\medskip
We prove Theorem \ref{th-1} by induction. When $t=0$, $\|\hba^{(0)}-\hba\|_2=\mu_0$, we have (\ref{th-1-1}) holds. Then we assume (\ref{th-1-1}) holds for general $t$ and consider the estimation of $\hbb^{(t+1)}$.
Now we define the following matrices,
\begin{align}\label{mat1}
	\bSigma=&(1/n)\sum_{i=1}^n\tbX_i^\T \hba^{(t)}(\hba^{(t)})^\T\tbX_i,	\cr \bTheta=&(1/n)\sum_{i=1}^n\tbX_i^\T\hba^{(t)}\ba^\T\tbX_i, \ \
	\cr \bE=&(1/n) \sum_{i=1}^n\eps_i\tbX_i^\T \hba^{(t)}. 
\end{align}
To estimate $\hbb^{(t+1)}$given the normalized $\hba^{(t)}$, we have
\bes
\hbb^{(t+1)}=\bSigma^{-1}(\bTheta\bb +\bE)=\langle \hba^{(t)},\ba \rangle\bb-\bSigma^{-1}\left(\langle \hba^{(t)},\ba \rangle\bSigma-\bTheta\right)\bb + \bSigma^{-1}\bE .
\ees
It follows by Lemma \ref{lm-1} that
\bel{pf-1-1}
\frac{\|\hbb^{(t+1)}-\bb\|_2}{\|\bb\|_2}\le\underbrace{|1-\langle \hba^{(t)},\ba \rangle|}_{\text{A1}} + \underbrace{\frac{\delta_2}{1-\delta_2}\sqrt{1-\langle \hba^{(t)},\ba \rangle^2}}_{\text{A2}}  + \underbrace{\frac{\|\bSigma^{-1}\bE\|_2}{\|\bb\|_2} }_{\text{A3}}. 
\eel
For A1, we note that $\hba^{(t)}$ and $\ba$ are normalized, thus
\bes
\|\hba^{(t)}-\ba\|_2=\sqrt{2(1-\langle \hba^{(t)},\ba \rangle)}.
\ees
Therefore, 
\bel{pf-1-2}
|1-\langle \hba^{(t)},\ba \rangle|=\frac{1}{2}\|\hba^{(t)}-\ba\|_2^2.
\eel
For A2, we have
\bel{pf-1-3}
\sqrt{1-\langle \hba^{(t)},\ba \rangle^2}\le\sqrt{2(1-\langle \hba^{(t)},\ba \rangle)}=	\|\hba^{(t)}-\ba\|_2.
\eel
where the inequality holds as $\langle \hba^{(t)},\ba \rangle\le 1$.
For A3,
\bel{pf-1-4}
\|\bSigma^{-1}\bE\|_2\le \sigma_*^{-1}(\bSigma)\|\bE\|_2\le \sigma_*^{-1}(\bSigma)\tau_2\le \frac{\tau_2}{1-\delta_1}\le \frac{\tau_2}{1-\delta_2}.
\eel
where the third inequality holds due to the RIP condition.
As $\mu_0\ge (\kappa_2\nu_1+\nu_2)(1-\kappa_1\kappa_2)^{-2}$, by assumption, 
\bel{pf-1-5}
\|\hba^{(t)}-\ba\|_2 \leq (\kappa_1\kappa_2)^{t} \mu_0 + \frac{\kappa_2\nu_1+\nu_2}{1-\kappa_1\kappa_2}\le \mu_0,
\eel
It then follows from (\ref{pf-1-1}) to (\ref{pf-1-5}) that
\bel{pf-1-6}
\frac{\|\hbb^{(t+1)}-\bb\|_2}{\|\bb\|_2}&\le&\left(\frac{\mu_0}{2}+ \frac{\delta_2}{1-\delta_2}\right)\|\hba^{(t)}-\ba\|_2 + \frac{\tau_2}{(1-\delta_2)\|\bb\|_2}\cr 
&\le&\kappa_1\|\hba^{(t)}-\ba\|_2 + \nu_1\cr &\le&  \kappa_1^{t+1}\kappa_2^{t} \mu_0 + \frac{\kappa_1\nu_2 +\nu_1}{1-\kappa_1\kappa_2}
\eel
Thus (\ref{th-1-2}) holds. Moreover,
\bel{pf-1-7}
(1-\kappa_1\mu_0)\|\bb\|_2 - \frac{\tau_2}{1-\delta_2}\le \|\hbb^{(t+1)}\|_2\le (1+\kappa_1\mu_0)\|\bb\|_2 + \frac{\tau_2}{1-\delta_2}.
\eel

Now consider estimate $\hba^{(t+1)}$ given $\hbb^{(t+1)}$. For the non-normalized $\tba^{(t+1)}$,
\bes
\tba\in \min_{\ba} \left\{\frac{1}{2n} \sum_{i=1}^n\left(y_i- \ba^\T\tbX_i \hbb^{(t+1)} \right)^2 + \lambda\|\ba\|_{1}\right\},
\ees
While for the truth
\bes
y_i=\ba^\T\tbX_i \hbb^{(t+1)}+ \underbrace{\left(\eps_i+ \ba^\T\tbX_i (\hbb^{(t+1)}-\bb)\right)}_{\widetilde{\eps}_i}
\ees
We need $\lam$ satisfy 
\bes
\left\|\frac{1}{n}\sum_{i=1}^n \widetilde{\eps}_i(\tbX_i \hbb^{(t+1)})\right\|_\infty\le \frac{\lambda}{2}
\ees
To bound $\left\|\frac{1}{n}\sum_{i=1}^n \widetilde{\eps}_i(\tbX_i \hbb^{(t+1)})\right\|_\infty$, we note that
\bes
\left\|\frac{1}{n}\sum_{i=1}^n \widetilde{\eps}_i(\tbX_i \hbb^{(t+1)})\right\|_\infty\le \underbrace{\left\|\frac{1}{n}\sum_{i=1}^n \eps_i(\tbX_i \hbb^{(t+1)})\right\|_\infty}_{B1} + \underbrace{\left\|\frac{1}{n}\sum_{i=1}^n \tbX_i \hbb^{(t+1)}(\hbh_b^{(t+1)})^\T\tbX_i\ba\right\|_\infty }_{B2}.
\ees
Let $\hbh_b^{(t+1)} = \hbb^{(t+1)} - \bb$. We have for the term B1, 
\bes
\left\|\frac{1}{n}\sum_{i=1}^n \eps_i(\tbX_i \hbb^{(t+1)})\right\|_\infty\le \max_{j,k}\frac{1}{n}\left\|\tbX_{jk}^\T\bep\right\|_2\|\hbb^{(t+1)}\|_2 =\tau_1\|\hbb^{(t+1)}\|_2.
\ees
For the term B2, 
\bes
&&\left\|\frac{1}{n}\sum_{i=1}^n \tbX_i \hbb^{(t+1)}(\hbh_b^{(t+1)})^\T\tbX_i\ba\right\|_\infty 
\cr =&&\max_{jk}\left(\frac{1}{n}\sum_{i=1}^n \tvec\big(\{\bX_i\}^{d_1,d_2}_{jk}\big) \hbb^{(t+1)}(\hbh_b^{(t+1)})^\T\tbX_i\ba\right)
\cr \le&&\max_{jk}\left(\frac{1}{n}\sum_{i=1}^n \Big(\tvec\big(\{\bX_i\}^{d_1,d_2}_{jk}\big)\hbb^{(t+1)}\Big)^2 \right)^{1/2}\left(\frac{1}{n}\sum_{i=1}^n \Big((\hbh_b^{(t+1)})^\T\tbX_i\ba\Big)^2\right)^{1/2}
\cr \le&&\ttheta(1+\delta_2)^{1/2}\|\hbb^{(t+1)}\|_2\|\hbh_b^{(t+1)}\|_2
\cr \le && \ttheta(1+\delta_2)^{1/2}\|\hbb^{(t+1)}\|_2\cdot \lfR{\|\bb\|_2}\left(\kappa_1^{t+1}\kappa_2^{t} \ \mu_0 + \frac{\kappa_1\nu_2 +\nu_1}{1-\kappa_1\kappa_2}\right).
\ees
Therefore, when $\lam^{(t+1)}=2\|\hbb^{(t+1)}\|_2 \left\{\tau_1+\ttheta(1+\delta_2)^{1/2}\left[\kappa_1^t\kappa_2^{t-1} \mu_0 + (\kappa_1\nu_2 +\nu_1)(1-\kappa_1\kappa_2)^{-1}\right]\right\}$, we have 
\bes
\|\tba^{(t+1)}-\ba\|_2&&\le \lfR{\frac{1.5\lam\sqrt{s} }{(1-\delta)\|\hbb^{(t+1)}\|_2^2}}
\cr &&\le \frac{3\tau_1\sqrt{s}(1-\lfR{\delta_2})^{-1}}{\|\hbb^{(t+1)}\|_2} +  \frac{3\ttheta\lfR{\sqrt{s}}(1+\delta_2)^{1/2}(1-\delta_2)^{-1}\left(\kappa_1^{t+1}\kappa_2^{t} \mu_0 + \frac{\kappa_1\nu_2 +\nu_1}{1-\kappa_1\kappa_2}\right)\lfR{\|\bb\|_2}}{\|\hbb^{(t+1)}\|_2} \\
&&\le \frac{3\tau_1\sqrt{s}+3\ttheta\sqrt{s}(1+\lfR{\delta_2})^{1/2}\left(\kappa_1^{t+1}\kappa_2^{t} \mu_0 + \lfR{\frac{\kappa_1\nu_2 +\nu_1}{1-\kappa_1\kappa_2}}\right)\lfR{\|\bb\|_2}}{\|\bb\|_2(1-\kappa\mu_0)(1-\delta_2) -\tau_2} \\
&&\le \frac{1}{2}(\kappa_1\kappa_2)^{t+1} \mu_0 +  \lfR{\frac{\kappa_2\nu_1+\nu_2}{2(1-\kappa_1\kappa_2)}},
\ees
Furthermore, for the normalized $\hba^{(t+1)}$,
\bes
\|\hba^{(t+1)} - \ba \|_2 &&= \|\hba^{(t+1)} -\tba^{(t+1)} + \tba^{(t+1)} -\ba\|_2
\cr &&\le \left|1-\|\tba^{(t+1)}\|_2\right| + \|\tba^{(t+1)} - \ba \|_2
\cr &&\le  \|\tba^{(t+1)} - \ba \|_2 + \|\lfR{\tba^{(t+1)}} - \ba \|_2
\cr &&= 2 \|\lfR{\tba^{(t+1)}} - \ba \|_2
\le (\kappa_1\kappa_2)^{t+1} \mu_0 +  \lfR{\frac{\kappa_2\nu_1+\nu_2}{1-\kappa_1\kappa_2}},
\ees
Finally,
\bes
\|\hba^{(t+1)}(\hbb^{(t+1)})^\T-\ba\bb^\T\|_F &&= \|\hba^{(t+1)}(\hbb^{(t+1)}-\bb)^\T+(\hba^{(t+1)}-\ba)\bb^\T\|_2
\cr &&\le \|\hba^{(t+1)}\|_2\|\hbb^{(t+1)}-\bb\|_2+\|\bb\|_2\|\hba^{(t+1)}-\ba\|_2
\cr &&\le \|\hbb^{(t+1)}-\bb\|_2+\|\bb\|_2\left(\kappa_2\frac{\|\hbb^{(t+1)}-\bb\|_2}{\|\bb\|_2}+ \nu_2\right)
\cr &&\le (1+\kappa_2) \|\hbb^{(t+1)}-\bb\|_2 + \nu_2\|\bb\|_2.
\ees
This completes the proof of Theorem \ref{th-1}.
\bigskip

\noindent {\it \bf Proof of Theorem \ref{th-2}.} To prove Theorem  \ref{th-2}, we need to prove (\ref{asy}) in the main paper holds.  First consider $\tau_1$. 
Note that
\bes
\tau_1=\max_{j,k}\frac{1}{n}\left\|\tbX_{jk}^\T\bep\right\|_2\le \frac{1}{n}\left\{\sum_{l=1}^{d_1d_2}\max_{j,k}\left[\left(\tbX_{jk}\right)_{\cdot,l}^\T\bep\right]\right\}^{1/2}, 
\ees
where $\left(\tbX_{jk}\right)_{\cdot,l}$ is the $l$-th row of $\tbX_{jk}$. When $\bep$ is a subGaussian vector,
\bes
\P\left(\max_{j,k}\left[\left(\tbX_{jk}\right)_{\cdot,l}^\T\bep\right]>\sqrt{n\log(p_1p_2)}\right)\le \sqrt{\log^{-1}(p_1p_2)}.
\ees
As a consequence,
\bes
\P\left(\tau_1\le \sqrt{d_1d_2\log(p_1p_2)/n}\right)\le 1-\sqrt{\log^{-1}(p_1p_2)}\rightarrow 1.
\ees
Now we consider $\tau_2$. we first note that
\bes
\tau_2^2&=&\sup\left\{\frac{1}{n^2}\left\|\sum_{i=1}^n\eps_i\tbX_i^\T \ba\right\|_2^2: \|\ba\|_2=1,\ba\in\mathbb{R}^{p_1p_2}\right\}\cr&=&\sup\left\{\frac{1}{n^2}\left\|\sum_{i=1}^n\eps_i\tbX_i \bb\right\|_2^2: \ \|\bb\|_2=1, \bb\in\mathbb{R}^{d_1d_2}\right\}
\cr &=&\sup\left\{\frac{1}{n^2}\sum_{l=1}^{p_1p_2}\left(\sum_{i=1}^n\eps_i\tbX_{i,l} \bb\right)^2: \quad \|\bb\|_2=1,\bb\in\mathbb{R}^{d_1d_2}\right\}
\ees
By the Hoeffding-type inequality in Proposition 5.10 of \citet{vershynin2010introduction}, we have
\bes
\P\left\{\left(\sum_{i=1}^n\eps_i\tbX_{i,l}\bb\right)^2\ge c_1\log n\sum_{i=1}^n(\tbX_{i,l}\bb)^2\right\}\le \frac{1}{n}
\ees
holds for some constant $c_1$. Also note that $\sum_{i=1}^n(\tbX_{i,l}\bb)^2=\|\tbX_{jk}\bb\|_2^2$.
Thus,
\bes
\P\left(\tau_2^2\ge c_1\frac{\log n}{n^2}\sum_{jk}\|\tbX_{jk}\bb\|_2^2\right) \le \frac{1}{n}
\ees
When $(1/n)\sum_{jk}\|\tbX_{jk}\bb\|_2^2$ are bounded, there exists $c_2$ such that
\bes
\P\left(\tau_2< c_2\sqrt{\frac{\log n}{n}}\right)\rightarrow 1.
\ees
Now we show that Theorem \ref{th-2} follows from Theorem \ref{th-1} and (\ref{asy}). By (\ref{asy}) and the definition of $\nu_1$ and $\nu_2$, we further have $\nu_1\asymp \tau_2\asymp \sqrt{n^{-1}\log(n)}$ and $\nu_2\asymp \sqrt{s}\tau_1\asymp \sqrt{n^{-1}sd_1d_2\log(p_1p_2)}$ hold with high probability. 
It then follows that
\bes
\frac{\kappa_2\nu_1+\nu_2}{1-\kappa_1\kappa_2}\asymp\frac{\kappa_1\nu_2+\nu_1}{1-\kappa_1\kappa_2}\asymp \nu_1+\nu_2\asymp \sqrt{\frac{n+sd_1d_2\log(p_1p_2)}{n}}.
\ees
Therefore, we have after $t$ times iteration, the statistical error for estimating $\hba^{(t)}$, $\hbb^{(t)}$ and $\hbC^{(t)}$ are of order $\sqrt{n^{-1}[n+sd_1d_2\log(p_1p_2)]}$.

\bigskip
\noindent {\it \bf Proof of Theorem \ref{th-3}.}
By Lemma 2.1 of \citet{jain2010guaranteed}
\bes
\left\|\tbX\tvec\big(\hba^{(0)}(\hbb^{(0)})^\T\big)-\by\right\|_2^2\le \|\bep\|_2^2+\frac{\delta_2}{1-\delta_2}\|\tbX\tvec\big(\ba\bb^\T\big)\|_2^2
\ees
By algebra,
\bes
\left\|\tbX\tvec\big(\hba^{(0)}(\hbb^{(0)})^\T-\ba\bb^\T\big)\right\|_2^2
\le \frac{\delta_2}{1-\delta_2}\|\tbX\tvec\big(\ba\bb^\T\big)\|_2^2 \lfR{+} \lfR{2}\bep^\T\left(\tbX\tvec\big(\hba^{(0)}(\hbb^{(0)})^\T-\ba\bb^\T\big)\right)
\ees
On the other hand, by RIP condition, we have $\|\tbX\tvec\big(\ba\bb^\T\big)\|_2^2\le (1+\delta_2)\|\bb\|_2^2$ and
\bes
(1-\delta_2)\big\|\hba^{(0)}(\hbb^{(0)})^\T-\ba\bb^\T\big\|_F^2\le\left\|\tbX\tvec\big(\hba^{(0)}(\hbb^{(0)})^\T-\ba\bb^\T\big)\right\|_2^2\le (1+\delta_2)\big\|\hba^{(0)}(\hbb^{(0)})^\T-\ba\bb^\T\big\|_F^2.
\ees
As a consequence,
\bes
\big\|\hba^{(0)}(\hbb^{(0)})^\T-\ba\bb^\T\big\|_F^2\le	\frac{\delta_2(1+\delta_2)}{(1-\delta_2)^2}\|\bb\|_2^2+\lfR{2} \frac{1+\delta_2}{1-\delta_2}\|\bep\|_2\big\|\hba^{(0)}(\hbb^{(0)})^\T-\ba\bb^\T\big\|_2
\ees
Thus
\bel{pf-2-2}
\big\|\hba^{(0)}(\hbb^{(0)})^\T-\ba\bb^\T\big\|_F&\le& \frac{1}{2}\left\{\lfR{2}\frac{1+\delta_2}{1-\delta_2}\|\bep\|_2+\sqrt{\frac{\lfR{4}(1+\delta_2)^2}{(1-\delta_2)^2}\|\bep\|_2^2+\frac{4\delta_2(1+\delta_2)}{(1-\delta_2)^2}\|\bb\|_2^2}\right\}
\cr &\lfR{\le}&\frac{\lfR{2}(1+\delta_2)(\|\bep\|_2/\|\bb\|_2)+\sqrt{\delta_2(1+\delta_2)}}{1-\delta_2}\|\bb\|_2
\eel
On the other hand,
\bel{pf-2-3}
\big\|\hba^{(0)}(\hbb^{(0)})^\T-\lfR{\ba}\bb^\T\big\|_F^2\ge\big\|\{\bI-\hba^{(0)}(\hba^{(0)})^\T\}\ba\bb^\T\big\|_F^2=\|\bb\|_2^2\left(1-\langle\hba^{(0)},\ba\rangle^2\right),
\eel
Combine (\ref{pf-2-2}) and (\ref{pf-2-3}), we have
\bes
1-\langle\hba^{(0)},\ba\rangle^2\le \lfR{\left(\frac{2(1+\delta_2)(\|\bep\|_2/\|\bb\|_2)+\sqrt{\delta_2(1+\delta_2)}}{1-\delta_2}\right)^2},
\ees
when $\|\bep\|_2/\|\bb\|_2\le \lfR{0.1}(1-\delta_2)$ and $\delta_2\le \lfR{0.1}$, we have
$\langle\hba^{(0)},\ba\rangle\ge \lfR{0.52}$. It further follows that
\bes
\mu_0=\|\hba^{(0)}-\ba\|_2=\sqrt{2(1-\langle\hba^{(0)},\ba\rangle)}< \lfR{0.99},
\ees
and
\bes
\kappa_1= (1/2)\mu_0+ \delta_2(1-\delta_2)^{-1} <\lfR{0.61}.
\ees
Moreover, $\kappa_2\in(0,1)$ also holds as $1-\kappa_1\mu_0>0$ and $s\ll p_1p_2$.
\bigskip

\noindent {\it \bf Proof of Theorem \ref{th-4}.}
We first define the following matrices 
\bes
&&\bbA=[\ba_1,\ba_2,\ldots,\ba_R]\in\mathbb{R}^{(p_1p_2)\times R}
\cr &&\bbB=[\bb_1,\bb_2,\ldots,\bb_R]\in\mathbb{R}^{(d_1d_2)\times R}
\cr &&\hbbA^{(t)}=[\hba_1^{(t)},\hba_2^{(t)},\ldots,\hba_R^{(t)}]\in\mathbb{R}^{(p_1p_2)\times R}
\cr &&\tbbA^{(t)}=[\tba_1^{(t)},\tba_2^{(t)},\ldots,\tba_R^{(t)}]\in\mathbb{R}^{(p_1p_2)\times R}
\cr &&\hbbB^{(t)}=[\hbb_1^{(t)},\hbb_2^{(t)},\ldots,\hbb_R^{(t)}]\in\mathbb{R}^{(d_1d_2)\times R}
\ees
and the resulting vectorization, $\bba=\tvec(\bbA)$,  $\hbba^{(t)}=\tvec(\hbbA^{(t)})$, $\tbba^{(t)}=\tvec(\tbbA^{(t)})$, $\bbb=\tvec(\bbB)$, $\hbbb^{(t)}=\tvec(\hbbB^{(t)})$. Moreover, we let $\tbbX_i$ be
\bes
\tbbX_i = \diag(\tbX_i,\tbX_i,\cdots,\tbX_i) = \left[\begin{array}{ccc}
	\tbX_{i}                       & \cdots & \mathbf{0}_{p_1p_2\times d_1d_2} \\
	\vdots                         & \ddots & \vdots                         \\
	\mathbf{0}_{p_1p_2\times d_1d_2} & \cdots & \tbX_{i}
\end{array}\right]_{(Rp_1p_2) \times (Rd_1d_2)}.
\ees
Finally, we define
\begin{align}
	\label{equa:BCDE}
	\begin{split}
		&\bSigma \stackrel{\text { def }}{=}\left[\begin{array}{ccc}
			\bSigma_{1 1} & \cdots & \bSigma_{1 R} \\
			\vdots    & \ddots & \vdots    \\
			\bSigma_{R 1} & \cdots & \bSigma_{R R}
		\end{array}\right]_{(Rd_1d_2) \times (Rd_1d_2)}, \ \ 
		\bTheta \stackrel{\text { def }}{=}\left[\begin{array}{ccc}
			\bTheta_{11}  & \cdots & \bTheta_{1 R} \\
			\vdots    & \ddots & \vdots    \\
			\bTheta_{R 1} & \cdots & \bTheta_{R R}
		\end{array}\right]_{(Rd_1d_2) \times (Rd_1d_2)},\\
		&\bD \stackrel{\text { def }}{=}\left[\begin{array}{ccc}
			\bD_{1 1} & \cdots & \bD_{1 R} \\
			\vdots    & \ddots & \vdots    \\
			\bD_{R 1} & \cdots & \bD_{R R}
		\end{array}\right]_{(Rd_1d_2) \times (Rd_1d_2)}, \ \ 
		\bE \stackrel{\text { def }}{=}\left[\begin{array}{ccc}
			\bE_{1} \\
			\vdots  \\
			\bE_{R}
		\end{array}\right]_{R \times (d_1d_2)},
	\end{split}
\end{align}
with
\begin{align}
	\bSigma_{uv} &= (1/n)\sum_i \tbX_i^\T\hba_u\hba_v^\T\tbX_i, \ \ \ \bTheta_{uv} = (1/n)\sum_i \tbX_i^\T\hba_u\ba_v^\T\tbX_i, 
	\cr \bD_{uv} &= \langle \hba_u,\ba_v \rangle \cdot \bI_{d_1d_2\times d_1d_2}, \ \ \ \bE_u = (1/n)\sum_i \tbX_i^\T\hba_u\epsilon_i.
\end{align}
Here we suppress the superscript $\hba^{(t)}$ for short. Now we are ready to prove Theorem 5.

To estimate $\hbbB^{(t+1)}$ given an orthonormal matrix $\hbbA^{(t)}$, we have
\bes
\hbbb^{(t+1)} = \bSigma^{-1}(\bTheta \bbb + \bE) = \bD \bbb - \bSigma^{-1}(\bSigma \bD - \bTheta)\bbb + \bSigma^{-1}\bE.
\ees
It follows that
\begin{align}\label{pf-2-1}
	\frac{\|\hbbB^{(t+1)} -\bbB\|_F}{\|\bbB\|_F} \leq \underbrace{\|\bD- \bI_{Rd_1d_2}\|_{op}}_{\text{A1}} + \underbrace{\|\bSigma^{-1}(\bSigma \bD - \bTheta)\|_{op}}_{\text{A2}} + \underbrace{\frac{\|\bSigma^{-1}\bE\|_{op}}{\|\bbB\|_F}}_{\text{A3}}.
\end{align}

Now we bound $A_1$ to $A_3$ separately. For $A_1$, using the spectrum property of Kronecker product, we have
\bes
\|\bD- \bI_{Rd_1d_2}\|_{op} = \|(\bbA^\T\hbbA^{(t)})^\T-\bI_R) \otimes \bI_{d_1d_2}\|_{op} =
\|\bI_R - \bbA^\T\hbbA^{(t)}\|_{op}
\ees
As both $\bbA$ and $\hbbA^{(t)}$ are orthonormal, we have $\|\bbA^\T\hbbA^{(t)}\|_{op} \leq \|\bbA\|_{op}\|\hbbA^{(t)}\|_{op} = 1$. This implies that $\bI_R - \bbA^\T\hbbA^{(t)}$ have no negative eigen values.
As a consequence,
\bel{pf-2-1a}
\|\bI_R - \bbA^\T\hbbA^{(t)}\|_{op} \leq \tr(\bI -  \bbA^\T\hbbA^{(t)}) = R - \tr(\bbA^\T\hbbA^{(t)}) = \frac{1}{2}\|\hbbA^{(t)} - \bbA\|_F^2
\eel
For $A_2$, and $A_3$, \lfR{we derive following inequalities by uding Lemma (\ref{lemma:3})}
\begin{align}
	\label{pf-2-2}
	\|\bSigma^{-1}(\bSigma \bD - \bTheta)\|_{op}  \leq \|\bSigma^{-1}\|_{op}\|(\bSigma \bD - \bTheta)\|_{op}  \leq \frac{\d_{2R}}{1-\d_{2R}}\cdot \|\hbbA^{(t)} - \bbA\|_F.
\end{align}
and
\begin{align}
	\label{pf-2-3}
	\lfR{\|\bSigma^{-1}\bE\|_{op} \leq \|\bSigma^{-1}\|_{op}\|\bE\|_{op} \leq \|\bSigma^{-1}\|_{op} \tau_2\leq \frac{\tau_2}{ 1- \d_{2R}}.}
\end{align}
Combining (\ref{pf-2-1}) to (\ref{pf-2-3}), we have
\bel{pf-2-4}
\frac{\|\hbbB^{(t+1)} -\bbB\|_F}{\|\bbB\|_F} &\leq& \left(\frac{\mu_0}{2} + \frac{\d_{2R}}{1-\d_{2R}}\right)\|\hbbA^{(t)} - \bbA\|_F + \frac{\tau_2}{(1- \d_{2R})\|\bbB\|_F}
\cr &\le&  \kappa_1^{t+1}\kappa_2^{t} \mu_0 + \frac{\kappa_1\nu_2 +\nu_1}{1-\kappa_1\kappa_2}
\eel

Now consider estimate $\tbba^{(t+1)}$ given $\hbbb^{(t+1)}$,
\bes
\tbba\in \min_{\bba} \left\{\frac{1}{2n} \sum_{i=1}^n\left(y_i- \bar{\ba}^\T \tbbX_i \hbbb^{(t+1)} \right)^2 + \lambda\|\bba\|_{1}\right\}.
\ees
Similar to the one-term case,
\bes
y_i=\bba^\T\tbbX_i \hbbb^{(t+1)}+ \underbrace{\left(\eps_i+ \bba^\T\tbbX_i (\bbb - \hbbb^{(t+1)})\right)}_{\widetilde{\eps}_i}
\ees
We need $\lam$ satisfy
\bes
\left\|\frac{1}{n}\sum_{i=1}^n \widetilde{\eps}_i(\tbbX_i \hbbb^{(t+1)})\right\|_\infty\le \frac{\lambda}{2}.
\ees
To bound $\left\|\frac{1}{n}\sum_{i=1}^n \widetilde{\eps}_i(\tbbX_i \hbbb^{(t+1)})\right\|_\infty$, we note that
\bes
&&\left\|\frac{1}{n}\sum_{i=1}^n \widetilde{\eps}_i(\tbbX_i \hbbb^{(t+1)})\right\|_\infty
\cr\le &&\underbrace{\left\|\frac{1}{n}\sum_{i=1}^n \eps_i(\tbbX_i \hbbb^{(t+1)})\right\|_\infty}_{B1} + \underbrace{\left\|\frac{1}{n}\sum_{i=1}^n \tbbX_i \hbbb^{(t+1)}\bba^\T   \tbbX_i   \hbh_b^{(t+1)}\right\|_\infty }_{B2}
\cr \le&&\lfR{\tau_1\|\hbbb^{(t+1)}\|_2 +\sqrt{R}\ \ttheta(1+\delta_2)^{1/2}\|\hbbb^{(t+1)}\|_2\left(\kappa_1^{t+1}\kappa_2^{t} \ \mu_0 + \frac{\kappa_1\nu_2 +\nu_1}{1-\kappa_1\kappa_2}\right).}
\ees
where $\hbh_b^{(t+1)} = \hbbb^{(t+1)} - \bbb$. For the term B1,
\bes
\left\|\frac{1}{n}\sum_{i=1}^n \eps_i(\tbbX_i \hbbb^{(t+1)})\right\|_\infty\le \max_{j,k}\frac{1}{n}\left\|\tbX_{jk}^\T\bep\right\|_2\|\hbbb^{(t+1)}\|_2 =\tau_1\|\hbbb^{(t+1)}\|_2.
\ees
\lfR{where $\tau_1$ is same as (14).}	For term B2,
\bes
&&\left\|\frac{1}{n}\sum_{i=1}^n \tbbX_i \hbbb^{(t+1)}\bba^\T\tbbX_i\hbh_b^{(t+1)}\right\|_\infty
\cr =&&\max_{jk,r}\left(\frac{1}{n}\sum_{i=1}^n \tvec^\T\big(\{\bX_i\}^{d_1,d_2}_{jk}\big) \hbb_r^{(t+1)}\bba^\T\tbbX_i\hbh_b^{(t+1)}\right)
\cr \le&&\max_{jk,r}\left(\frac{1}{n}\sum_{i=1}^n \Big(\tvec^\T\big(\{\bX_i\}^{d_1,d_2}_{jk}\big)\hbb_r^{(t+1)}\Big)^2 \right)^{1/2}\left(\frac{1}{n}\sum_{i=1}^n \Big(\bba^\T\tbbX_i\hbh_b^{(t+1)}\Big)^2\right)^{1/2}
\cr \le&&\max_{j,k,r}\|\tbX_{jk}\|_2\|\hbb_r^{(t+1)}\|_2 \left\{(1+\delta_2)^{1/2}\|\bba\|_2\|\hbh_b\|_2\right\}
\cr \le&& \sqrt{R}\ \ttheta(1+\delta_2)^{1/2}\|\hbbb^{(t+1)}\|_2\|\hbh_b^{(t+1)}\|_2
\cr \le && \sqrt{R}\ \ttheta(1+\delta_2)^{1/2}\|\hbbb^{(t+1)}\|_2\left(\kappa_1^{t+1}\kappa_2^{t} \ \mu_0 + \frac{\kappa_1\nu_2 +\nu_1}{1-\kappa_1\kappa_2}\right)\lfR{\|\bbb\|_2}.
\ees
Therefore, 
\bes
&&\|\tbba^{(t+1)}-\bba\|_2\cr\le &&\lfR{\frac{1.5\lam\sqrt{s} }{(1-\delta_{2R})\|\hbbb^{(t+1)}\|_2^2}}
\cr \le&& \frac{3\tau_1\sqrt{s}(1-\delta_{2R})^{-1}}{\|\hbbb^{(t+1)}\|_2} +  \frac{3\sqrt{R}\ttheta(1+\delta_{2R})^{1/2}(1-\delta_{2R})^{-1}\left(\kappa_1^{t+1}\kappa_2^{t} \ \mu_0 + \frac{\kappa_1\nu_2 +\nu_1}{1-\kappa_1\kappa_2}\right)\lfR{\|\bbb\|_2}}{\|\hbbb^{(t+1)}\|_2} \\
\le&& \frac{3\tau_1\sqrt{s}+3\sqrt{R}\ttheta\sqrt{s}(1+\delta_{2R})^{1/2}\left(\kappa_1^{t+1}\kappa_2^{t} \ \mu_0 + \frac{\kappa_1\nu_2 +\nu_1}{1-\kappa_1\kappa_2}\right)\lfR{\|\bbb\|_2}}{\|\bbb\|_2(1-\kappa\mu_0)(1-\delta_{2R}) -\tau_2} \\
=&& \frac{1}{2}(\kappa_1\kappa_2)^{t+1} \mu_0 +  \frac{\kappa_2\nu_1+\nu_2}{2(1-\kappa_1\kappa_2)},
\ees
As a consequence,
\bes
\|\tbbA^{(t+1)}-\bbA\|_F \leq \frac{1}{2}(\kappa_1\kappa_2)^{t+1} \mu_0 +  \frac{\kappa_2\nu_1+\nu_2}{2(1-\kappa_1\kappa_2)},
\ees
Furthermore, we can now show that the nearest orthomormal matrix $\hbbA = \tbbA(\tbbA^\T \tbbA)^{-\frac{1}{2}}$. It is easy to verify that $\hbbA$ constructed as above is orthonormal, i.e., $\hbbA^\T \hbbA = \bI_R$.	By standard nearest orthonormal matrix results in  \cite{horn1988closed}, we have
\bes
\|\hbbA-\tbbA\|_F^2   \leq \|\bbA-\tbbA^{(t+1)}\|_F^2 .
\ees
As a consequence,
\bes
\|\hbbA^{(t+1)}-\bbA\|_F &&\leq \|\hbbA^{(t+1)}-\tbbA^{(t+1)}\|_F + \|\tbbA^{(t+1)}-\bbA\|_F\\
&&\leq \|\bbA-\tbbA^{(t+1)}\|_F + \|\tbbA^{(t+1)}-\bbA\|_F \\
&&= 2\|\tbbA^{(t+1)}-\bbA\|_F.
\ees
Finally,
\bes
\left\|\sum_{r=1}^R\hbA^{(t)}\otimes\hbB^{(t)}-\sum_{r=1}^R\bA\otimes\bB\right\|_F&=&\|\hbbA^{(t)}(\hbbB^{(t)})^\T-\bbA\bbB^\T\|_F\cr&=&\|\hbbA^{(t)}(\hbbB^{(t)}-\hbbB)^\T+(\hbbA^{(t)}-\bbA)\bbB^\T\|_F\cr&\le&\|\hbbA^{(t)}\|\|\hbbB^{(t)}-\hbbB\|_F+\|\hbbA^{(t)}-\bbA\|_F\|\bbB\|_F
\cr&\le&\sqrt{R}\|\hbbB^{(t)}-\hbbB\|_F+\|\bbB\|_F\left(\kappa_2\frac{\|\hbbB^{(t+1)}-\bbB\|_F}{\|\bbB\|_F}+ \nu_2\right)
\cr &\le& (\sqrt{R}+\kappa_2) \|\hbbB^{(t+1)}-\bbB\|_F + \nu_2\|\bbB\|_F.
\ees
\bigskip

\noindent {\it \bf Proof of Theorem \ref{th-rd}.}
We first note that 
\begin{align}\label{pf-4-1}
	\left\|\Psi\left(\hba^{(t)}-\ba\right)\right\|_\infty=&\frac{1}{n}\left\|\sum_{i=1}^n \tbX_i \bb\bb^\T\tbX_i\left(\hba^{(t)}-\ba\right)\right\|_\infty
	\cr =&\max_{jk}\left(\frac{1}{n}\sum_{i=1}^n \tvec\big(\{\bX_i\}^{d_1,d_2}_{jk}\big) \bb\bb^\T\tbX_i\big(\hba^{(t)}-\ba\big)\right)
	\cr \le&\max_{jk}\left(\frac{1}{n}\sum_{i=1}^n \Big(\tvec\big(\{\bX_i\}^{d_1,d_2}_{jk}\big)\bb\Big)^2 \right)^{1/2}\left(\frac{1}{n}\sum_{i=1}^n \Big[\bb^\T\tbX_i\big(\hba^{(t)}-\ba\big)\Big]^2\right)^{1/2}
	\cr \le&\ttheta(1+\delta_2)^{1/2}\|\bb\|_2\|\hba^{(t)}-\ba\|_2
	\cr \le & c\ttheta\sqrt{\frac{\log(n)+sd_1d_2\log(p_1p_2)}{n}}\|\bb\|_2^2.
\end{align}
holds for some constant $c$, where the last inequality holds due to Theorem \ref{th-2}.  On the other hand, as the $j$-th component of $\left\|\Psi\left(\hba^{(t)}-\ba\right)\right\|_\infty$ is
\bel{pf-4-2}
\left(\Psi\big(\hba^{(t)}-\ba\big)\right)_j = \Psi_{jj}\big(\hba^{(t)}_j-\ba_j\big)+\sum_{k=1,k\neq j}^{p_1p_2}\Psi_{jk}\big(\hba^{(t)}_k-\ba_k\big)
\eel
It then follows from (\ref{pf-4-1}) and (\ref{pf-4-2}) that
\begin{align}
	\psi_1\|\bb\|_2^2\left\|\hba^{(t)}-\ba\right\|_\infty\le c\ttheta\sqrt{\frac{\log(n)+sd_1d_2\log(p_1p_2)}{n}}\|\bb\|_2^2 +\psi_2\|\bb\|_2^2
	\left\|\hba^{(t)}-\ba\right\|_1
\end{align}
Therefore,
\begin{align}
	\left\|\hba^{(t)}-\ba\right\|_\infty\le &c\frac{\ttheta+\psi_2\sqrt{s}}{\psi_1}\sqrt{\frac{\log(n)+sd_1d_2\log(p_1p_2)}{n}}
	\cr\le &c\sqrt{\frac{\log(n)+d_1d_2\log(p_1p_2)}{n}}.
\end{align}
As a consequence, the selection consistency can be guaranteed when 
\bes
\min_{l\in S}|a_j|> c\sqrt{\frac{\log(n)+d_1d_2\log(p_1p_2)}{n}}.
\ees 
\bigskip

\begin{lemma}\label{lm-1}
	Let $\bSigma$, $\bTheta$ and $\bE$ be as in (\ref{mat1}). Suppose $\|\hba^{(t)}\|_2=\|\ba\|_2=1$.	
	Then
	\bes
	\left\|\bSigma^{-1}\left(\langle \hba^{(t)},\ba \rangle\bSigma-\bTheta\right)\right\|_{op} \le \frac{\delta_2}{1-\delta_2}\sqrt{1-\langle \hba^{(t)},\ba \rangle^2} .
	\ees
\end{lemma}

We omit the proof of Lemma \ref{lm-1} as it can be found in Lemma 4.3 of \citet{jain2013low}.
\medskip

\begin{lemma}
	\label{lemma:RIP}
	Suppose matrix $\bX\in\mathbb{R}^{n \times (D'D'')} $ satisfies the 2R-RIP condition with constant $\d_{2R}$, then we have
	\bes
	\left|\left\langle\frac{1}{n}\bX\tvec(\bU), \frac{1}{n}\bX \tvec(\bV)\right\rangle - \langle \bU,\bV\rangle\right|	\leq \d_{2R}\|\bU\|_F\|\bV\|_F
	\ees
	holds for any matrix $\bU, \bV \in \mathbb{R}^{D'\times D''}$, if they satisfy $\rank(\bU) \leq 2R$ and $\rank(\bV) \leq 2R$.
\end{lemma}
We omit the proof of Lemma \ref{lemma:RIP} as it can be found in Lemma B.1 of \citet{jain2013low}.
\medskip

\begin{lemma}
	\label{lemma:3}
	Let $\bSigma$, $\bTheta$ and $\bE$ be as in (\ref{equa:BCDE}) with both $\hbA^{(t)}$ and $\bA$ being orthonormal. Then we have
	\bes
	\|\bSigma^{-1}\|_{op} \le \frac{1}{1-\delta_{2R}} 
	\ees 
	and
	\bes  
	\|\bSigma\bD - \bTheta\|_{op} \leq \d_{2R}\|\hbbA^{(t)} - \bbA\|_F .
	\ees
\end{lemma}
\bigskip

\noindent {\it \bf Proof of Lemma \ref{lemma:3}.}
Define any vector $\bw,\bz \in \mathbb{R}^{d_1d_2}$, such that matrix $\bW = [\bw_1,\bw_2,\ldots,\bw_R], \bZ = [\bz_1,\bz_2,\ldots,\bz_R]$ satisfies $\|\bW\|_F = 1,\|\bZ\|_F = 1$ respectively. Denotes vector $\bw_v = (\bw_1^\T,\bw_2^\T,\ldots,\bw_R^\T)^\T, \bz_v = (\bz_1^\T,\bz_2^\T,\ldots,\bz_R^\T)^\T$.
First, we observe 
\bes
\bw_v^\T\bSigma\bw_v &&= \sum_{p= 1}^{R}\sum_{q= 1}^{R}\bw_p^\T\bSigma_{pq}\bw_q \cr&&= \sum_{p= 1}^{R}\sum_{q= 1}^{R}\bw_p^\T\left((1/n)\sum_i \tbX_i^\T\hba_p\hba_q^\T\tbX_i\right)\bw_q \\
&&= \frac{1}{n}\sum_{i=1}^{n}\left(\sum_{p= 1}^{R}\bw_p^\T\tbX_i^\T\hba_p\right)\left(\sum_{q= 1}^{R}\hba_q^\T\tbX_i\bw_q\right) \cr&&= \frac{1}{n}\sum_{i=1}^{n}\left(tr(\tbX_i^\T\hbbA\bW^\T)\right)^2\\
&& \geq (1-\delta_2)\|\hbbA\bW^\T\|_F^2 \\&&= (1-\delta_2)\|\hbbA\|_F^2\|\bW^\T\|_F^2 \\
&&= 1 - \delta_2.
\ees 
The second to last inequality holds by using Lemma (\ref{lemma:RIP}). Hence, we have 
\bes
\|\bSigma^{-1}\|_{op} \le \frac{1}{1-\delta_{2R}}.
\ees
Second, our purpose is to calculate
\bes
\|\bSigma\bD - \bTheta\|_{op} = \max\limits_{\|\bw_v\|_2,\|\bz_v\|_2=1} \bw_v^\T(\bSigma\bD - \bTheta)\bz_v = \sum\limits_{p = 1}^{R}\sum\limits_{q = 1}^{R}\bw_p(\bSigma\bD - \bTheta)_{pq}\bz_q.
\ees
For $(\bSigma\bD - \bTheta)_{pq}$, we have
\bes
(\bSigma\bD - \bTheta)_{pq} &&= \sum_{l=1}^{R} \bSigma_{p l} \bD_{l q}-\bTheta_{p q} \cr&& = \frac{1}{n}\sum_{l=1}^{R}\sum_{i=1}^n \tbX_i^\T\hba_p\hba_l^\T(\hba_l^\T\ba_q)\tbX_i - \frac{1}{n}\sum_{i=1}^{n} \tbX_i^\T\hba_p\ba_q^\T\tbX_i\\
&&= \frac{1}{n}\sum_{i=1}^n \tbX_i^\T\hba_p\ba_q^\T(\sum_{l=1}^{R}\hba_l\hba_l^\T)\tbX_i - \frac{1}{n}\sum_{i=1}^{n} \tbX_i^\T\hba_p\ba_q^\T\tbX_i \\
&& = \frac{1}{n}\sum_{i=1}^n \tbX_i^\T\hba_p\ba_q^\T(\hbbA\hbbA^\T - \bI_{d_1d_2})\tbX_i.
\ees
Therefore,

\bes
\bw_v^\T(\bSigma\bD - \bTheta)\bz_v &&= \sum\limits_{p = 1}^{R}\sum\limits_{q = 1}^{R}\bw_p^\T \cdot \left(\frac{1}{n}\sum_{i=1}^n \tbX_i^\T\hba_p\ba_q^\T(\hbbA\hbbA^\T  - \bI_{d_1d_2})\tbX_i\right)\bz_q \\
&& = \frac{1}{n}\sum_{i=1}^n\sum\limits_{p = 1}^{R}\sum\limits_{q = 1}^{R}\bw_p^\T\tbX_i^\T\hba_p\ba_q^\T(\hbbA\hbbA^\T - \bI_{d_1d_2})\tbX_i\bz_q \\
&& = \frac{1}{n}\sum_{i=1}^n \tr\left(\tbX_i^\T\hbbA\bW^\T\right)\tr\left(\tbX_i^\T(\hbbA\hbbA^\T  - \bI_{d_1d_2})\bbA\bZ^\T\right) \\
&& \leq \langle \hbbA\bW^\T,(\hbbA \hbbA^\T  - \bI_{d_1d_2})\bbA\bZ^\T \rangle + \d_{2R}\|\hbbA\bW^\T\|_F\|(\hbbA \hbbA^\T - \bI_{d_1d_2})\bbA\bZ^\T \|_F.
\ees
The last inequality holds since we suppose $\tbX$ satifies RIP and use lemma (\ref{lemma:RIP}). Then, we have
\bes
\langle \hbbA\bW^\T,(\hbbA\hbbA^\T - \bI_{d_1d_2})\bbA\bZ^\T \rangle = \langle (\hbbA\hbbA^\T - \bI_{d_1d_2})\hbbA\bW^\T,\bbA\bZ^\T \rangle = 0
\ees
Noting that $\|\hbbA\|_F = 1$, $\|\bW\|_F = 1$ and $\|\bZ\|_F = 1$. Thus
\bes
&&\|\hbbA\bW^\T\|_F\|(\hbbA\hbbA^\T - \bI_{d_1d_2})\bbA\bZ^\T \|_F \cr\le&&  \|(\hbbA\hbbA^\T - \bI_{d_1d_2})\bbA\|_F \cr=&& \sqrt{R - \tr\left(\bbA^\T\hbbA(\bbA^\T\hbbA)^\T \right)}\\
\leq&&  \sqrt{R - \frac{1}{R}(\tr(\bbA^\T\hbbA))^2} \\
=&&  \sqrt{R - \frac{1}{R}\left(\frac{1}{2}\|\hbbA - \bbA\|_F^2 - R\right)^2} \\
=&&  \sqrt{\|\hbbA - \bbA\|_F^2\cdot \left(1- \frac{1}{4R}\|\hbbA - \bbA\|_F^2\right)}\\
\leq&&  \|\hbbA - \bbA\|_F.
\ees
Therefore,
$\|\bSigma\bD - \bTheta\|_{op} \leq \d_{2R}\|\hbbA - \bbA\|_F$.

\section{Additional Simulations} \label{appendix:sumulation}
\subsection{The effects of rank $R$}
In this subsection, we conduct simulation study to demonstrate the performance of SKPD under different rank $R$. In particular, we implement 1-term, 3-term, 5-term SKPD along with the $R$-term SKPD with $R$ tuned by the BIC criteria discussed in the paper. In addtion, we also compare the performance of SKPD with a local smoothing method, in which a local average is taken over a block and Lasso is applied on the downsized image. That is to say, the local smoothing approach can be viewed as a special case of our one-term SKPD with the dictionary $\bB$ being an all-one matrix. 

\begin{figure}
	\centering
	\includegraphics[width=0.9\columnwidth]{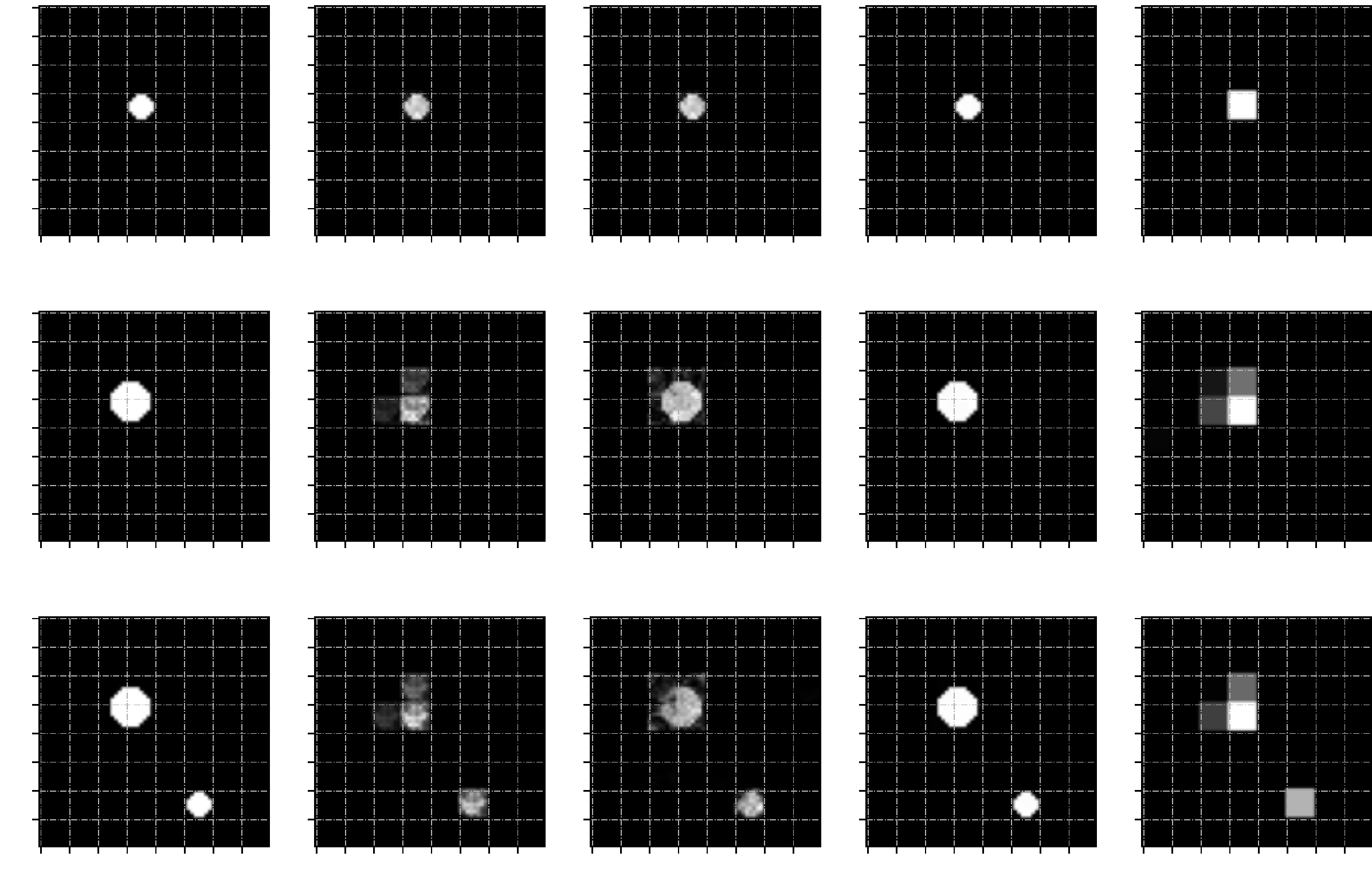}
	\caption{An illustration of coefficients estimated by local smoothing and SKPDs with different rank $R$. From left to right: True signal; 1-term SKPD; 3-term SKPD; 5-term SKPD; Local smoothing. Clearly, the signal shapes could be clearly recovered by multi-term SKPDs, but not by local smoothing.}
	\label{fig:study5}
\end{figure}

\begin{figure}
	\centering
	\includegraphics[width=\columnwidth]{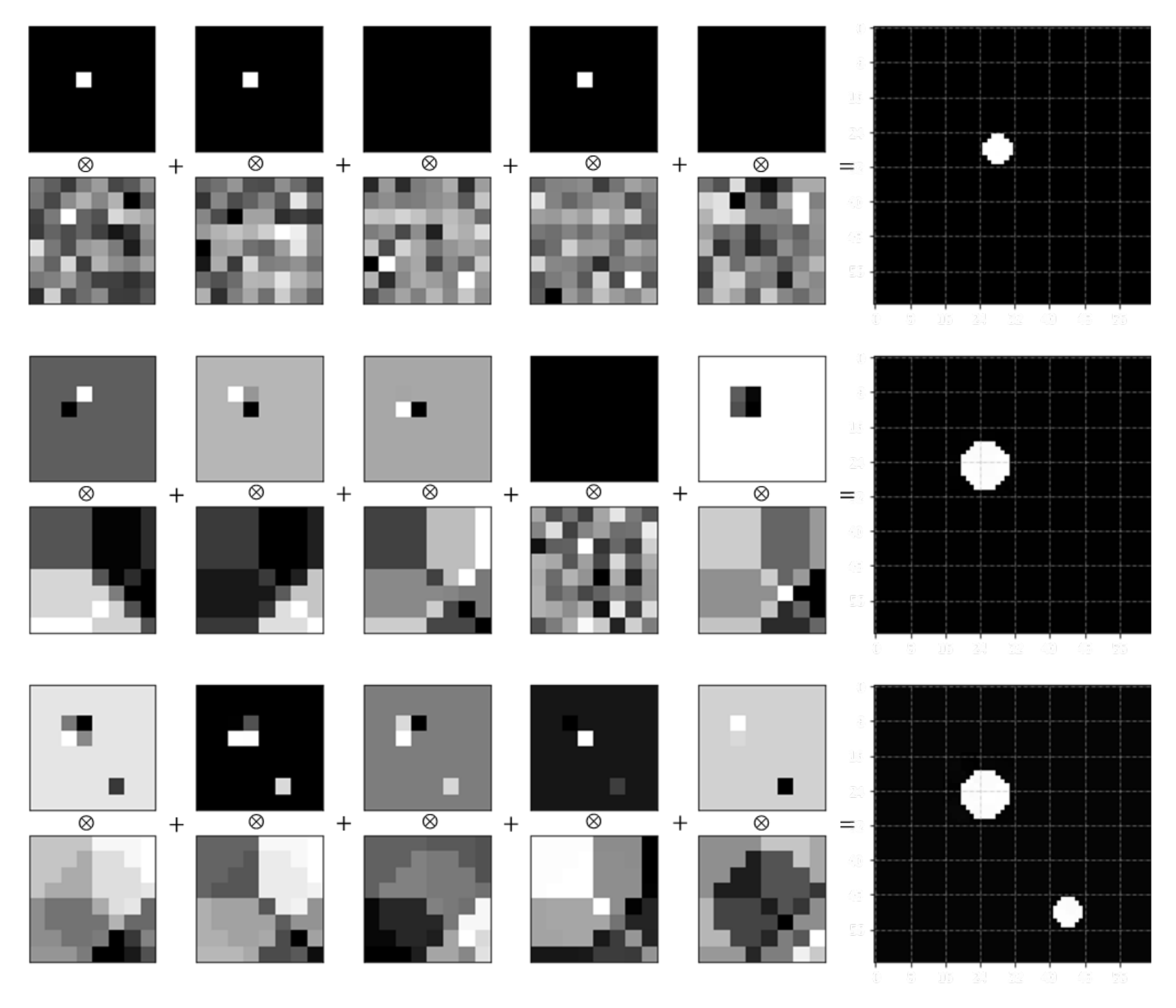}
	\caption{An illustration of separate terms estimated by 5-term SKPD across different signal shapes.}
	\label{fig:study5-5-terms}
\end{figure}

Three signal shapes are considered: 1)``one-circle'' within a block exactly, 2)``one-circle'' but not in one block, and 3) ``two-circles'' that is a combination of above two cases. Here sample size $n = 1000$ and image is of size $D_1 = D_2 = 64$. For local smoothing and all SKPD method, we fix $d_1 = d_2 = 8$. We plot the estimated coefficients in Figure \ref{fig:study5} and report the estimation/region detection accuracy in Table 
\ref{table:study5-1}. To better illustrate SKPD, we also plot in Figure \ref{fig:study5-5-terms} the separate terms estimated by the multi-term SKPD. Clearly, SKPD outperforms local smoothing. The estimation performance of SKPD could be significantly improved when we increase the SKPD terms $R$. As a comparison, it is impossible for local smoothing.

\begin{table}[H]
	\caption{\label{table:study5-1}A comparison between 1-term and R-term SKPD along with local smoothing.}
\centering
\setlength{\tabcolsep}{2.5mm}{
	\begin{tabular}{*{10}{lcccccc}}
		\hline
		\\
		\multicolumn{4}{c}{FPR($\times 100\%$) }& \multicolumn{3}{c}{TPR($\times 100\%$) }&
		\multicolumn{3}{c}{RMSE ($\times 100$) }
		\\
		
		&one-tm &R-tm & local & one-tm &R-tm & local &one-tm &R-tm & local & \\ 
		Case 1	   &0.7&0.7&0.7 &100.0 &100.0&100.0&1.2&1.2&6.2&\\
		Case 2	 &2.5& 7.1&4.0 &93.8 &100.0&93.8&9.8&4.5&10.5& \\ 
		Case 3 &3.2& 7.8&4.7 &95.5 &100.0&95.5&11.3&6.9&12.3& & 
		\\
		\hline
\end{tabular}}
\end{table}%

\subsection{The effects of grid sizes}

In this subsection, we restrict our attention to 1-term SKPD and consider the effects of grid sizes. Specifically, consider images of size $D_1=D_2=120$ and 8 different sizes of $\bB$: $(d_1,d_2)=(1,1), (2,2),(5,5), (8,8), (10,10), (12,12), (15,15), (20,20)$. 
Note that under the case $(d_1,d_2)=(1,1)$, the SKPD reduced to standard Lasso. 
We report the region detection and estimation results in Table 
\ref{table:bs} below. 
First note that besides the extreme case $(d_1,d_2)=(1,1)$, the coefficients $\bC$ is unable to be written as the form $\bA\otimes \bB$ under any of the other settings. However, the special Lasso case performs the worst in terms of both estimation and region detection. As a comparison, 1-term SKPD with ``moderately small'' blcoks, e.g., $(d_1,d_2)=(5,5), (8,8), (10,10), (12,12)$, performs particularly well under two different signal shapes.

\begin{figure}[H]
\centering
	\includegraphics[width=\columnwidth]{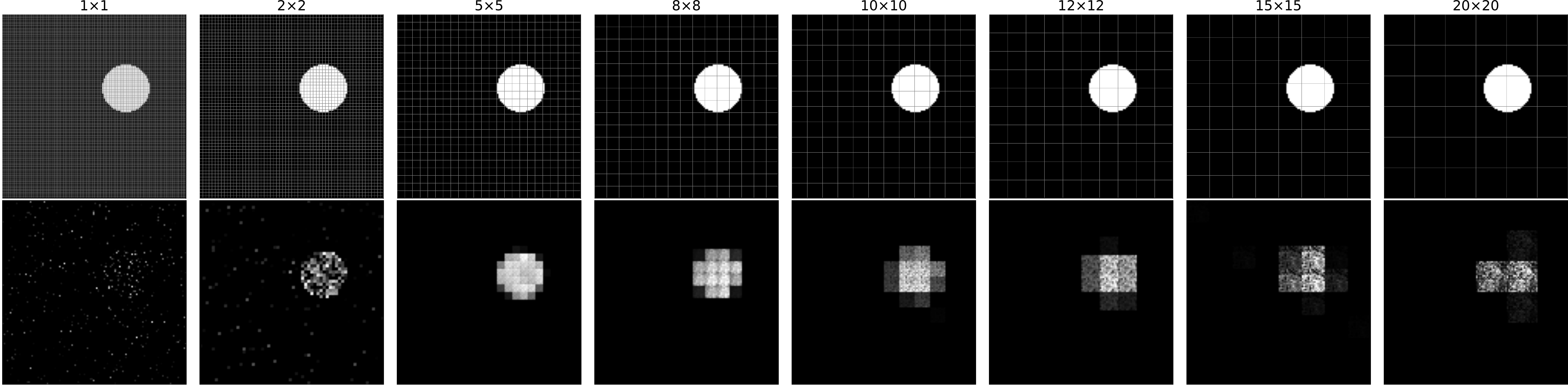}
	\includegraphics[width=\columnwidth]{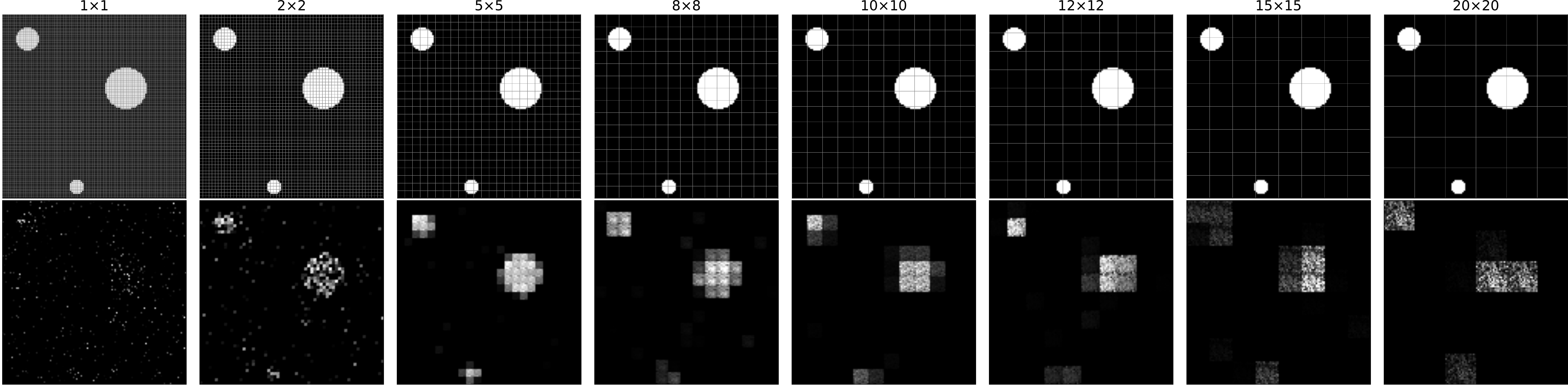}
\caption{An illustration of the coefficients estimated by 1-term SKPD under different $(d_1,d_2)$. The true signals are``one circle'' and ``three circles'' . Row 1 and Row 3 represent true signal, Row 2 and Row 4 the estimated coefficients by one-term SKPD.}
\label{fig:bs}
\end{figure}

\begin{table}
\caption{\label{table:bs} The performance of one-term SKPD under different $(d_1,d_2)$. The true signal is ``one circle'' and ``three circles'' signals, sample size $n = 1000$, noise level $\sigma = 1$.}
\centering
\setlength{\tabcolsep}{2mm}{
	\begin{tabular}{*{8}{lcccccc}}
		\hline
		\multicolumn{8}{c}{``one circle''} & 
		\\
		Measures &$1 \times 1$&$2 \times 2$      &$5 \times 5$&$8 \times 8$&$10 \times 10$&$12 \times 12$& $15 \times 15$& $20 \times 20$& \\ 
		
		FPR($\times 100\%$)   &0.7&0.6&1.9 & 3.0 & 5.2&5.1 &  7.8 &12.3 & \\ 
		TPR($\times 100\%$)    &6.2&37.8&99.0& 100.0&98.8 &98.0 &98.9&92.5 &\\
		RMSE ($\times 100$) &22.6&20.9&8.5& 9.4 &12.5 & 12.1 &14.9& 18.0  &\\ 
		\\
		\multicolumn{8}{c}{``three circles''} & 
		\\
		Measures     &$1 \times 1$&$2 \times 2$   &$5 \times 5$  & $8 \times 8$&$10 \times 10$&$12 \times 12$& $15 \times 15$& $20 \times 20$& \\ 
		
		FPR($\times 100\%$)   &0.8&0.8&3.0& 4.9& 6.5&7.0&12.9 &21.0 & \\ 
		TPR($\times 100\%$)   &6.4&36.2 &96.1& 96.8&93.0 &95.0 &92.5&86.1 &\\
		RMSE ($\times 100$)&23.8&22.0 &11.7& 14.2 &14.4 & 14.4 &18.8&22.6  &\\ 
		\hline
\end{tabular}}
\end{table}%

\subsection{The effects of grid sizes + ranks}
\begin{figure}[H]
	\centering
	\includegraphics[width=1\columnwidth]{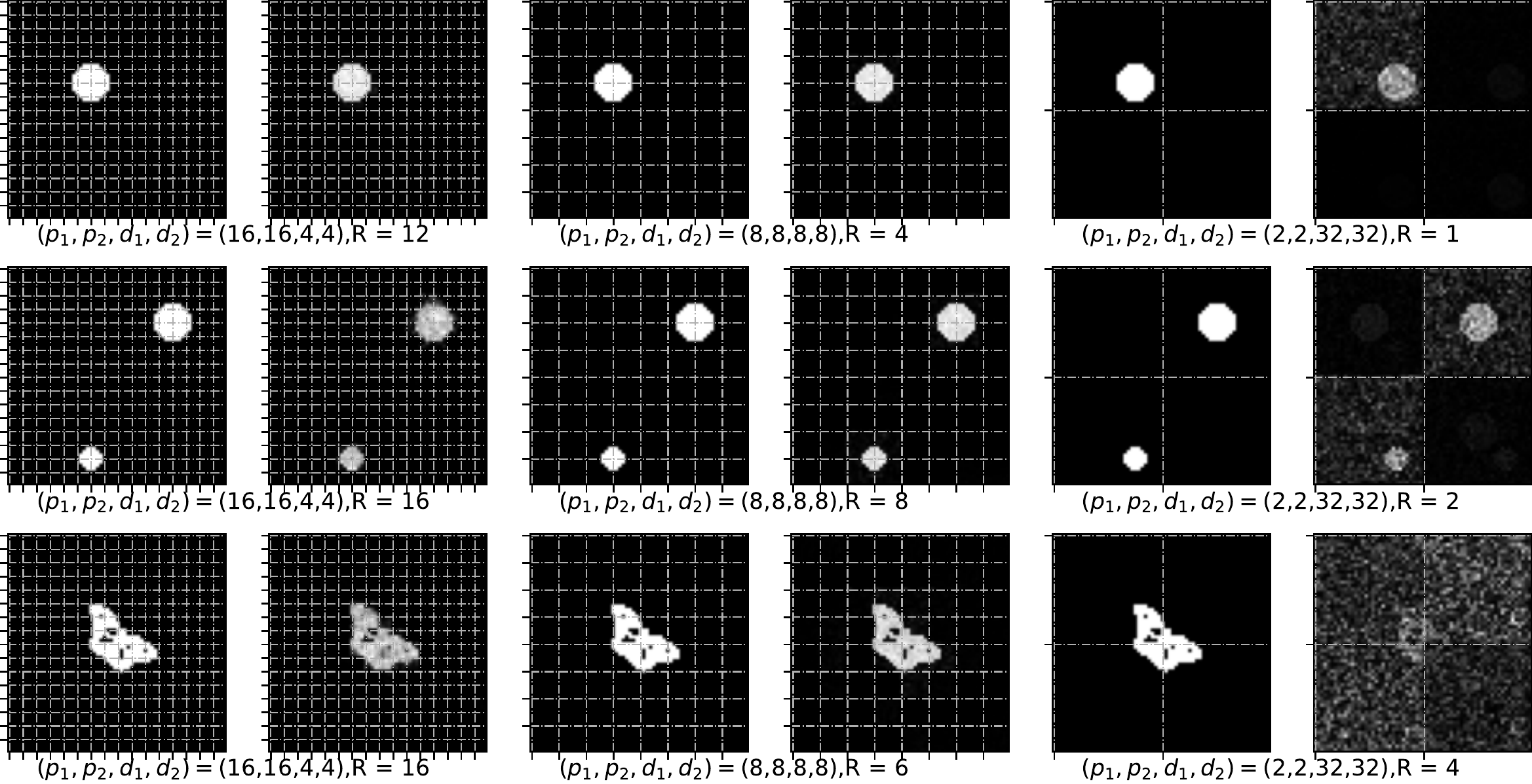}
	\caption{The true signals (Columns 1,3,5) and the SKPD estimates (Columns 2,4,6) under different rank and block sizes. In this simulation,  $D_1=D_2=64$, $n=1000$.}
\label{fig:Q1(c)}
\end{figure}

In this subsection, we vary both the grid sizes and ranks and check their combined effects to SKPD. The results are illustrated in Figure \ref{fig:Q1(c)}. Note that columns 1, 3, 5 are the true signals under different block partitions, columns 2,4,6 are the corresponding SKPD estimations.
Clearly, the SKPD performed well with relatively small block sizes (first two cases). But when the blocks becomes very large, the $\bA_r$s are not sparse enough and SKPD did not demonstrate a satisfactory performance, especially under the butterfly case. The results further support our choice of ``moderately small'' blocks.

\subsection{Comparisons between SKPD and local smoothing}
In this subsection, we compare the performance of SKPD with different types of smoothing filters. Specifically, we  considered mean local smoothing with different grid sizes: $2 \times 2, 4 \times 4$ and $8\times 8$ (denoted as loc-2, loc-4 and loc-8 respectively). Moreover, the Gaussian filter is also considered (denoted as GF). Their performance are compared with SKPD (grid size $(d_1,d_2)=(8\times 8)$, $R$ tuned by BIC) under five cases with different signal shapes and intensities. In Cases 1-3, we consider three different signal shapes (one circle, one big circle, two circles), but all with intensity 1. In Case 4, the signal intensity is ``Gaussian filter style'', i.e., strong signals in the center and weak outside. In Case 5, the signal intensity is generated by standard normal distribution. Figure \ref{fig:study5_round3} below illustrates the five signal shapes and their corresponding estimates of different methods. Table \ref{tables2} summarize the region detection and estimation performance.
The SKPD clearly out-performs local smoothing approaches under all five cases for both region detection and estimation.

\begin{figure}[H]
\centering
\includegraphics[width=\columnwidth]{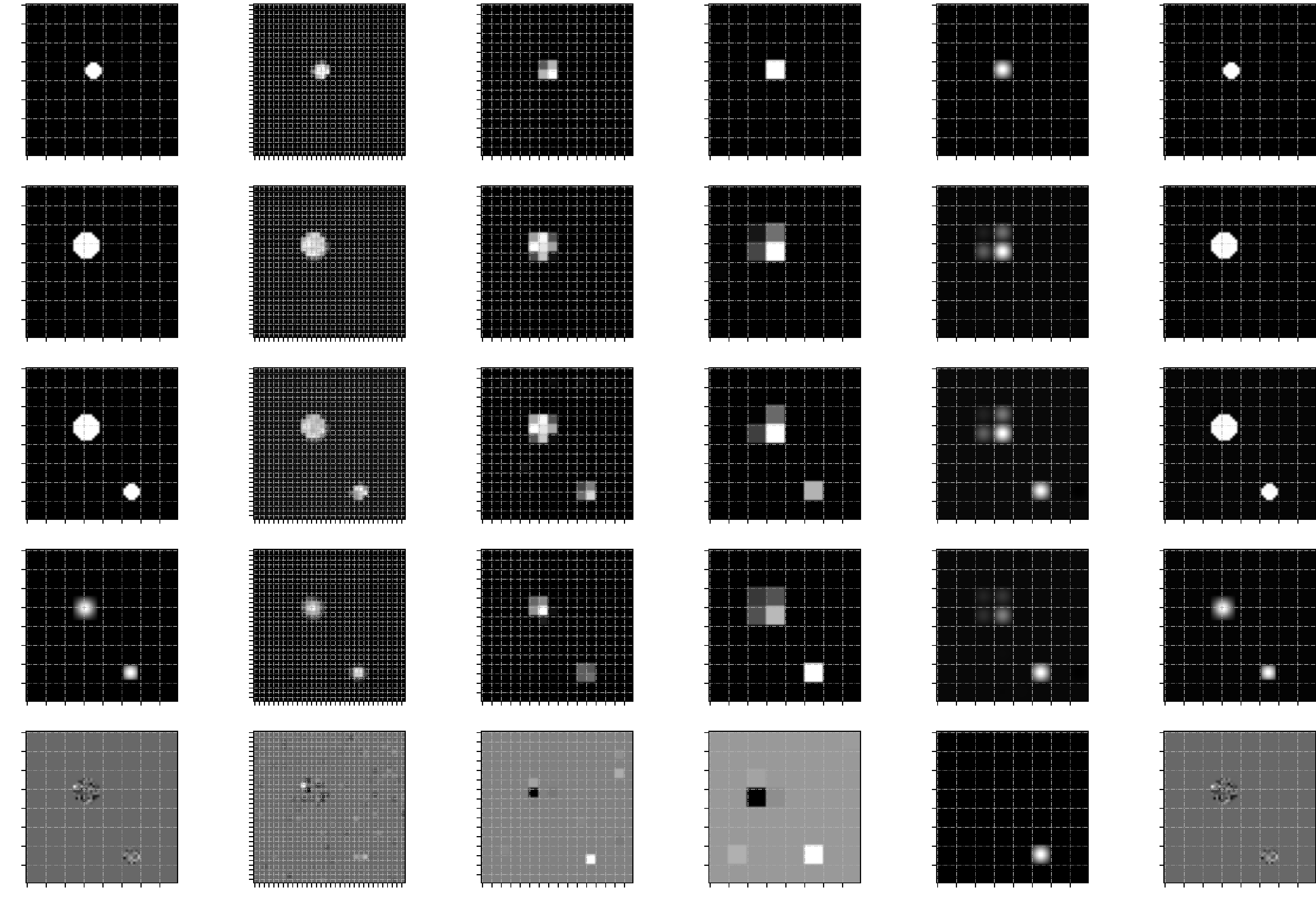}
\caption{An illustration of the estimation results of different local smoothings and SKPD under various signal shapes and intensities. From left to right: True signal, Local smoothing with grid sizes $(2\times 2), (4 \times 4)$ and $(8 \times 8)$, Local smoothing with Gaussian filter, and SKPD.  In this simulation,  $D_1=D_2=64$, $n=1000$.}
\label{fig:study5_round3}
\end{figure}

\begin{table}
\caption{\label{tables2} Region detection and estimation performance of different local smoothing methods and SKPD in simulation study. }
\centering
\setlength{\tabcolsep}{2.3mm}{
\begin{tabular}{*{11}{lcccccc}}
	\hline
	\multicolumn{6}{c}{TPR($\times 100\%$) }& \multicolumn{6}{c}{FPR($\times 100\%$) }
	\\
	
	& loc-2 &loc-4& loc-8 &GF & SKPD&&  & loc-2 &loc-4& loc-8 &GF&SKPD& \\ 
	Case 1	  &97.3&100.0&100.0& 100.0&100.0 && &0.5 &1.1&0.7&2.2&0.7 &\\
	Case 2	 &99.0&96.9&93.8&89.7&100.0&&&1.2 &2.0&4.0&4.0& 7.1 & \\ 
	Case 3 &97.8&97.8&95.5&95.5&100.0& &&2.0 &2.3 &4.7&4.7& 7.8 &
	\\
	Case 4 &81.0&91.2&100.0&88.2&100.0&&&0.7 &2.1 &6.7&3.4&4.6 &  \\
	Case 5 &61.6&61.2&59.7&27.6&100.0&&&16.0 &25.1 &5.2&0.7&6.3 & 
	\\
	\\
	\multicolumn{8}{c}{RMSE($\times 100$) }
	\\
	
	& loc-2 & loc-4 &loc-8 & \multicolumn{1}{c}{GF }&SKPD& \\ 
	Case 1	 &4.4 & 6.0&6.2 &4.4 &1.2&&\\
	Case 2	&6.1 &7.3&10.5 & 11.6& 4.5& \\ 
	Case 3 &7.8 &9.7 &12.3 &12.4 & 6.9& \\
	Case 4 &5.0 &6.0&7.9& 7.3& 1.2&  \\
	Case 5 &16.5&17.3&17.5& 17.6& 3.3 & \\
	\hline
\end{tabular}}
\end{table}%

\subsection{Additional implementation details} \label{appendix:b5}
We provide in this subsection additional details on the implementation of different approaches.
MatrixReg and TR Lasso were both implemented by Matlab toolbox ``TensorReg''. For MatrixReg, the only one parameter that need to be selected is penalty level $\lambda$, which is chosen by the Bayesian information criterion. For TR Lasso, we follow the suggestions of \cite{zhou2013tensor} and set the term rank $r = 3$ for the Gaussian image. The tuning parameter $\lambda$ is selected by BIC from a range of $1000$ to $5000$.  We shall note that the performance of TR Lasso is rather unstable under the UK Biobank study. The three-term TR Lasso is unable to produce a reasonable solution even with carefully tuned $\lambda$. Therefore, we also implemented TR Lasso with one and two terms. The reported results are based on the best performance.   The STGP was implemented use the R package ``STGP''. STGP has four tuning parameters: the number of knots $(m_1,m_2)$ on each axis and prior for the threshold $(t_{min},t_{max})$. For the Gaussian image, we set all parameters on default values because of the heavy computation. For our SKPDs, we let $\lambda$ range from 0.4 to 2.
Our Nonlinear SKPD can be implemented with Pytorch as other CNN models. As we mentioned before, the nonlinear SKPD is equivalent to a two-layer CNN with one convolutional layer and one fully-connected layer. We use 3 filters of size is $8 \times 8$. 
The learning rate was set to 0.02 and times decay weight 0.98 at every 10 epochs. Besides, the number of epochs was set to 100. We take the Adam Optimizer for trainning with mini-batches of size $32$. We evaluate the prediction error for the predefined candidate set to tune the hyper-parameter $\lambda$.
For CNN, we decrease stride size from $8 \times 8$ to $1 \times 1$, and keep other settings as the Nonlinear SKPD.


\end{document}